\documentclass[10pt,twocolumn,letterpaper]{article}

\usepackage{iccv}              

\usepackage{csquotes}
\usepackage{subcaption}
\usepackage{stfloats}
\usepackage{placeins}

\definecolor{iccvblue}{rgb}{0.21,0.49,0.74}
\usepackage[pagebackref,breaklinks,colorlinks,allcolors=iccvblue]{hyperref}

\def\paperID{*****} 
\def\confName{ICCV}
\def\confYear{2025}

\title{Sea-ing Through Scattered Rays: Revisiting the Image Formation Model for Realistic Underwater Image Generation}

\author{Vasiliki Ismiroglou\thanks{Corresponding author}
\and
Malte Pedersen
\and
Stefan H. Bengtson
\and
Andreas Aakerberg
\and
Thomas B. Moeslund
\\Visual Analysis and Perception Laboratory, Aalborg University, Denmark\\
Pioneer Centre for Artificial Intelligence, Denmark\\
{\tt\small \{vasilikii, mape, shbe, anaa, tbm\}@create.aau.dk}}

\begin{document}
\maketitle
\begin{abstract}
In recent years, the underwater image formation model has found extensive use in the generation of synthetic underwater data. Although many approaches focus on scenes primarily affected by discoloration, they often overlook the model's ability to capture the complex, distance-dependent visibility loss present in highly turbid environments. In this work, we propose an improved synthetic data generation pipeline that includes the commonly omitted forward scattering term, while also considering a nonuniform medium. Additionally, we collected the BUCKET dataset under controlled turbidity conditions to acquire real turbid footage with the corresponding reference images. Our results demonstrate qualitative improvements over the reference model, particularly under increasing turbidity, with a selection rate of 82. 5\% by survey participants. Data and code can be accessed on the project page: \href{https://vap.aau.dk/sea-ing-through-scattered-rays/}{vap.aau.dk/sea-ing-through-scattered-rays}.

\end{abstract}  
\section{Introduction}
\label{sec:intro}
Underwater computer vision has gained significant momentum in recent years, driven largely by the growing urgency of environmental monitoring. As marine ecosystems face unprecedented threats from climate change, pollution, and overfishing, there is a need for tools that can support long-term ecological assessment and conservation efforts. Given the significant progress of data-driven methods in more common terrestrial applications, they present a strong foundation for addressing this problem. However, underwater settings come with unique challenges compared to their above-sea counterparts.

\begin{figure}[!t]
    \centering
    \includegraphics[width=\linewidth]{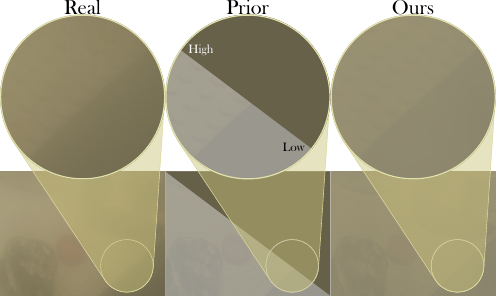}
    \caption{In high-turbidity environments, blur due to forward scattering is not negligible. Left: the real underwater image. Middle: results from a standard IFM using both low and high attenuation settings. Right: our proposed model under high attenuation. When attenuation is overestimated (middle), backscatter dominates before blur becomes perceptible. Best viewed when zoomed in.}
    \label{fig:blurdisplay}
\end{figure}

The physical underwater environment makes collecting data particularly hard. Among other things, getting recording equipment underwater is not trivial, as it requires specialized casings, divers, or expensive remotely operated vehicles (ROVs) and a degree of maintenance that is not always feasible or practical. 

Another hindrance is the low visibility that comes as a result of the properties of the medium. Light is attenuated differently in water compared to the atmosphere, with longer wavelengths absorbed at very short depths, leading to discolorations of the footage \cite{mobley_light_1994}. Natural light does not penetrate the medium after a certain depth, making artificial lighting necessary. That in turn amplifies backscattering and introduces new degradations, such as uneven illumination \cite{sooknanan_improving_2012, wang_underwater_2025}. The attenuation caused by seawater is further intensified by the presence of suspended particles and sediment, commonly referred to as turbidity. Seawater composition is often described using the Jerlov classification scheme \cite{jerlov_optical_1957}, which distinguishes between five coastal (1C, 3C, 5C, 7C, 9C) and five oceanic (I, IA, IB, II, III) water types based on optical depth. Coastal waters are generally more affected by turbidity.

Due to the challenges and high costs of collecting underwater footage, available data is sparse, with source locations heavily skewed towards clearer and 'interesting' habitats. In contrast, murky environments have been largely neglected.

Synthetic data generation has been widely proposed as a solution to data scarcity \cite{li_watergan_2017, wang_uwgan_2019}, particularly in the domain of image enhancement and restoration, where access to reference images is crucial. One common approach uses the underwater image formation model (IFM) to simulate various underwater effects on clean images \cite{hou_benchmarking_2020, desai_ruig_2021, desai_rsuigm_2024, wang_uwgan_2019, ueda_underwater_2019}. This method is computationally efficient and straightforward to implement. However, it often produces visually unrealistic results, especially in turbid conditions. We believe that this approach has been misrepresented in the literature, with oversimplified or inconsistent formulations and evaluation on only a limited subset of relevant environments.

In this work, we examine the subtle differences among IFM implementations and their impact on the synthesized images. Furthermore, we propose improvements with a particular focus on better modeling and synthesizing of high-turbidity environments. The contributions of this paper can be summarized as follows:
\begin{itemize}
    \item A detailed review of prior work in synthetic data generation using the underwater IFM, highlighting important challenges that have not been adequately addressed.
    \item A dataset of induced turbidity in a controlled environment. The dataset showcases the effects of particulates in the water medium with respect to scene discoloration and information loss.
    \item An improved synthetic data generation pipeline that includes the commonly omitted \textit{forward scattering} term of the underwater IFM and considers inhomogeneities in the medium. Qualitative results show that our method produces more realistic outputs than the default IFM, particularly in high turbidity conditions.
\end{itemize}
\section{Related work}
In this paper, the related work is categorized into two main areas to highlight their distinct focuses. The first, discussed in \cref{subsec:related:uifm}, concerns the development, evaluation, and refinement of the underwater IFM over time. The second area, presented in \cref{subsec:related:synth}, centers on one of the model's common applications: the generation of synthetic data for learning-based methods.

\subsection{Underwater image formation model}\label{subsec:related:uifm}
Only a portion of the light that gets reflected from an underwater scene towards a camera reaches the lens uninterrupted. The rest can be scattered at small angles within the field of view, scattered at larger angles and diverge from the camera, or be completely absorbed due to particles of varying sizes. Additionally, these particles can scatter light towards the camera before it even reaches the scene, providing no relevant information to the captured image.

These processes are typically approximated by the underwater image formation model, also known as the Jaffe-McGlamery model \cite{jaffe_computer_1990, mcglamery_computer_1980}. In its simplified form, it is defined as:
\begin{equation}
    \mathbf{I}(\mathbf{x}) = \mathbf{D}(\mathbf{x}) + \mathbf{F}(\mathbf{x}) + \mathbf{B}(\mathbf{x})
\end{equation}
where $\mathbf{I}(\mathbf{x})$ is the degraded image, $\mathbf{D}(\mathbf{x})$ is the direct transmission, $\mathbf{F}(\mathbf{x})$ is the small-angle forward scattering, and $\mathbf{B}(\mathbf{x})$ is the backscatter. \textbf{Bold letters} indicate 2-dimensional arrays representing images.

In \cite{schechner_clear_2004}, Schechner and Karpel argued that the effects of the forward scattering term are minimal and can be ignored. By assuming that forward scattering only introduces blur, they simplified the direct and forward components to be a signal of the form:
\begin{equation}
    \mathbf{D(x)} + \mathbf{F(x)} = \mathbf{D(x)}_{blurred}
\end{equation} 
and showed that reduced contrast due to backscatter seemingly overpowers any blurring effects.

\textbf{\textit{The direct transmission}}, $\mathbf{D(x)}$, is produced through attenuation of the latent true scene radiance, $\mathbf{J(x)}$, which represents the image that would be captured in the absence of water between the scene and the camera \cite{mcglamery_computer_1980}.
\begin{equation}
    \mathbf{D}(\mathbf{x}) = \mathbf{J}(\mathbf{x}) \cdot \mathbf{t(x)} = \mathbf{J}(\mathbf{x}) \cdot  e^{-\beta_c^D \cdot \mathbf{z}(\mathbf{x})}
    \label{eq:dx}
\end{equation}

Here, $\mathbf{t}(\mathbf{x})$ denotes the transmission map, which is the proportion of scene radiance that reaches the camera lens after traveling through the water medium. It is modeled according to the Beer--Lambert law of exponential decay and is a function of the distance between the camera and the scene, $\mathbf{z}(\mathbf{x})$, as well as an attenuation coefficient $\beta_c^D$~\cite{mobley_light_1994, akkaynak_what_2017,mcglamery_computer_1980,jaffe_computer_1990}.

$\beta(\lambda)$ is the collimated beam attenuation coefficient, and is an inherent optical property (IOP) of the medium \cite{mobley_light_1994}. It is the sum of an absorption, $a(\lambda)$, and a scattering, $b(\lambda)$, coefficient, both of which are dependent on the wavelength of light. Akkaynak et al.\@ \cite{akkaynak_what_2017} showed that $\beta_c^D$ is a function of $\beta(\lambda)$, along with the camera's spectral response and scene distance, and can diverge from the medium's IOPs.

\textbf{\textit{Backscattering}} is modeled similarly:
\begin{align}
    \mathbf{B(x)} &= \int_0^z{b(\lambda) \cdot E(d, \lambda)\cdot e^{-\beta_c^B\cdot \mathbf{z(x)}}} dz \notag \\
    &= B^\infty \cdot [1 - e^{-\beta^B_c \cdot \mathbf{z(x)})}]
    \label{eq:backscatter}
\end{align}
and represents the characteristic hazy veil commonly observed in underwater imagery. $B^{\infty}$ is the value of the veil at an infinite distance
\begin{equation}
    B^\infty(\lambda) = \frac{b(\lambda)E(d,\lambda)}{\beta(\lambda)}
    \label{eq:backlight}
\end{equation}
where $E(d,\lambda)$ is the ambient light at vertical depth $d$, this time calculated through the diffuse attenuation coefficient for downwelling irradiance $\mathcal{K}_d(\lambda)$:
\begin{equation}
    E( d,\lambda) = E(0,\lambda) \cdot e^{-\mathcal{K}_d(\lambda) \cdot d}
\end{equation}
Akkaynak et al. \cite{akkaynak_revised_2018}, showed that the backscattering attenuation coefficient $\beta_c^B$ is a distinct function of $\beta(\lambda)$, and it differs from the direct transmission coefficient $\beta_c^D$, which had previously been assumed to be identical. 

\textbf{\textit{The forward scattering}} component has been modeled to have the following form \cite{mcglamery_computer_1980, jaffe_computer_1990}: 
\begin{gather}
   \mathbf{F(x)} = \left[ \left(e^{-G(\lambda) \cdot \mathbf{z(x)}} - e^{\beta(\lambda) \cdot \mathbf{z(x)}}\right) \cdot \mathbf{J(x)} \right] \ast \mathbf{H(x)}
   \label{eq:fscatter} \\
   \mathbf{H(x)}=\mathcal{F}^{-1}(e^{-\phi\cdot \mathbf{z(x)}})
\end{gather}
where $G(\lambda)$ and $\phi \geq 0$ are empirical, $\ast$ signifies convolution and $\mathcal{F}^{-1}$ is the inverse Fourier transform. Intuitively, this is a weighted and blurred version of $\mathbf{J(x)}$. 

The papers of McGlamery \cite{mcglamery_computer_1980} and Jaffe \cite{jaffe_computer_1990} disagree on the use of $\mathbf{J(x)}$ or $\mathbf{D(x)}$ for the calculation, with \cite{mcglamery_computer_1980} specifying the former should be used, as it is the true source of the scattered particles. They also disagree on the empirical limits of $G(\lambda)$, with \cite{mcglamery_computer_1980} proposing $a(\lambda)\leq G(\lambda) \leq \mathcal{K}_d(\lambda)$ and \cite{jaffe_computer_1990} proposing $|G(\lambda)|\leq \beta(\lambda)$.

When using $\mathbf{J(x)}$ in the forward scattering term, the effective attenuation coefficient of $\mathbf{D(x)} + \mathbf{F(x)}$ is $G(\lambda)$. If $G(\lambda) \approx \beta(\lambda)$, the forward scattering term vanishes. However, as $\beta(\lambda)$ grows and $G(\lambda) < \beta(\lambda)$, the forward scattering term increases, \textit{reducing the effective signal attenuation} in addition to the introduction of blur.

The analysis of Schechner and Karpel \cite{schechner_clear_2004}, along with the computational difficulty of the term and the discrepancies between \cite{mcglamery_computer_1980} and \cite{jaffe_computer_1990}, have led to the forward scattering term being overlooked in recent years. This omission persists despite the fact that the majority of scattering happens at angles less than $90^\circ$ \cite{tuchow_sensitivity_2016}, potentially within the field of view of a camera. Forward scattering may significantly impact apparent attenuation, especially in turbid environments. Therefore, in this work, we reintroduce the term and study its effects in practice.

\subsection{Synthetic data using the underwater IFM} \label{subsec:related:synth}
\label{sec:formatting}

\begin{table*}[!b]
    \centering
    \begin{tabular}{|c|c|c|c|c|c|c|}\hline
         Work         & Data type   & $\mathbf{z(x)}$ & $B^\infty$   & $d$             & $\mathbf{F(x)}$ & Availability   \\\hline
         \cite{hou_benchmarking_2020}  & Terrestrial & -               & Extracted    & No              & No              & Dataset   \\
         \cite{li_underwater_2020, anwar_deep_2018}    & Terrestrial & Kinect          & Rand. homog. & Undefined       & No              & -      \\
         \cite{ueda_underwater_2019}  & Terrestrial & Kinect          & Rand. homog. & $b(\lambda)$    & No              & -      \\
         \cite{desai_ruig_2021} & Terrestrial & Kinect          & Analytical   & Undefined       & No              & -      \\  
         \cite{desai_rsuigm_2024} & Both        & Undefined       & Analytical   & $\mathcal{K}_d$ & No              & Dataset   \\
         \cite{liu_model-based_2022}   & Underwater  & megaDepth \cite{li_megadepth_2018}       & Analytical   & Incident        & No              & -      \\     
         Ours         & Underwater  & DepthAnythingv2\cite{yang_depth_2024} & Analytical   & $\mathcal{K}_d$ & Yes             & Code     \\
\hline         
    \end{tabular}
    \caption{Existing variations of synthetic data generation pipelines based on the underwater IFM. The column \textit{Availability} corresponds to either the synthesized dataset or code availability, that would allow for the full range of outputs to be accessed.}
    \label{tab:related}
\end{table*}

The underwater IFM has been widely used to synthesize data, using source images from both terrestrial and underwater environments. Although transforming terrestrial images into underwater scenes and modifying the attenuation of already underwater images are technically distinct tasks, their primary differences lie in the illumination profile, lens distortions, and content of the source images. However, these aspects are not addressed by the IFM, which solely models light attenuation in the medium between the scene and the camera. Therefore, within the scope of the IFM, there is little justification for treating the two tasks as fundamentally different.

Hou et al.\@ \cite{hou_benchmarking_2020} use a simplified version of the IFM that resolves the need for distance information. They propose calculating the complete transmission map of \textit{real} underwater images using the red channel prior \cite{galdran_automatic_2015} and applying it to terrestrial images. However, a transmission map should transfer shape information in the form of artifacts on the synthetically degraded images, an effect that does not appear in the resulting SUID dataset. It is thus assumed, that transmission coefficients were calculated instead. The synthesized images exhibit the characteristic blue-green tint typical of underwater footage, but lack horizontal depth dependence and the gradual fading observed over large distances. 

The authors in \@ \cite{li_underwater_2020, anwar_deep_2018, ueda_underwater_2019} apply the underwater IFM on the NYU-depth V2 dataset \cite{silberman_indoor_2012}, which provides depth maps. In all works, $B^\infty$ is selected randomly within a range, effectively decoupling its color from the attenuation coefficients. This allows for greater flexibility and variability, though the resulting color can be uncharacteristic of underwater environments. In \cite{li_underwater_2020, anwar_deep_2018}, the attenuation coefficient is simplified to the medium's inherent value. In contrast, \cite{ueda_underwater_2019} computes it across the full wavelength spectrum, taking into account all other dependencies—though it does not distinguish between the direct and backscattering terms and uses $\beta_c^D$ for both. All studies consider the effect of vertical depth; however, \cite{li_underwater_2020, anwar_deep_2018} do not clearly explain how depth is incorporated, and \cite{ueda_underwater_2019} uses the beam attenuation coefficient in place of the diffuse attenuation coefficient.

Desai et al.\@ \cite{desai_ruig_2021}, introduce the distinction between $\beta_c^D$ and $\beta_c^B$ as proposed by \cite{akkaynak_revised_2018}. Working on the same NYU-depth V2 dataset \cite{silberman_indoor_2012}, the resulting images appear to darken with changes in water type rather than showing increased turbidity, and lack the typical color distortions characteristic of underwater environments. It is unclear whether this visual deviation is due to the introduction of $\beta_c^B$, or to another aspect of their implementation.

Desai et al.\@ \cite{desai_rsuigm_2024} approach varying depth simulation through a downwelling term in the direct transmission, and analytical $B^\infty$ calculation using $\mathcal{K}_d$. They synthesize the RSUIGM\footnote{An underwater version of RSUIGM has been published by the authors in the context of a workshop challenge\cite{kiefer_3rd_nodate}.} dataset with source images from SUID \cite{hou_benchmarking_2020}, but do not clarify how distance information is obtained.

Utilizing the scene structure, Liu et al.\@ \cite{liu_model-based_2022}  simulate uniform, parallel, and artificial light conditions. They apply their approach to the EUVP \cite{islam_fast_2020} dataset, with depth maps extracted using megaDepth \cite{li_megadepth_2018} and random rescaling.

All aforementioned works have used the synthesized images to develop or evaluate image restoration frameworks. Their key implementation differences are summarized in \cref{tab:related}.

\subsection*{A critical look}
In much of the prior work, examples shown for coastal water types exhibit minimal visible attenuation, likely due to significantly underestimated horizontal distances. When accurate distances are available, the synthesized images often appear increasingly artificial as attenuation coefficients rise. At the same time, evaluations tend to focus on heavily discolored but low-scattering scenes, typically overlooking the quality of more turbid data.

This trend extends to work focused directly on the IFM. Measurements and studies of physical processes, such as forward scattering and its effects, are typically based on oceanic water types. However, due to the exponential nature of attenuation, deviations from physical realism are less noticeable in low-turbidity environments, making it easier to overlook model inaccuracies.

Although prior work has claimed to cover the full range of Jerlov water types, it remains unclear whether these approaches can reliably generalize to scattering-dominated, turbid environments. Therefore, in this work, we specifically shift our focus to that regime.
\section{Methods}
\label{sec:methods}
Our proposed method involves the inclusion of the forward-scattering term in the underwater IFM along with coefficient adjustments. Additionally, we introduce a stochastic approach to medium inhomogeneity, allowing for synthetic data with higher variation. 

\begin{figure}[t]
    \centering
    \includegraphics[width=\linewidth]{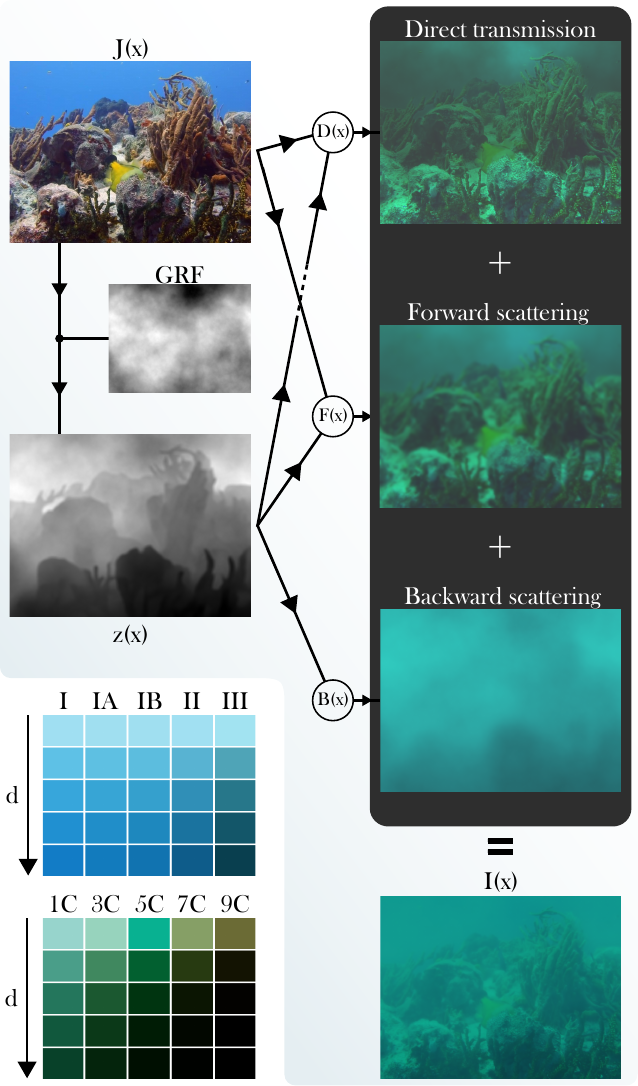}
    \caption{The proposed synthetic data generation pipeline. The inputs are a clear underwater image, along with a depth map multiplied with a Gaussian Random Field. The direct, forward, and backscattering components are calculated using the desired water type coefficients through their respective functions, \cref{eq:dx,eq:backscatter,eq:fscatter}. Their sum is the degraded image $\textbf{I(x)}$. The white-patch behavior for all Jerlov water types \cite{solonenko_inherent_2015} is displayed on the bottom left. *\textit{The direct and forward scattering images have had their contrast and brightness enhanced for visualization purposes.}}
    \label{fig:pipeline}
\end{figure}

Since code of prior work is not available, we define and implement the following reference model, on which we apply our modifications: The \textit{revised IFM} with $\beta_C^D$, $\beta_C^B$ and $\mathbf{B(x)}$ calculated for the entire range of wavelengths, as proposed by the authors in \cite{akkaynak_what_2017,akkaynak_revised_2018}, with the inclusion of downwelling attenuation, similar to \cite{desai_rsuigm_2024}. The full pipeline is shown in \cref{fig:pipeline}

\subsection{Small angle forward scattering}

We reintroduce the forward scattering term of \cref{eq:fscatter} into the underwater IFM, and parametrize $G_(\lambda)$ as:
\begin{equation}
    G(\lambda) = a(\lambda) + g\cdot b(\lambda)
\end{equation} 
Here, $a(\lambda)$ and $b(\lambda)$ are the absorption and scattering coefficients described in \cref{subsec:related:uifm} and $g\leq1$ represents the portion of light scattered away from the line of sight (LOS) between the scene and the camera. This parametrization is consistent with recent studies in water optical properties \cite{tuchow_sensitivity_2016, doxaran_improved_2016}. 

Similarly, we change the effective attenuation coefficient of the backscattering term to match $G(\lambda)$. This choice is important for consistency, i.e., if we assume some of the scattered rays from the scene reach the lens, the same must be true of backscattered rays.
\begin{equation}
    \textbf{B}(\textbf{x}) = B^\infty \cdot [1 - e^{-G^B_c \cdot \textbf{z}(\textbf{x})}]
\end{equation}

Finally, we reduce the scattering coefficient in the backlight calculation (\cref{eq:backlight}) through constant $\mu\leq1$, which is not necessarily the same as $g$. This reflects the assumption that not all light scattered by particles is initially directed toward the camera, even before accounting for attenuation along the LOS. The final backlight calculation is performed as follows:
\begin{equation}
    B^\infty(\lambda) = \frac{\mu \cdot b(\lambda) \cdot E(d,\lambda)}{G(\lambda)}
    \label{eq:backlightnew}
\end{equation}
Notice the use of $G(\lambda)$ which results from the integration in \cref{eq:backscatter}.
The convolution in \cref{eq:fscatter} is performed directly in the image domain, with a depth-variable Gaussian kernel and $\sigma=\phi \cdot \mathbf{z(x)}$.

\subsection{Medium non-uniformity}
A core assumption of the underwater IFM model is that the medium is uniform. Unlike marine snow, which consists of relatively large, visible particles, the non-uniformity referred to here arises from subtle, indistinguishable variations in the water that cause local fluctuations in absorption and scattering coefficients. These inhomogeneities cannot be analytically modeled or reliably approximated, as they are the result of chaotic fluid dynamics in unbounded volumes without clearly defined sources or initial conditions. Nonetheless, incorporating such randomness and variability can have a significant impact on the realism and effectiveness of synthetic data generation.

Gaussian random fields (GRFs) have been used to describe natural physical processes, such as the deviations in homogeneity of the Cosmic Microwave Background \cite{wandelt_gaussian_2013}, and have also been applied in generating synthetic image data \cite{li_multimodal_2025}. Specifically, scale-free Gaussian Random Fields have a fractal structure with spatial correlations on different scales, and visually resemble smoke effects. A GRF is generated for every image and its values are scaled to small variations around 1. It is then multiplied by the depth map, which effectively corresponds to density variations.

It is important to note that this process is not proposed as an extension of the underwater IFM, but rather as a general-purpose degradation technique.
\section{Experiments and results}
\label{sec:data_collection}

\subsection{The BUCKET dataset}

A reappearing problem in synthetic image generation is the lack of ground truth. Without a ground truth, it is not trivial to evaluate the quality of the result. Motivated by this, we propose an image dataset, the \textit{ Baseline for Underwater Conditions in Known Environments and Turbidity (BUCKET)}.

\begin{figure}[t]
    \centering
    \begin{subfigure}{0.45\linewidth}
        \centering
        \includegraphics[width=\linewidth]{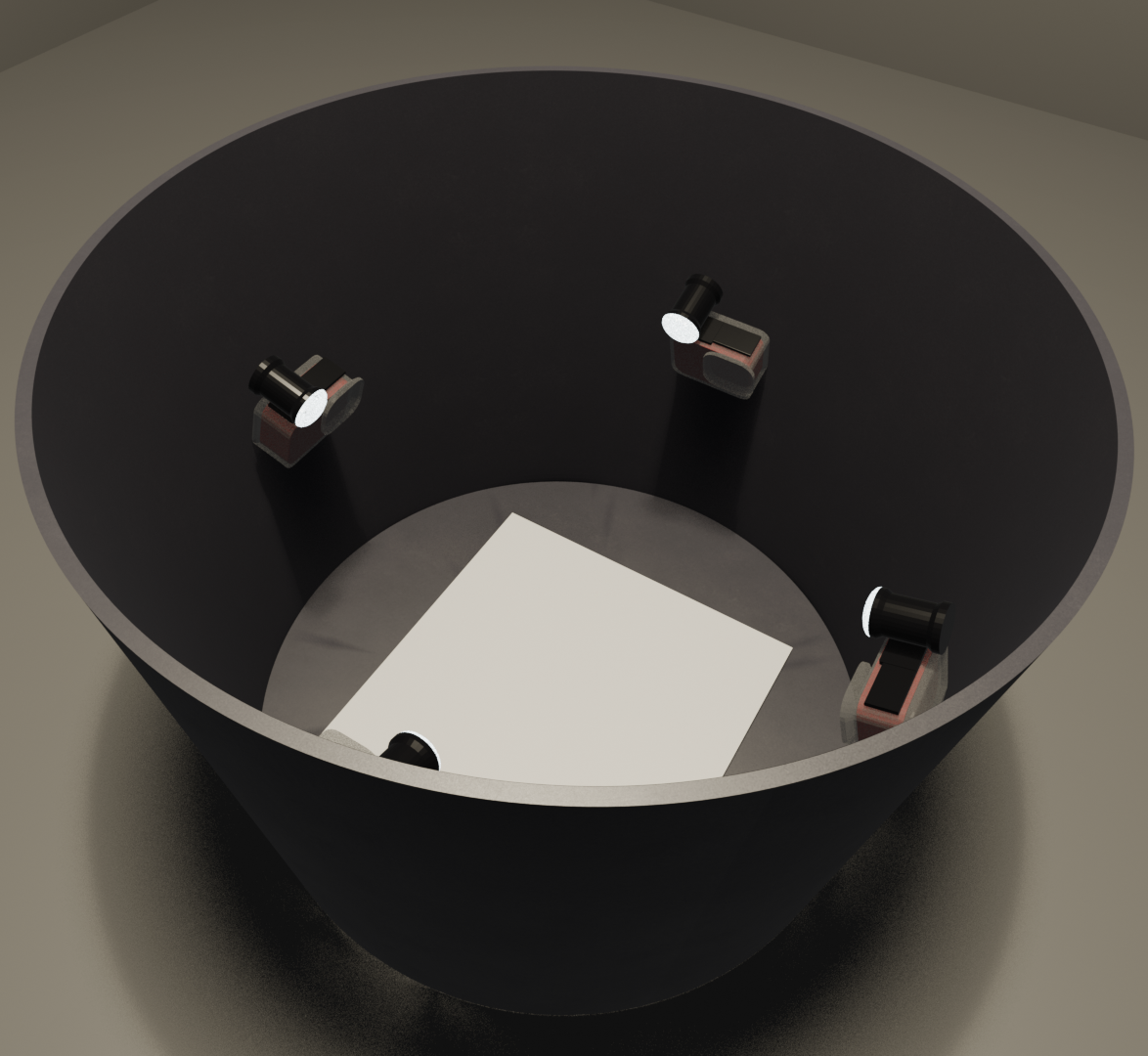}
        \label{}
        \caption{}
    \end{subfigure}
    \begin{subfigure}{0.49\linewidth}
        \includegraphics[width=0.49\linewidth]{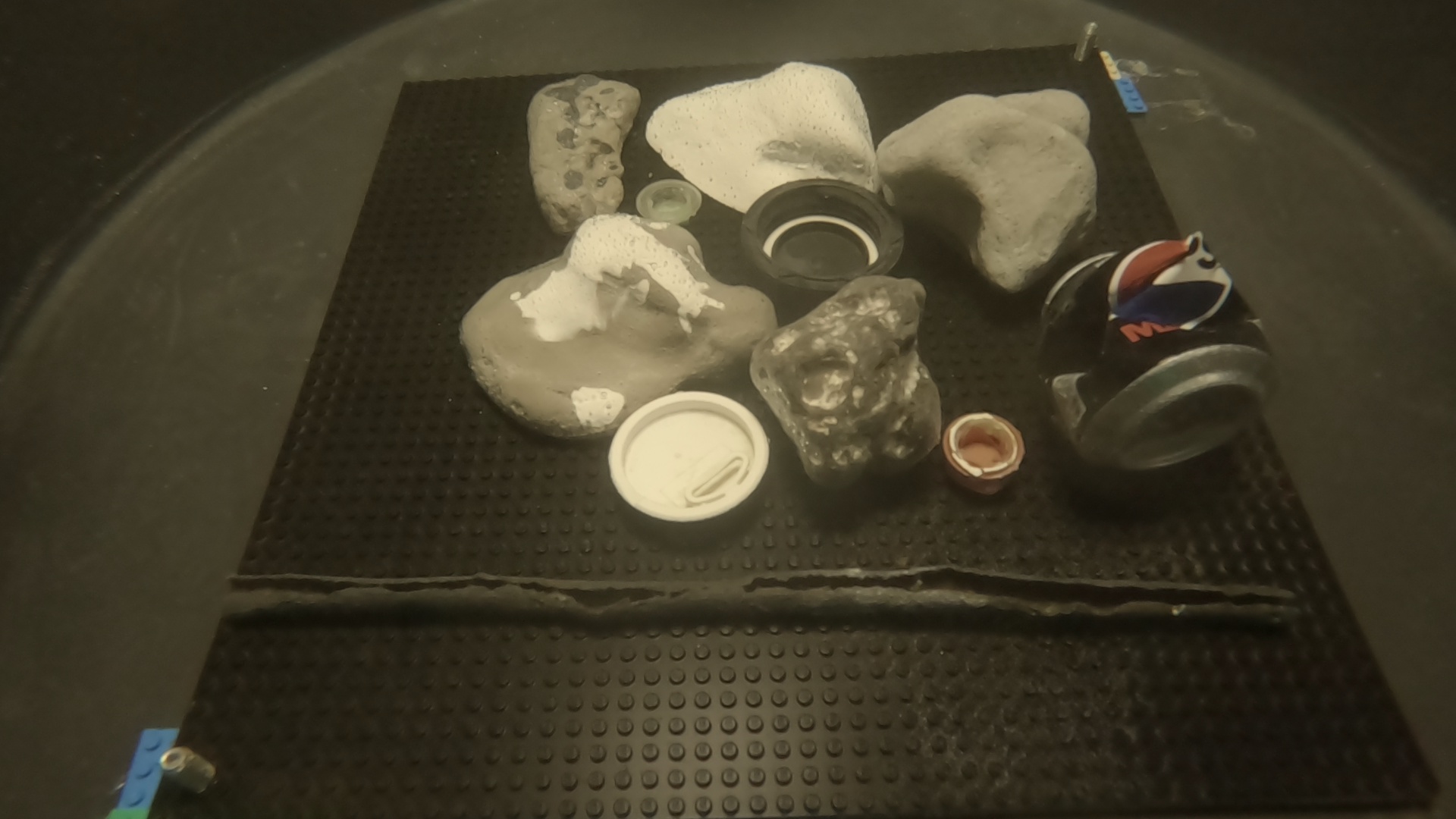}
        \includegraphics[width=0.49\linewidth]{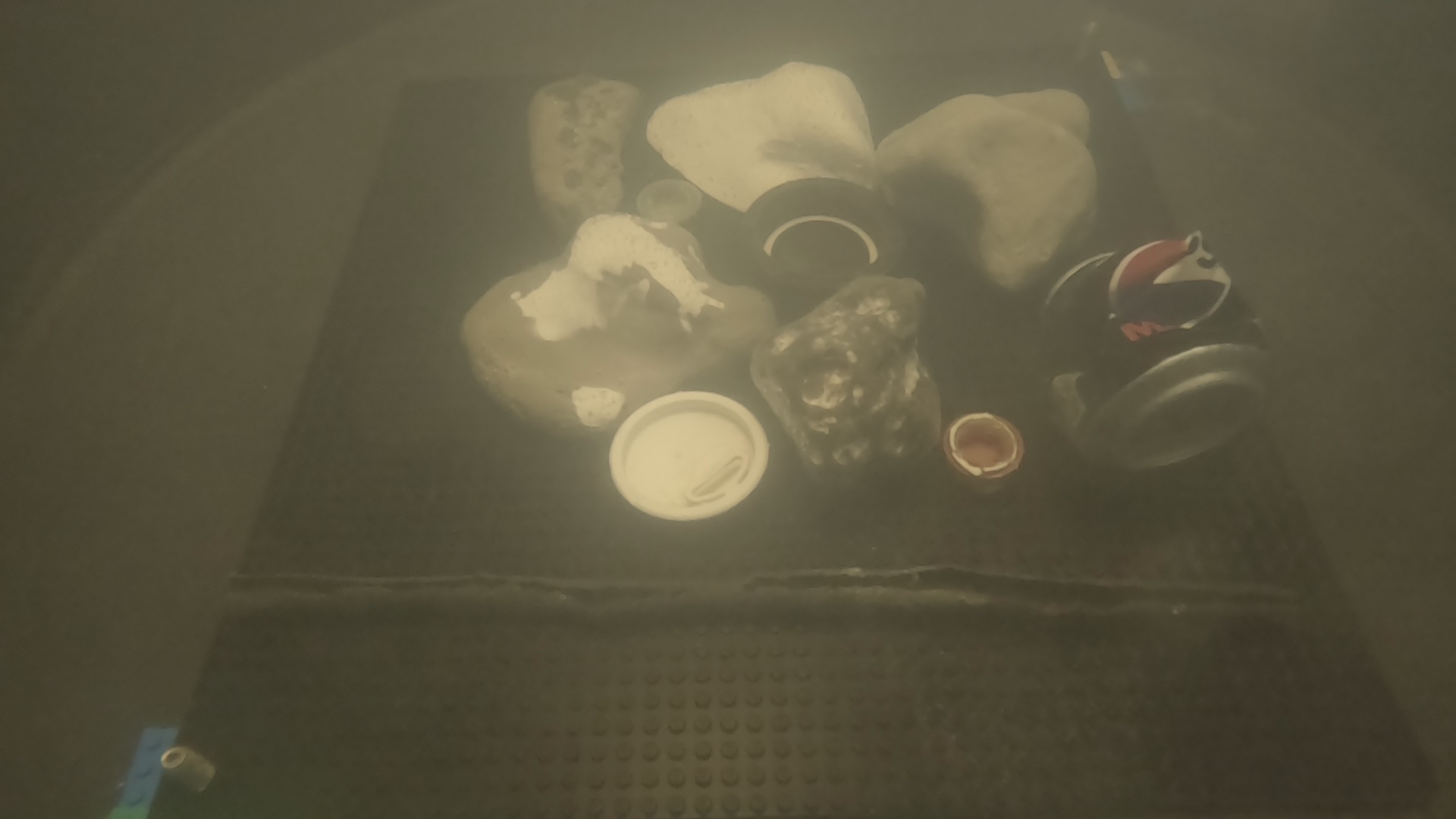}
        \includegraphics[width=0.49\linewidth]{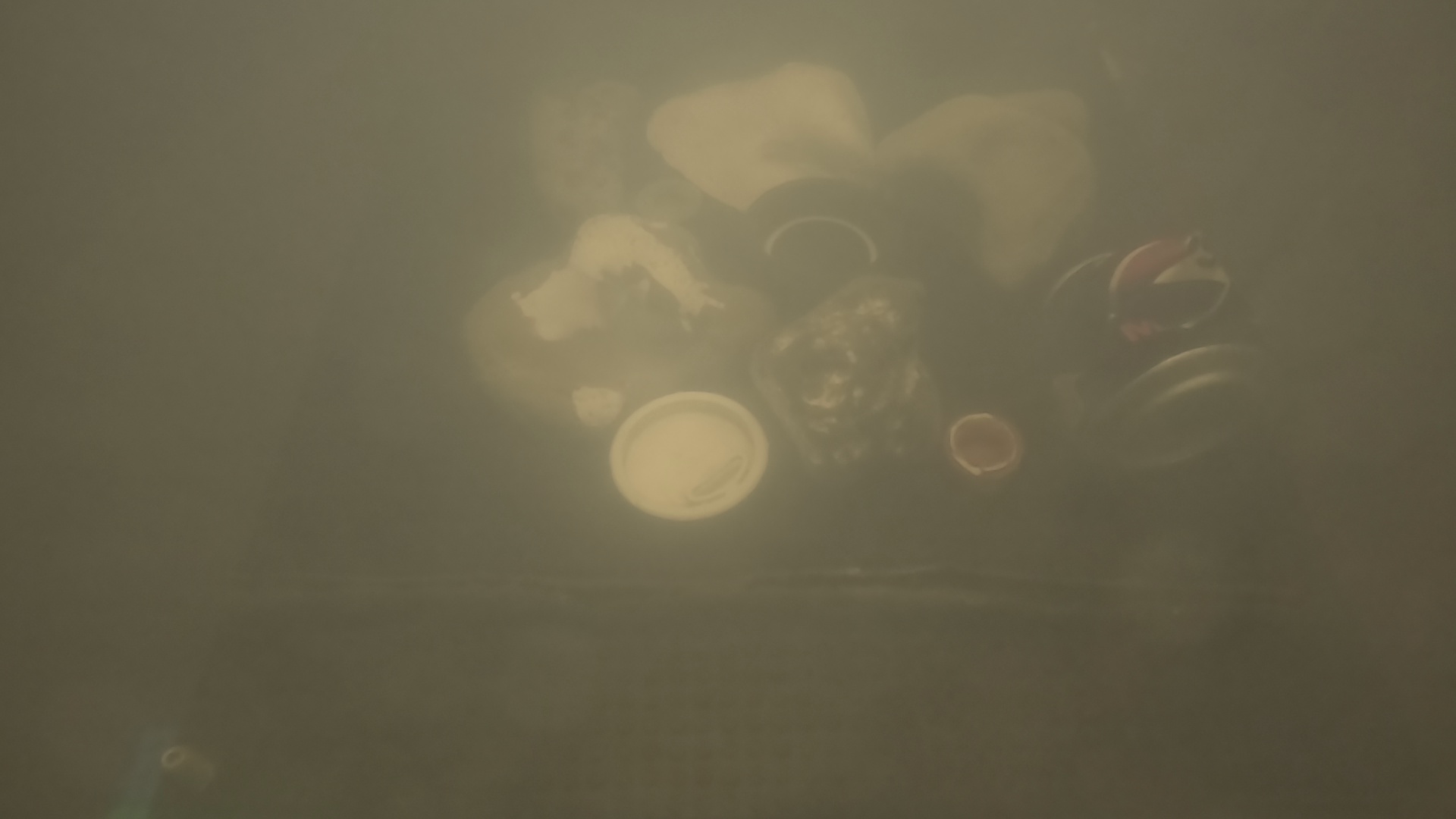}
        \includegraphics[width=0.49\linewidth]{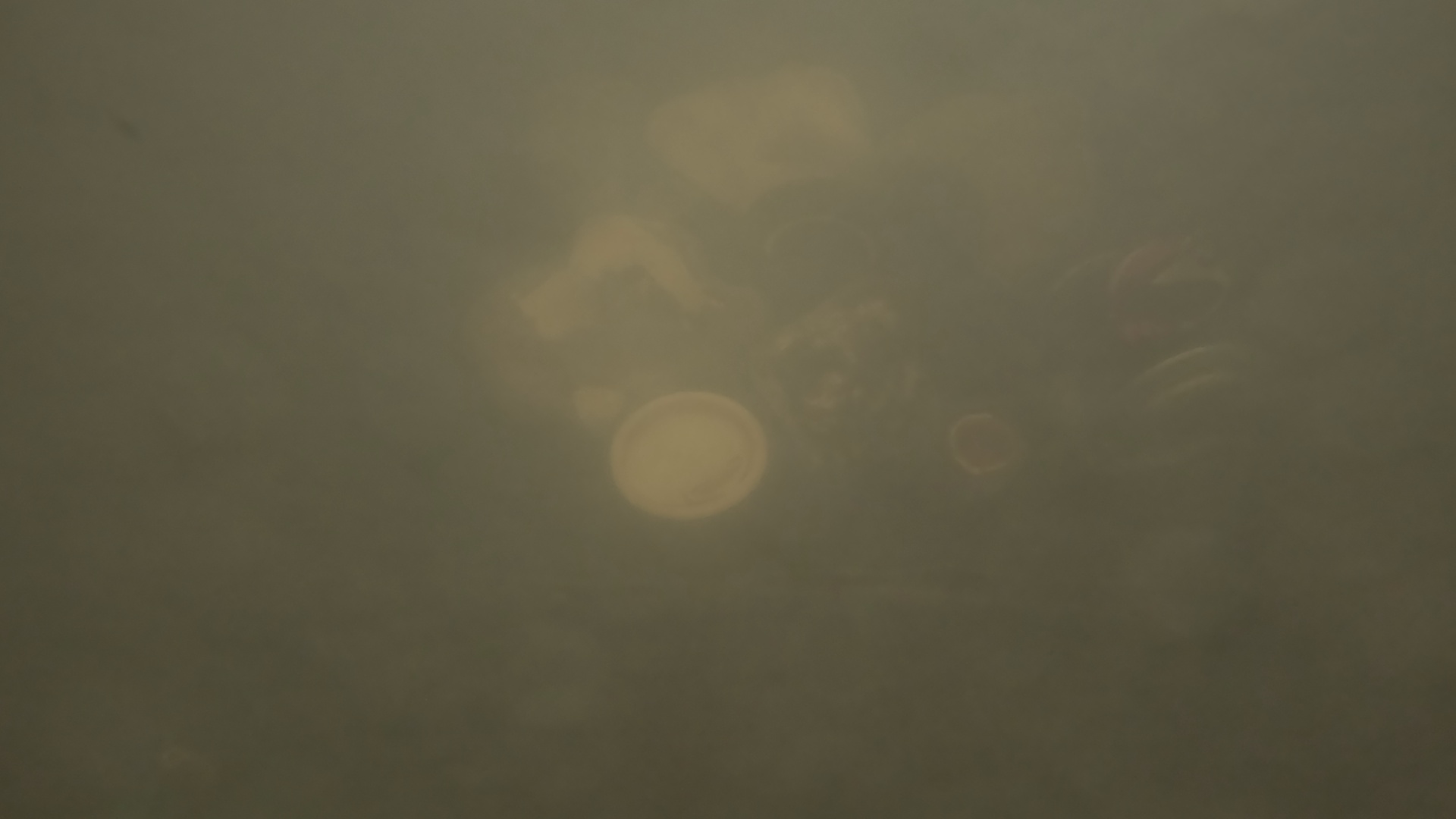}
        \includegraphics[width=0.49\linewidth]{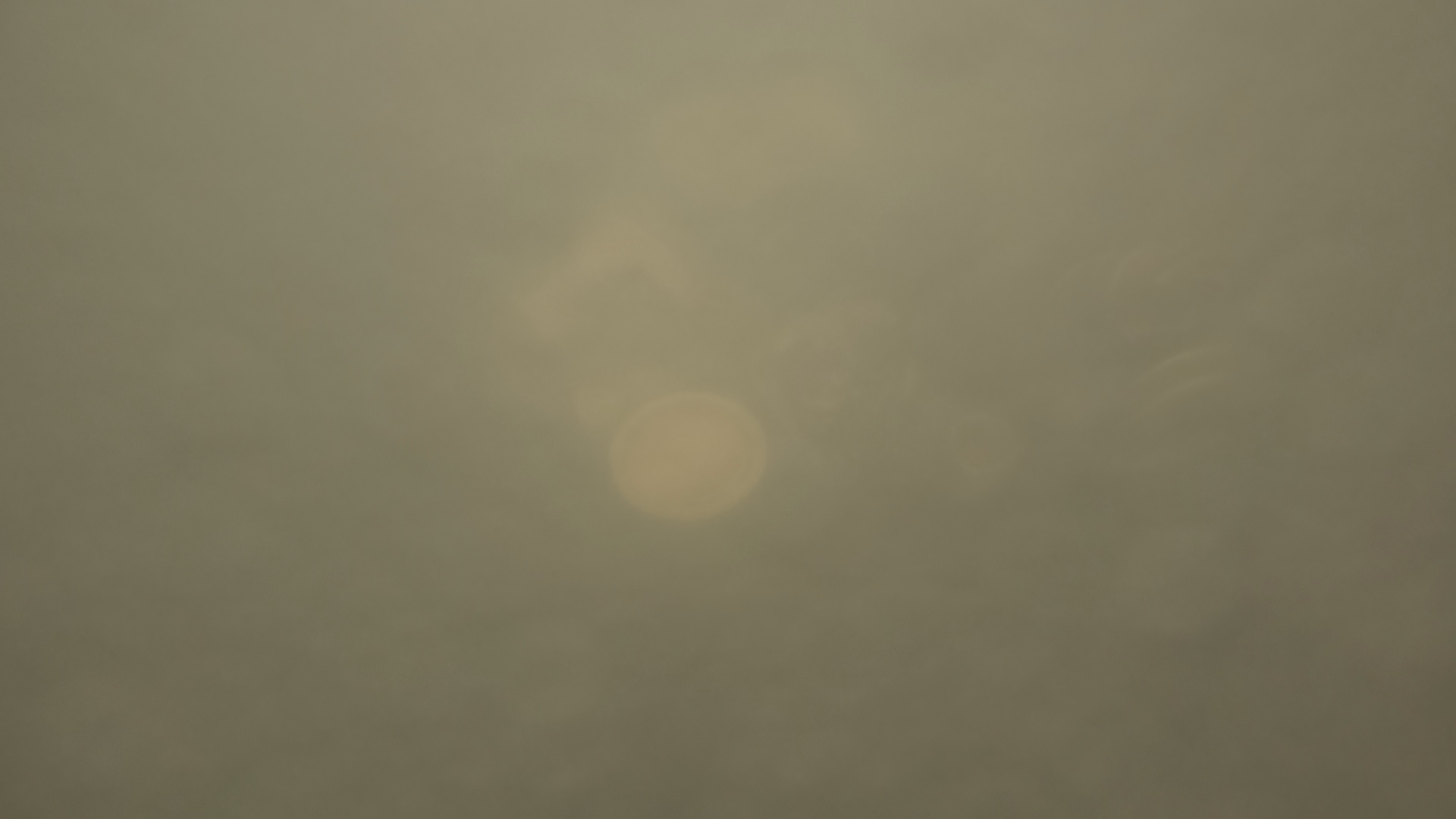}
        \includegraphics[width=0.49\linewidth]{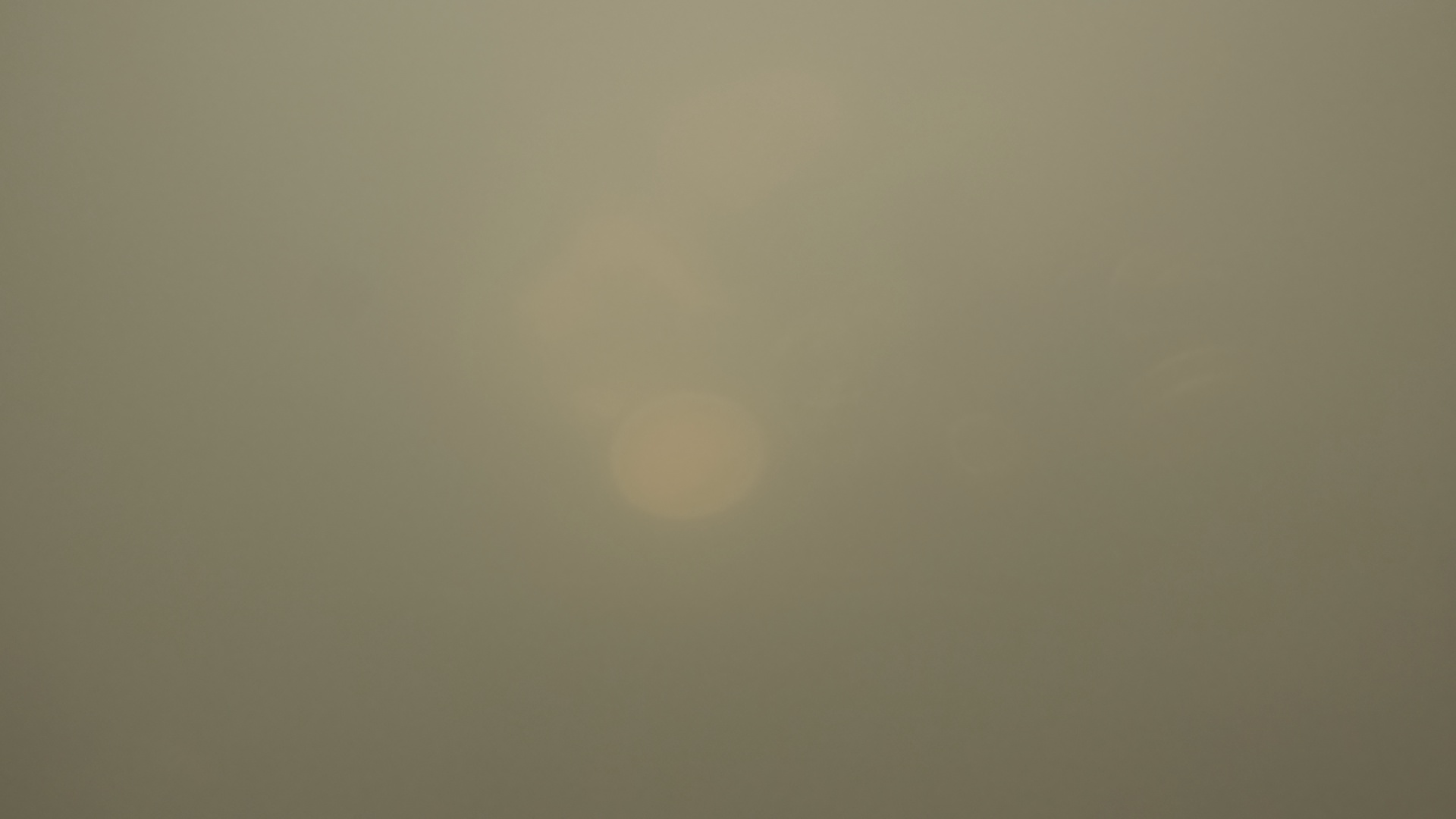}
        \label{}
        \caption{}
    \end{subfigure}
    \caption{The BUCKET dataset. a) The data collection setup. b) Some example images at increasing turbidity.}
    \label{fig:data:setup}
\end{figure}

The data collection setup involves four \textit{GoPro} cameras, equally spaced along the perimeter of a cylindrical tank with opaque walls. All data collection took place in a dark room, and lighting was introduced through four \textit{Blue Robotics} \textit{Lumen} subsea lights and a diffused lamp centered above the tank. Turbidity was simulated using increasing amounts of oat milk and clay, and was accurately measured. On the bottom of the tank, sets of objects were placed on a \textit{LEGO} baseplate, retaining object positions. This ensures the preservation of the scene between the clear reference images and the ones collected as turbidity increases. Objects include, but are not limited to, artificial and real rocks, as well as trash of different shapes and colors. A visualization of the setup, along with sample images, is presented in \cref{fig:data:setup}.

The setup and collection protocol were designed to generate a diverse range of conditions, enabling the resulting images to support multiple underwater vision tasks, such as image restoration, object detection, and multi-view reconstruction. A total of 64 images, captured only under ambient light from the overhead lamp, comprise a preliminary version of the dataset and are considered in this work. They will be made publicly available through the project page.

\subsection{Comparison with reference images}
To obtain usable depth information for the collected footage, we employed the state-of-the-art monocular depth estimation model, Depth-Anything-v2 \cite{yang_depth_2024}, and generated depth maps from the clean images. The predicted relative depths were then scaled using the known setup dimensions.

Since the induced turbidity does not correspond precisely to any standard Jerlov water type, coefficients were approximated to visually match the ground truth images: $\beta_c^D = \beta_c^B = \beta_c^{approx}$. As such, this is a simplified version of the reference model defined in \cref{sec:methods}. Detailed values are provided in the supplementary material.

\cref{fig:res:lab} shows a comparison of real and synthetic images. The turbidity measurements for the real underwater images are 1.2, 6.53, 16.07, 20.20, and 30.43 NTU, respectively, and the added substance was oatmilk. All synthesized images share a horizontal depth range of 0.3-0.7m and a vertical depth of 0.5m.

\begin{figure*}[b]
    \centering
    \begin{subfigure}{0.95\textwidth}
    \centering
        \includegraphics[width=0.19\linewidth]{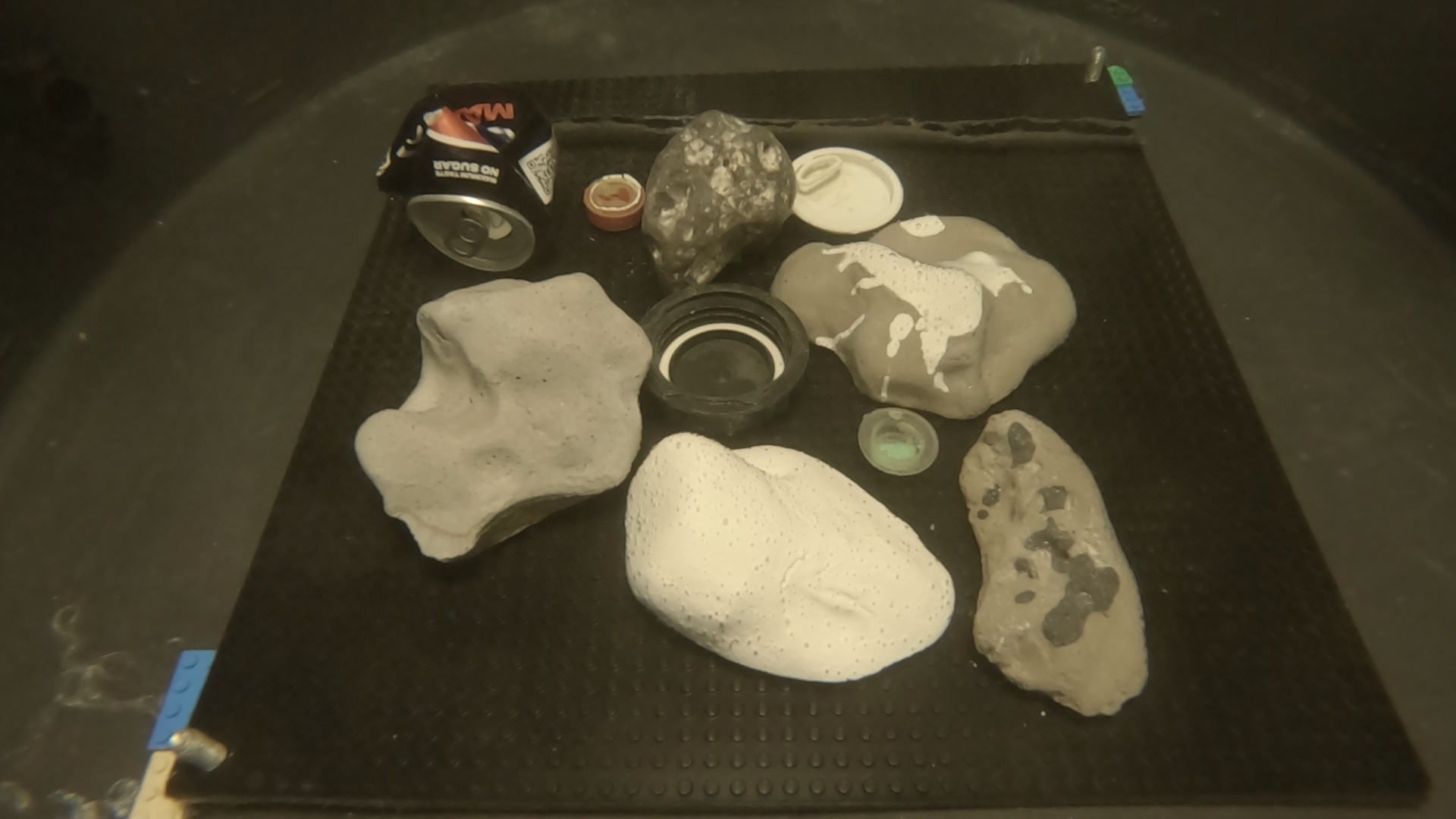}
        \includegraphics[width=0.19\linewidth]{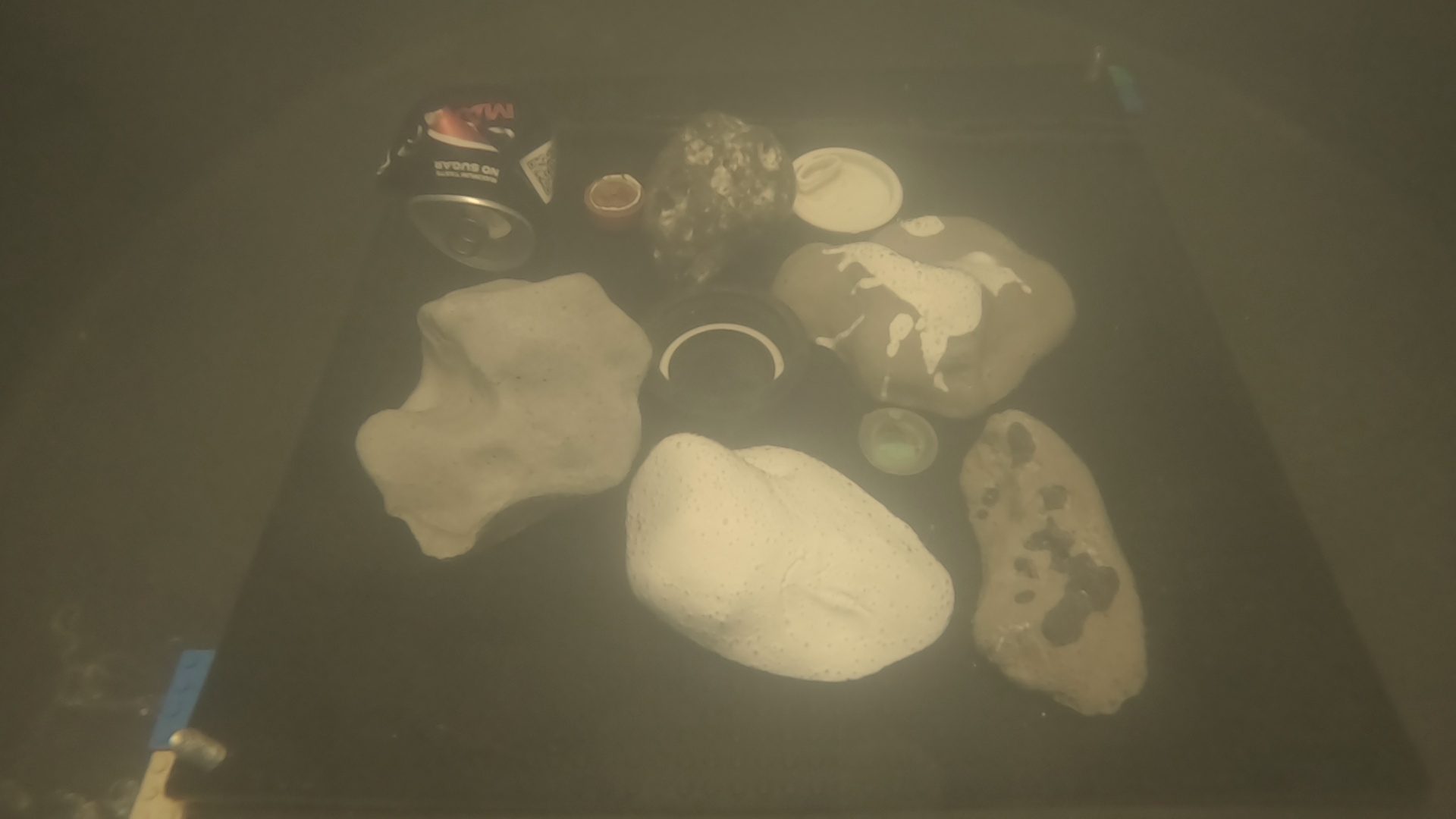}
        \includegraphics[width=0.19\linewidth]{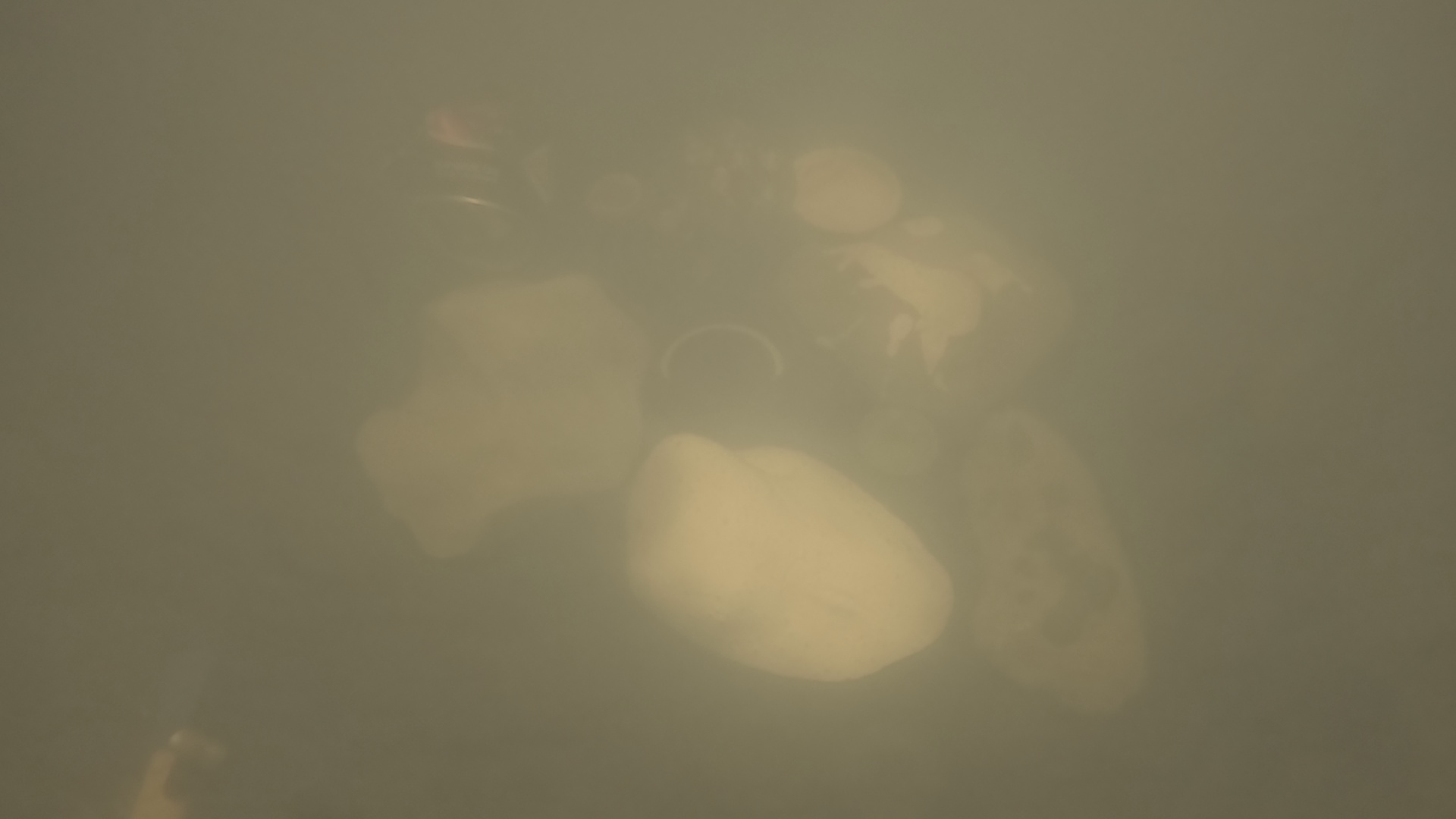}
        \includegraphics[width=0.19\linewidth]{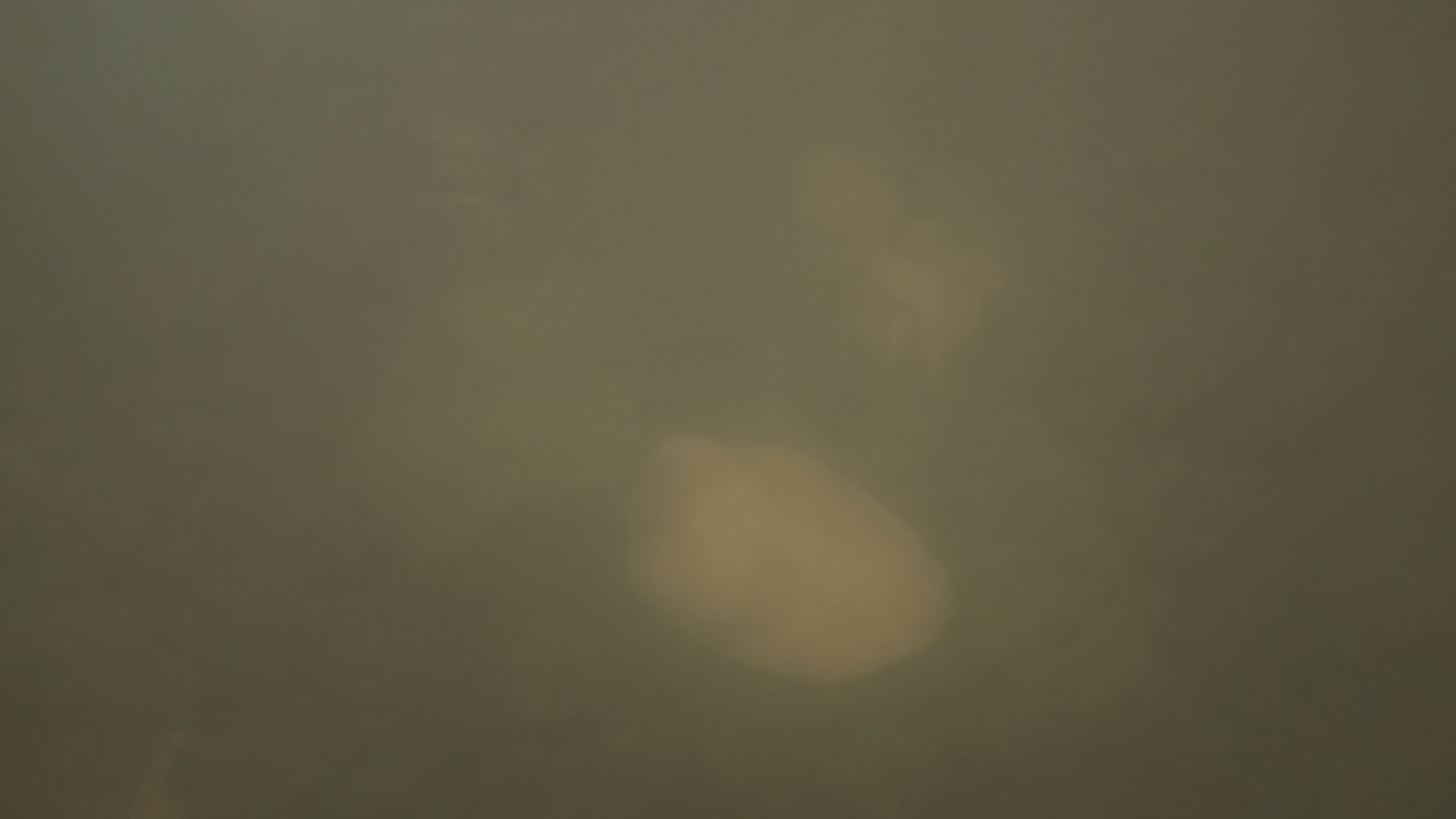}
        \includegraphics[width=0.19\linewidth]{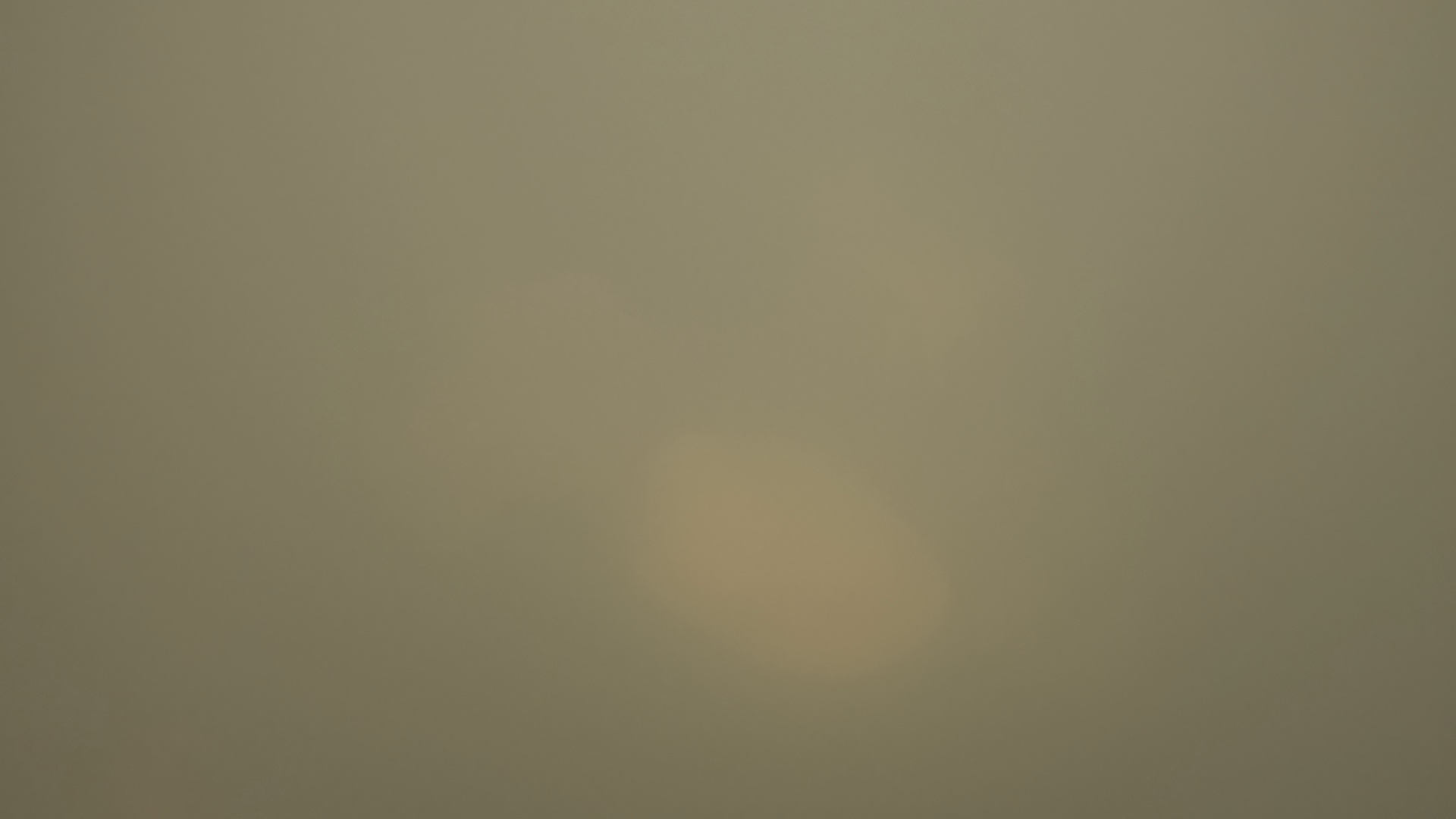}
        \caption{}
        \label{fig:res:lab:real}
    \end{subfigure}

    \begin{subfigure}{0.95\textwidth}
    \centering
        \includegraphics[width=0.19\linewidth]{figures/results/camera_3_new/camera_3_clean/camera_3_clean.jpg}
        \includegraphics[width=0.19\linewidth]{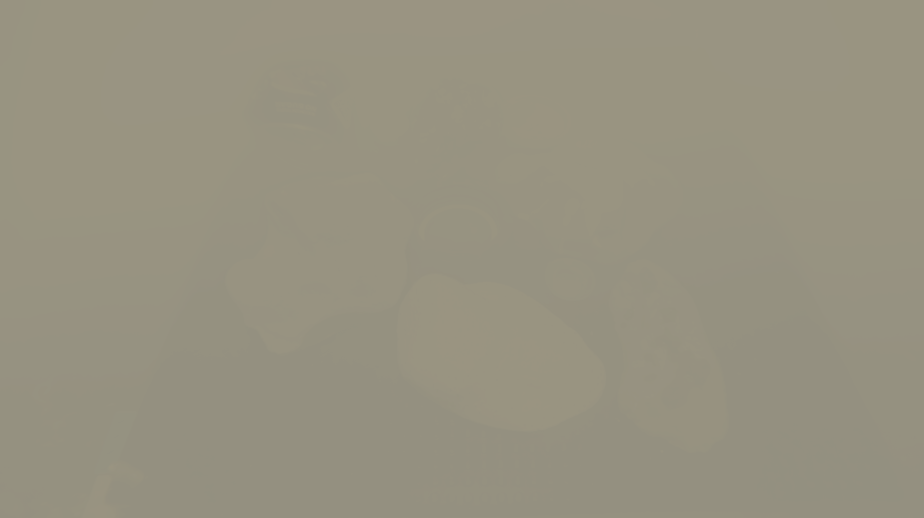}
        \includegraphics[width=0.19\linewidth]{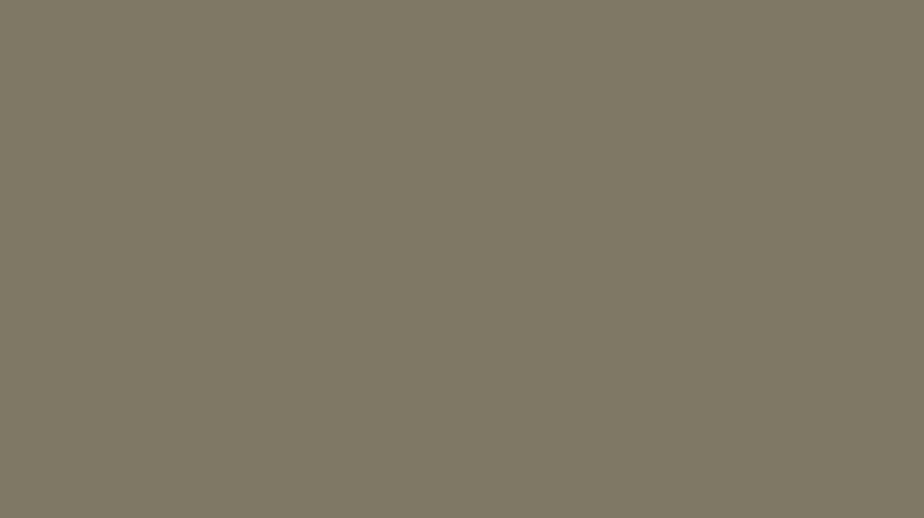}
        \includegraphics[width=0.19\linewidth]{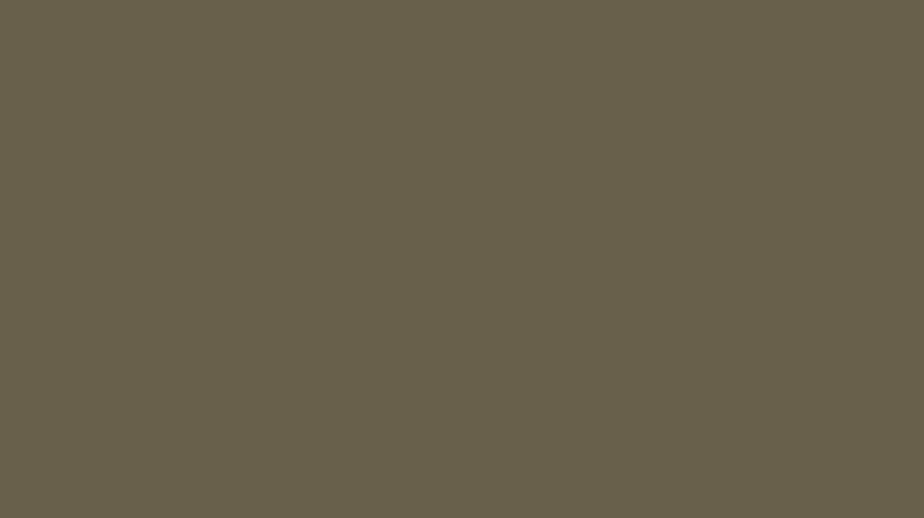}
        \includegraphics[width=0.19\linewidth]{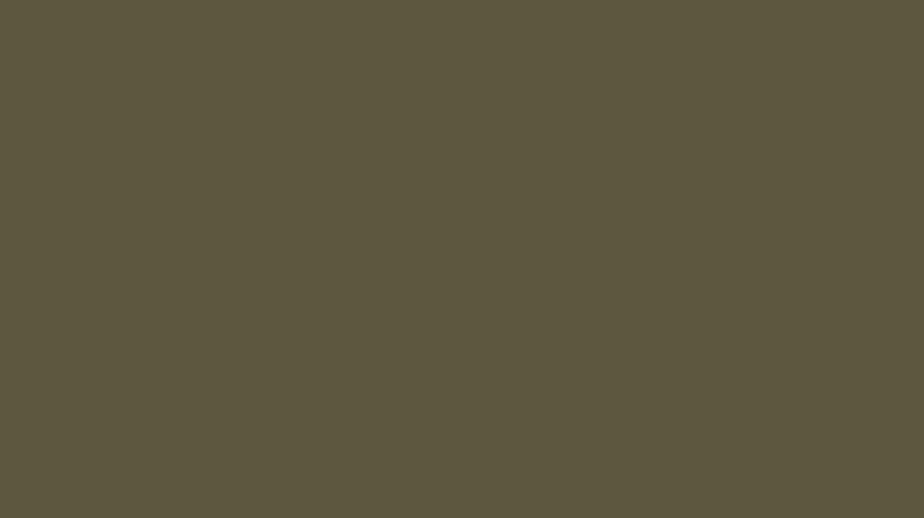}
        \caption{}
        \label{fig:res:lab:turb_old_full}
    \end{subfigure}

    \begin{subfigure}{0.95\textwidth}
        \centering
        \includegraphics[width=0.19\linewidth]{figures/results/camera_3_new/camera_3_clean/camera_3_clean.jpg}
        \includegraphics[width=0.19\linewidth]{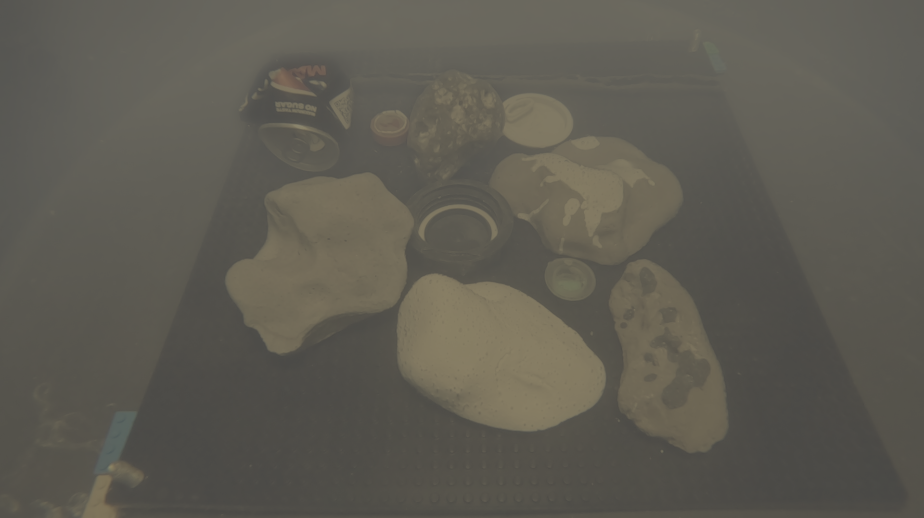}
        \includegraphics[width=0.19\linewidth]{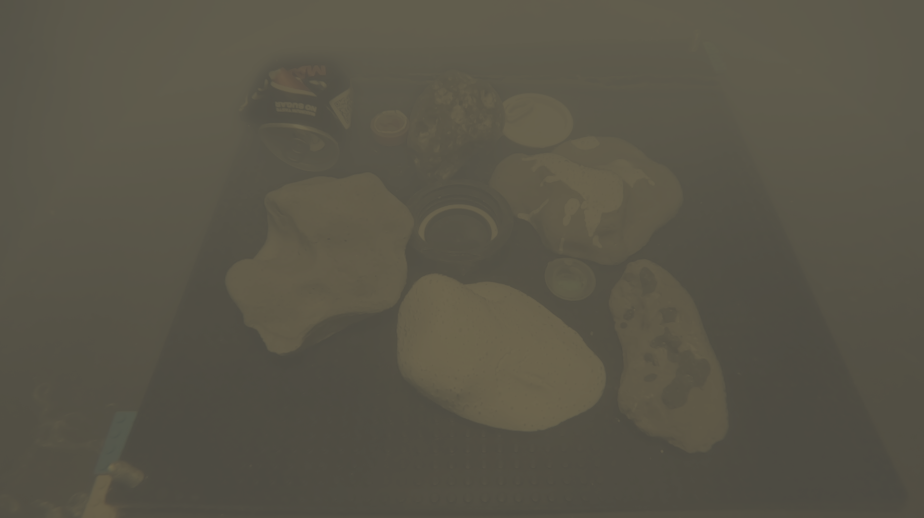}
        \includegraphics[width=0.19\linewidth]{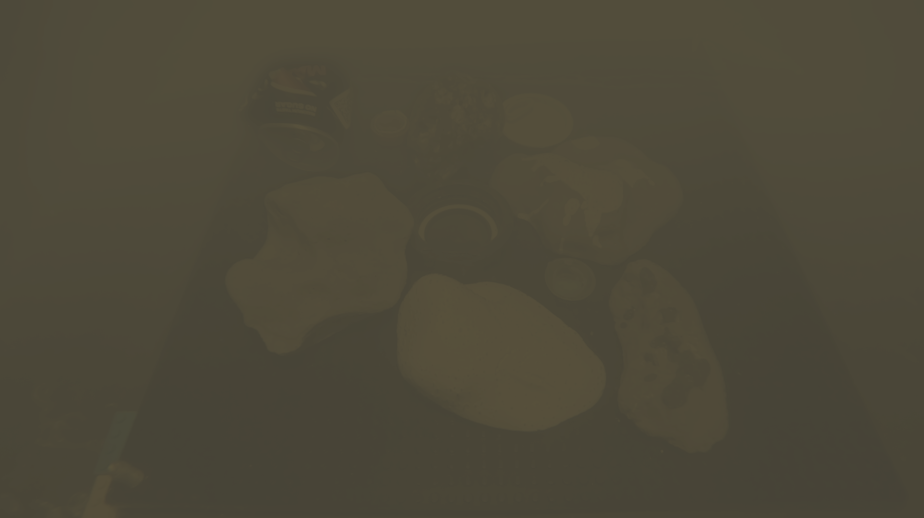}
        \includegraphics[width=0.19\linewidth]{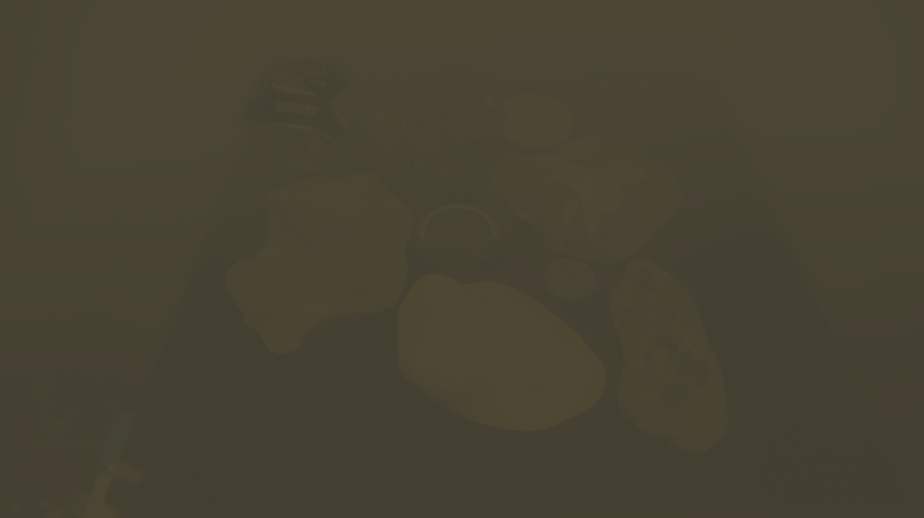}
        \caption{}
    \end{subfigure}
     \begin{subfigure}{0.95\textwidth}
        \centering
        \includegraphics[width=0.19\linewidth]{figures/results/camera_3_new/camera_3_clean/camera_3_clean.jpg}
        \includegraphics[width=0.19\linewidth]{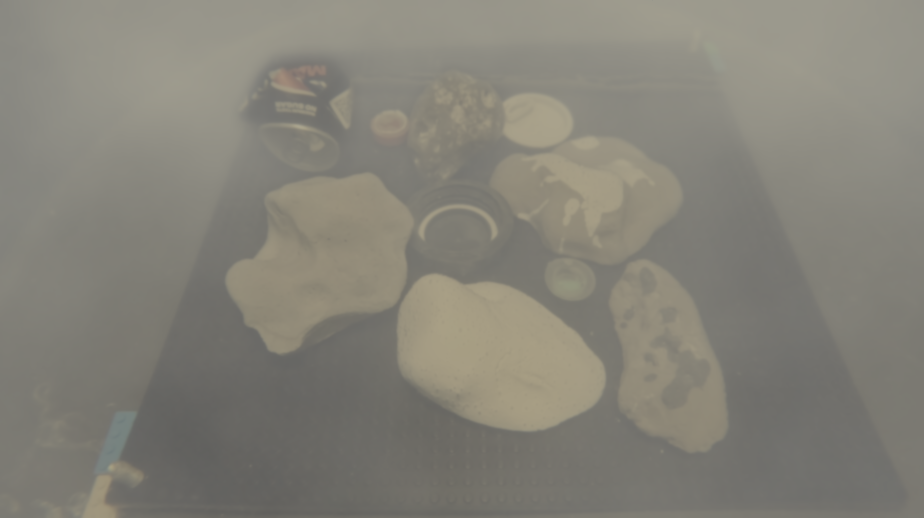}
        \includegraphics[width=0.19\linewidth]{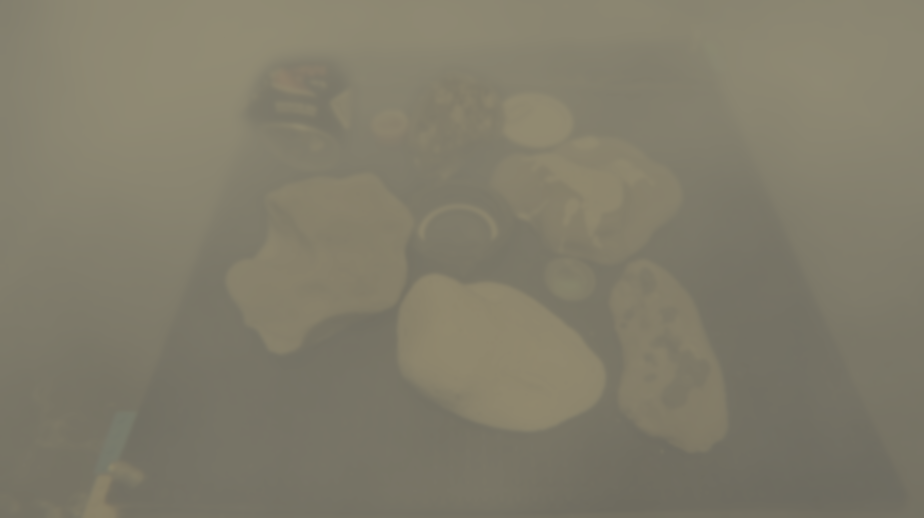}
        \includegraphics[width=0.19\linewidth]{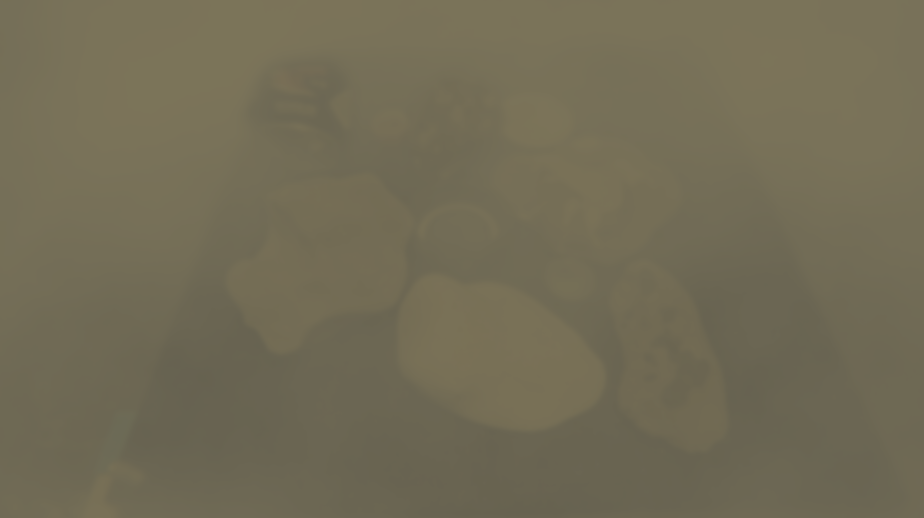}
        \includegraphics[width=0.19\linewidth]{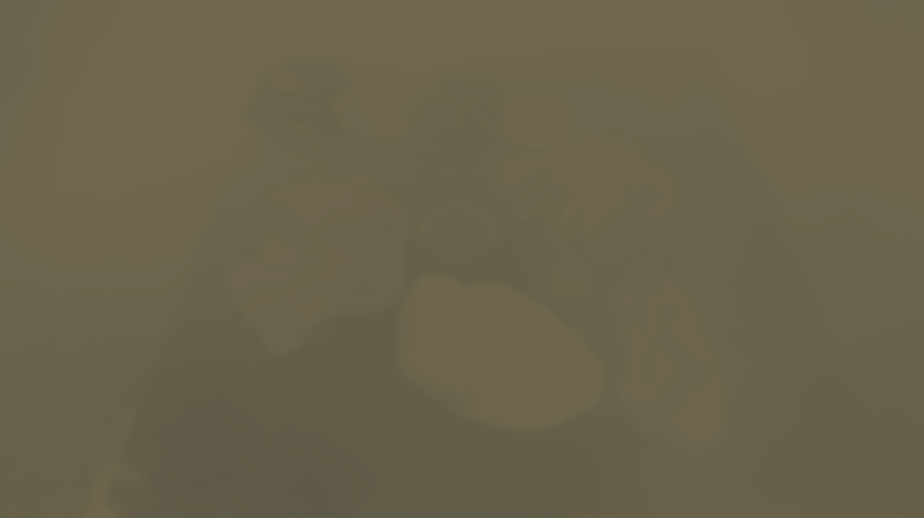}
        \caption{}
    \end{subfigure}
    \caption{Comparison of the underwater IFM with real data. Columns indicate increasing turbidity. a) Real underwater data. b) Synthetic data with the reference model. c) Synthetic data with the reference model but $b_c^{eff} = g \cdot b_c^{approx}$. d) Synthetic data with the proposed model. Rows \textit{c} and \textit{d} share the same effective attenuation coefficients. However, \textit{d} includes blur, inhomogeneity effects, and a backlight adjusted through $\mu$. Our proposed model is not equivalent to only reducing the scattering coefficient.}
    \label{fig:res:lab}
\end{figure*}

\begin{figure*}[p]
    \centering
    \begin{subfigure}{\textwidth}
        \begin{subfigure}{\textwidth}
        \centering
            \includegraphics[width=0.19\linewidth]{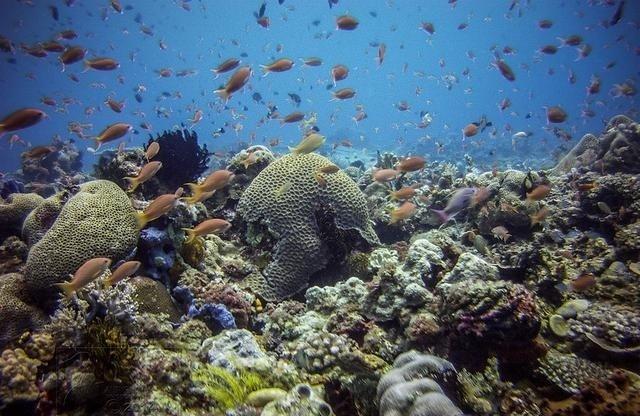}
            \includegraphics[width=0.19\linewidth]{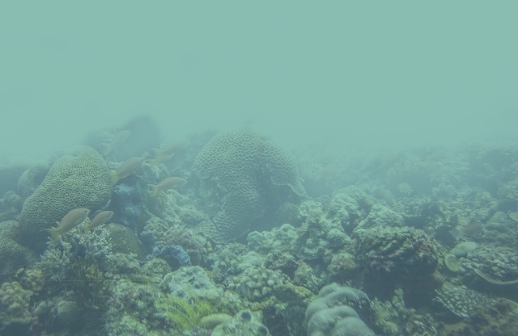}
            \includegraphics[width=0.19\linewidth]{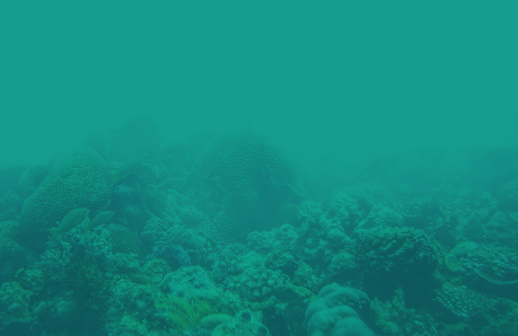}
            \includegraphics[width=0.19\linewidth]{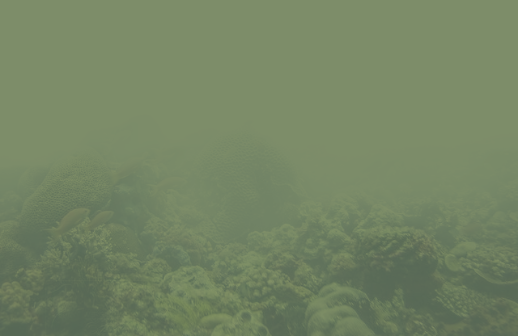}
            \includegraphics[width=0.19\linewidth]{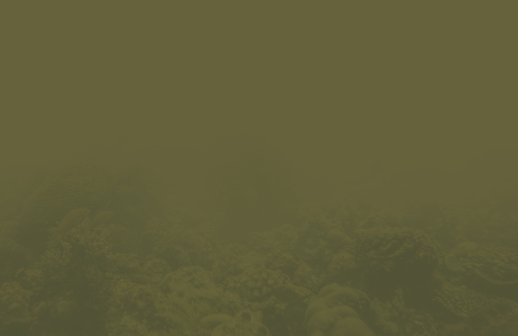}
        \end{subfigure}
    
        \begin{subfigure}{\textwidth}
            \centering
            \includegraphics[width=0.19\linewidth]{figures/results/gmn_8064up/gmn_8064up.jpg}
            \includegraphics[width=0.19\linewidth]{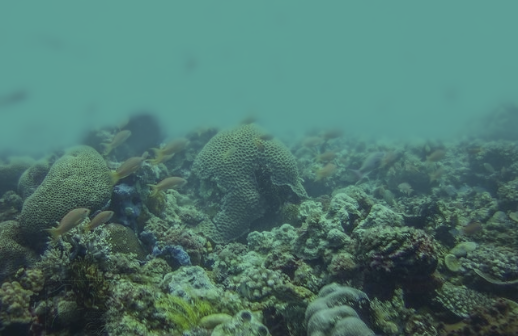}
            \includegraphics[width=0.19\linewidth]{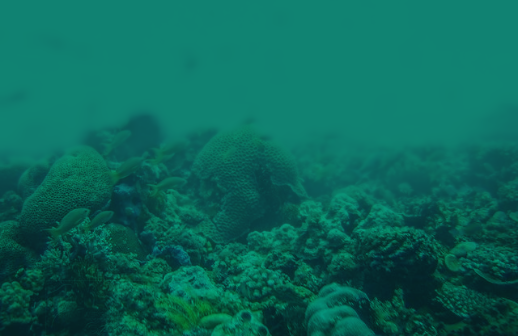}
            \includegraphics[width=0.19\linewidth]{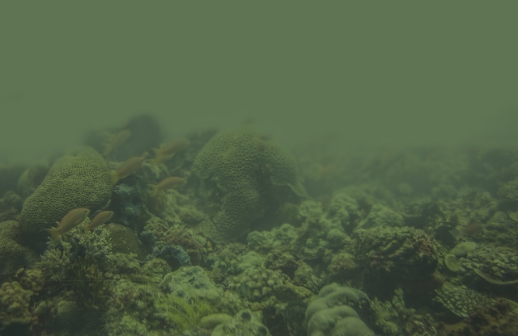}
            \includegraphics[width=0.19\linewidth]{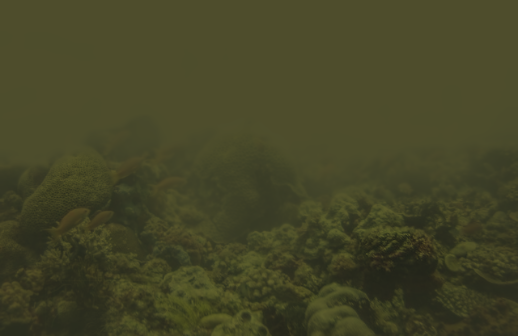}
        \end{subfigure}
        \caption{}
        \label{}
    \end{subfigure}
    \begin{subfigure}{\textwidth}
        \begin{subfigure}{\textwidth}
            \centering
            \includegraphics[width=0.19\linewidth]{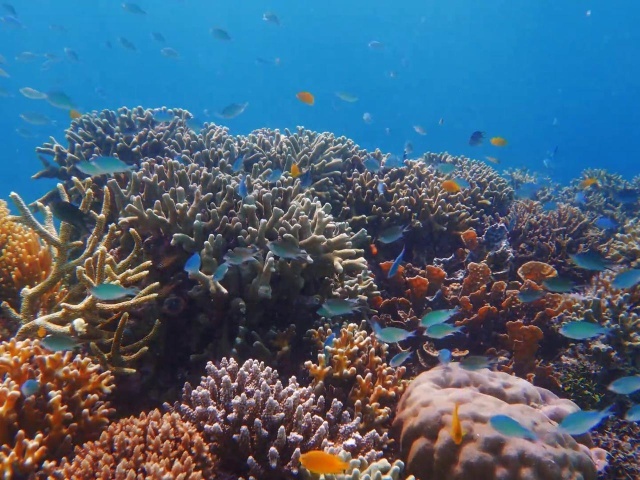}
            \includegraphics[width=0.19\linewidth]{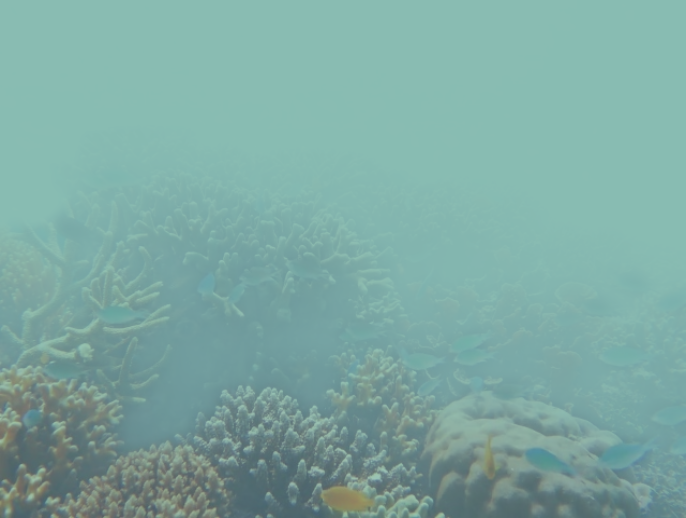}
            \includegraphics[width=0.19\linewidth]{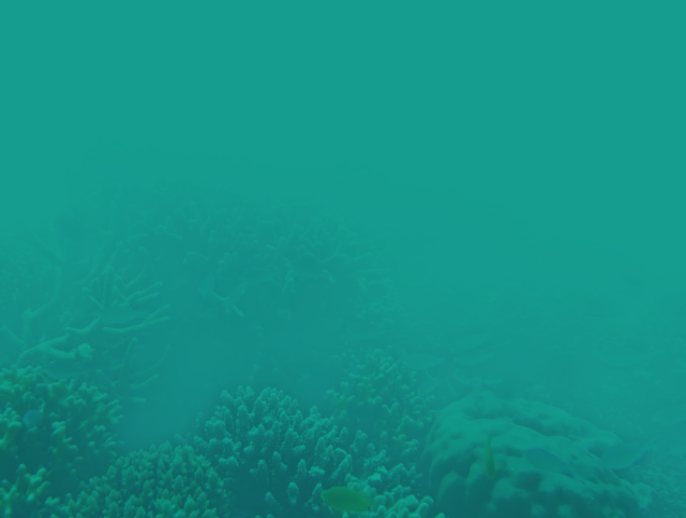}
            \includegraphics[width=0.19\linewidth]{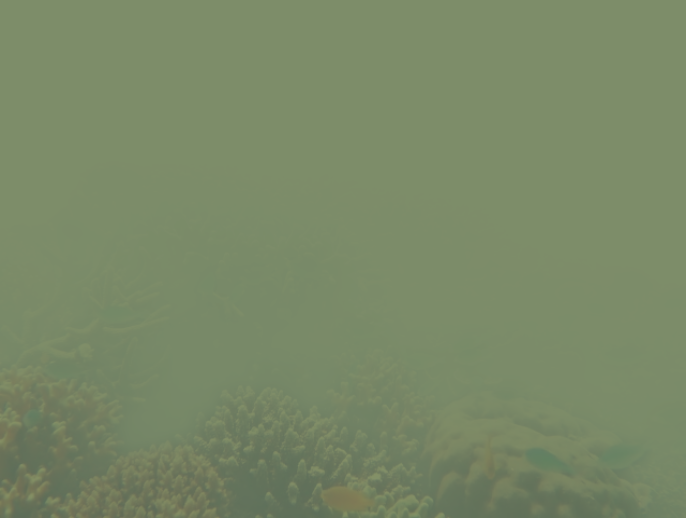}
            \includegraphics[width=0.19\linewidth]{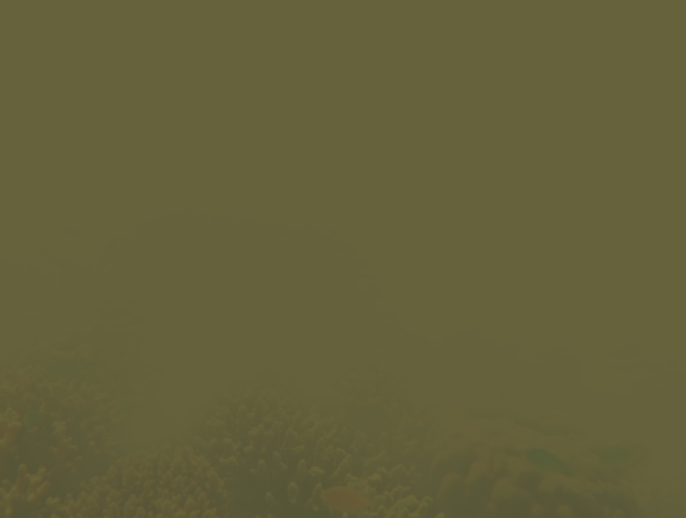}
        \end{subfigure}
        
        \begin{subfigure}{\textwidth}
            \centering
            \includegraphics[width=0.19\linewidth]{figures/results/gmn_7543up/gmn_7543up.jpg}
            \includegraphics[width=0.19\linewidth]{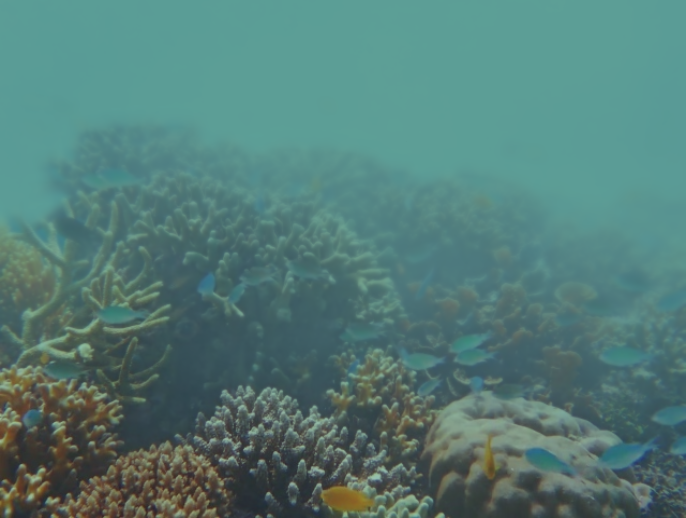}
            \includegraphics[width=0.19\linewidth]{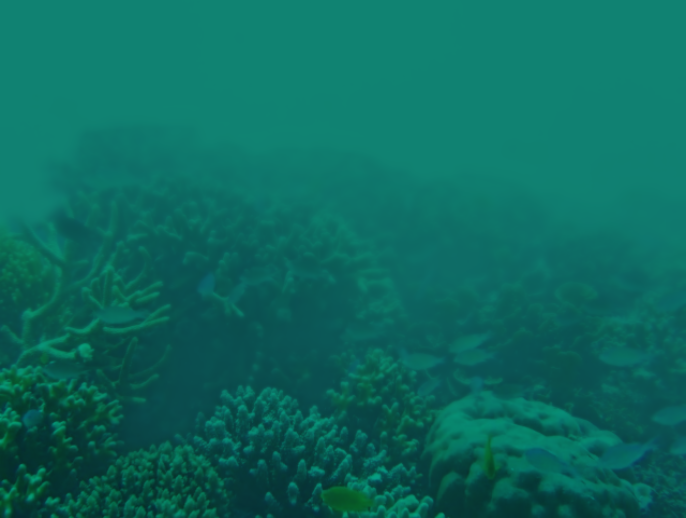}
            \includegraphics[width=0.19\linewidth]{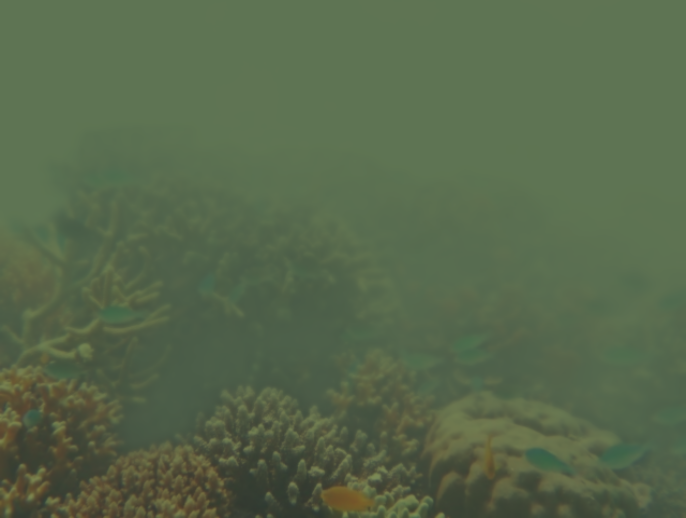}
            \includegraphics[width=0.19\linewidth]{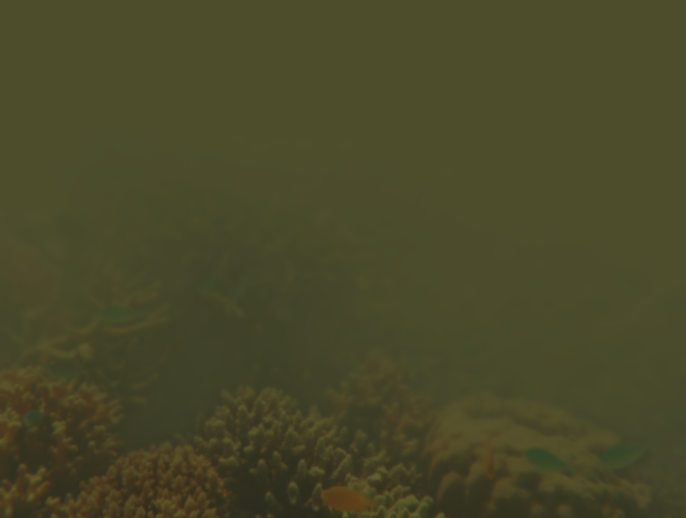}
        \end{subfigure}
        \caption{}
        \label{}
    \end{subfigure}
  
    \begin{subfigure}{\textwidth}
        \begin{subfigure}{\textwidth}
        \centering
            \includegraphics[width=0.19\linewidth]{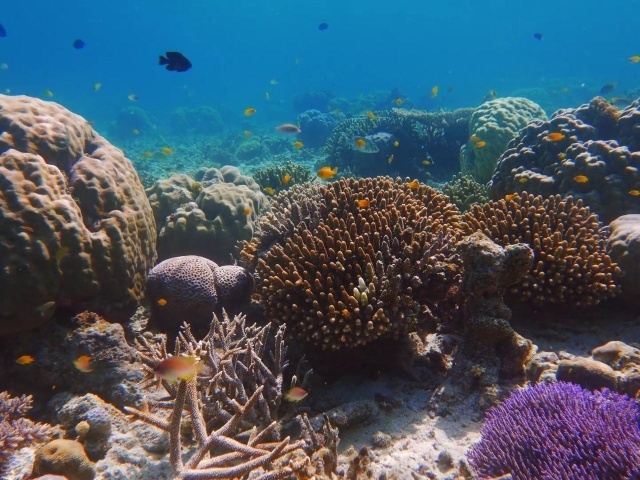}
            \includegraphics[width=0.19\linewidth]{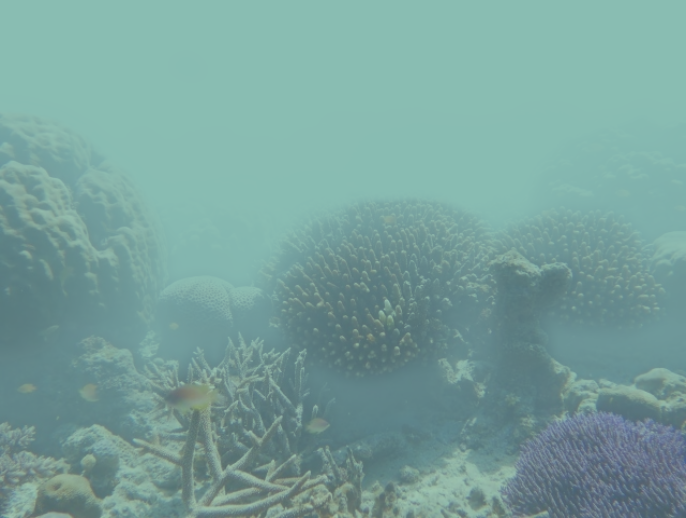}
            \includegraphics[width=0.19\linewidth]{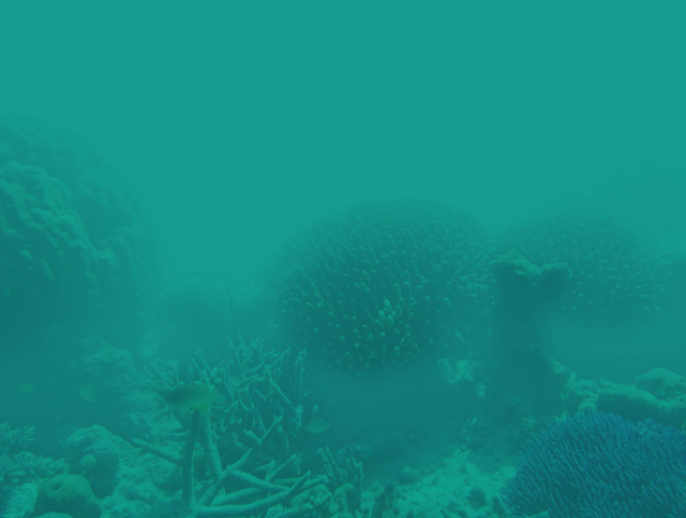}
            \includegraphics[width=0.19\linewidth]{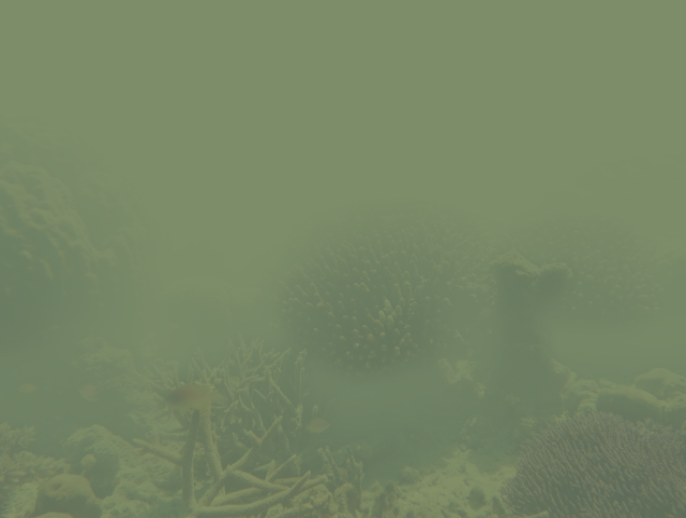}
            \includegraphics[width=0.19\linewidth]{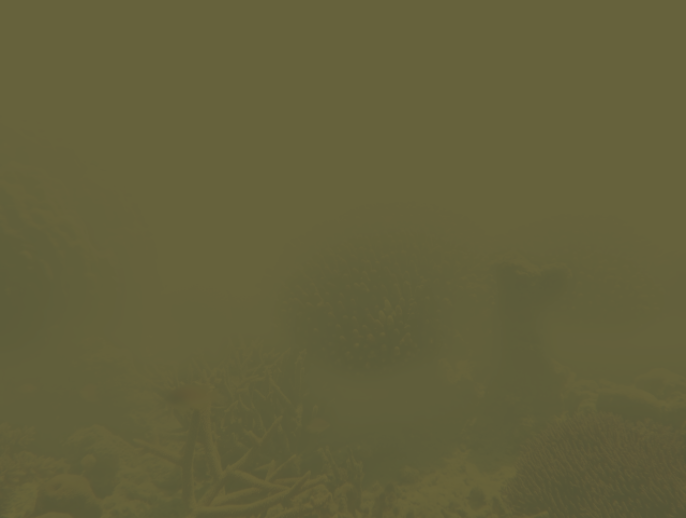}
        \end{subfigure}
    
        \begin{subfigure}{\textwidth}
            \centering
            \includegraphics[width=0.19\linewidth]{figures/results/gmn_8122up/gmn_8122up.jpg}
            \includegraphics[width=0.19\linewidth]{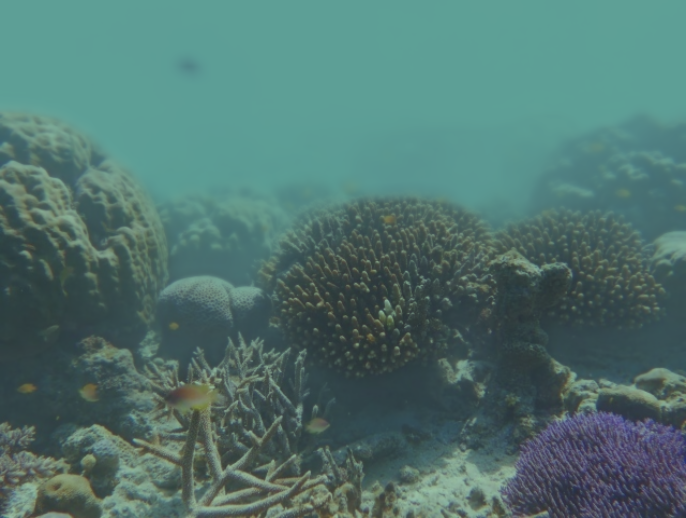}
            \includegraphics[width=0.19\linewidth]{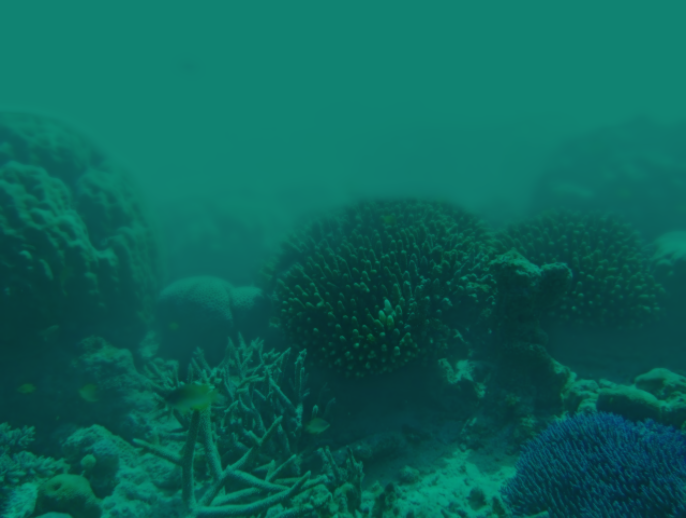}
            \includegraphics[width=0.19\linewidth]{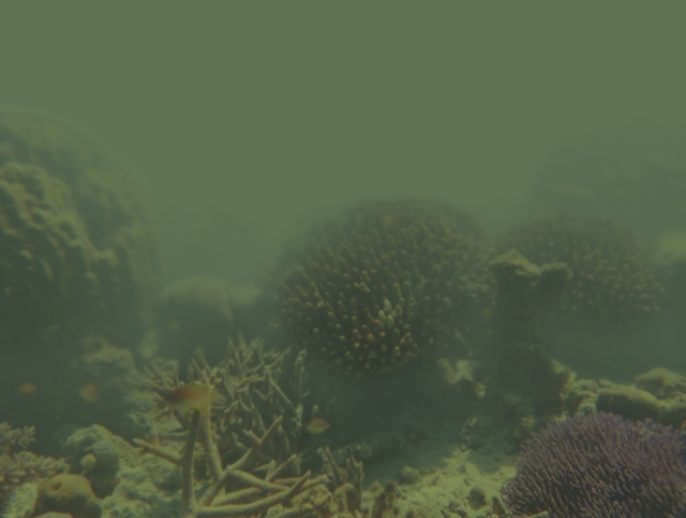}
            \includegraphics[width=0.19\linewidth]{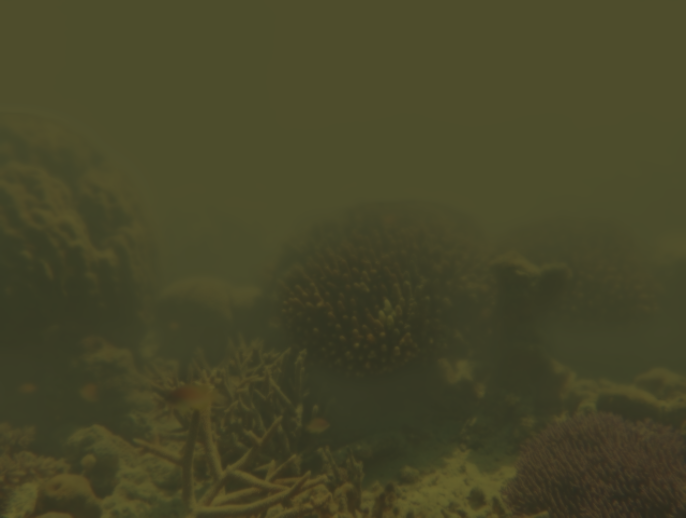}
        \end{subfigure}
        \caption{}
        \label{}
    \end{subfigure}
    \begin{subfigure}{\textwidth}
        \begin{subfigure}{\textwidth}
            \centering
            \includegraphics[width=0.19\linewidth]{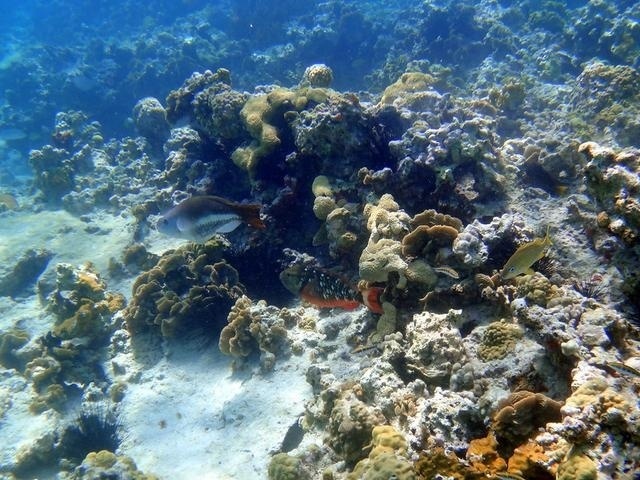}
            \includegraphics[width=0.19\linewidth]{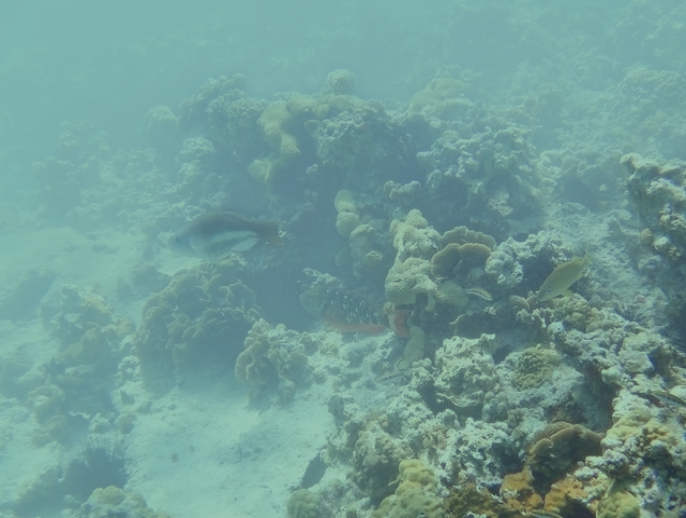}
            \includegraphics[width=0.19\linewidth]{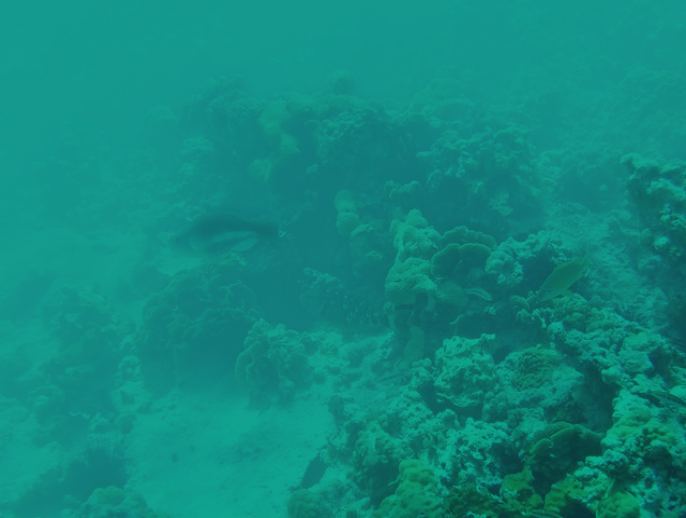}
            \includegraphics[width=0.19\linewidth]{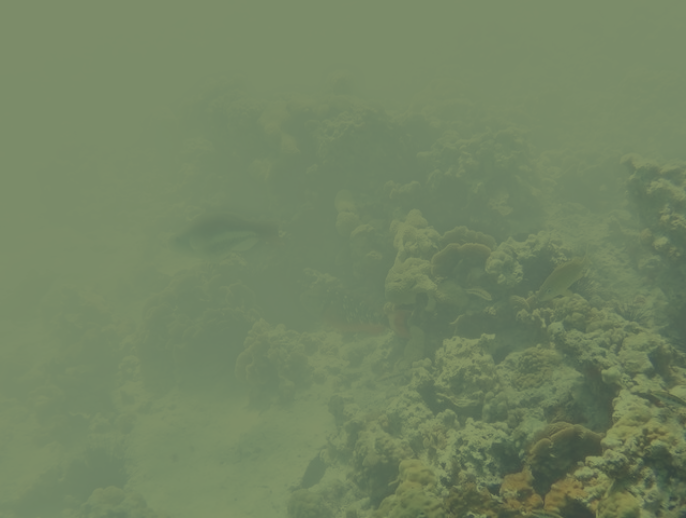}
            \includegraphics[width=0.19\linewidth]{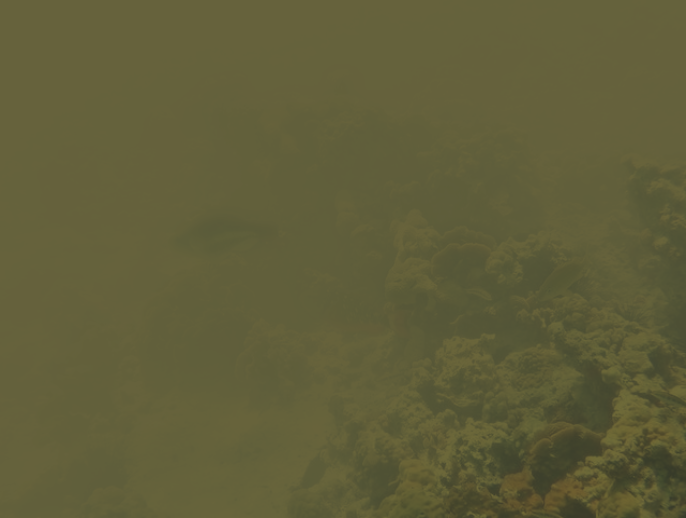}
        \end{subfigure}
        
        \begin{subfigure}{\textwidth}
            \centering
            \includegraphics[width=0.19\linewidth]{figures/results/gmn_7385up/gmn_7385up.jpg}
            \includegraphics[width=0.19\linewidth]{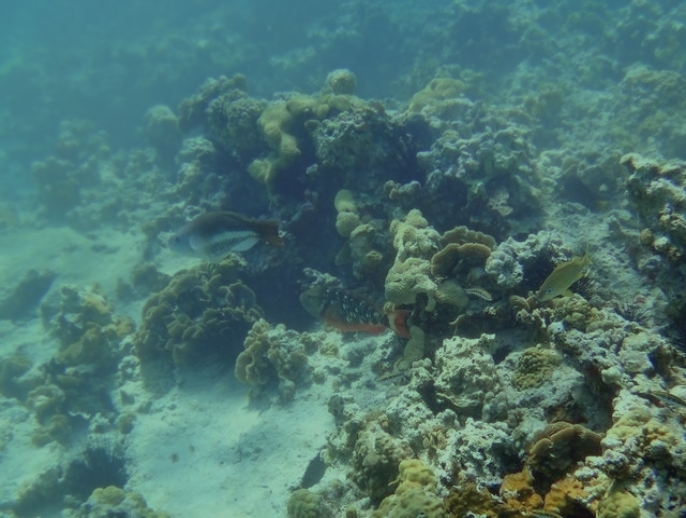}
            \includegraphics[width=0.19\linewidth]{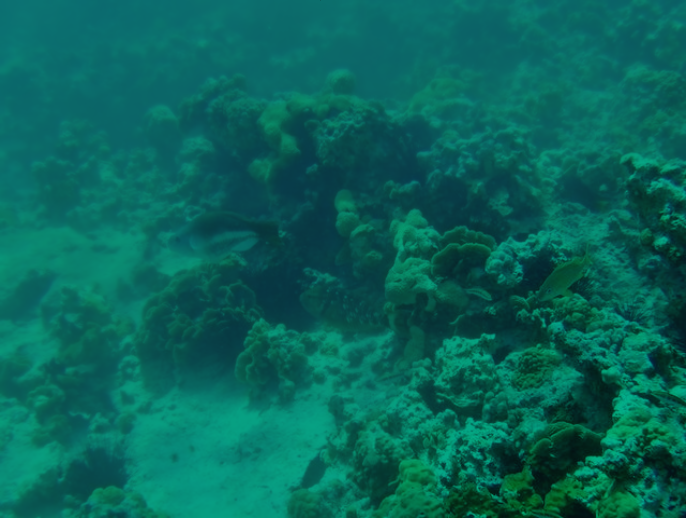}
            \includegraphics[width=0.19\linewidth]{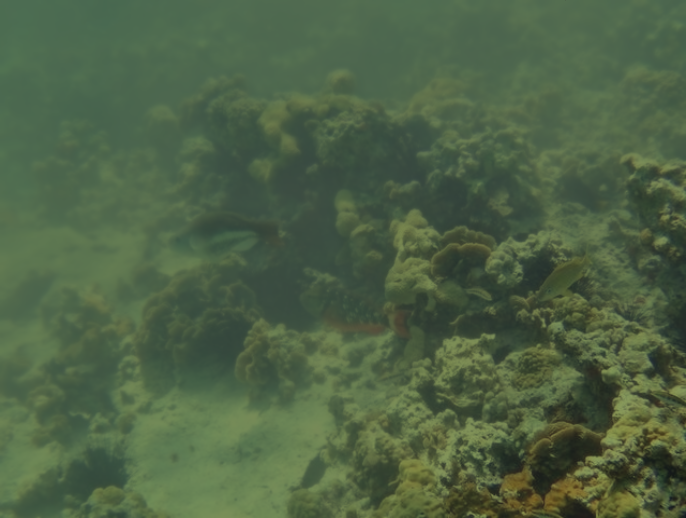}
            \includegraphics[width=0.19\linewidth]{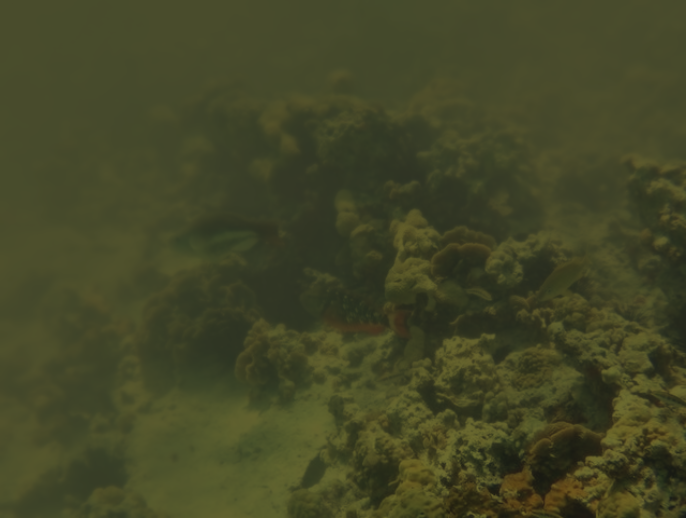}
        \end{subfigure}
        \caption{}
        \label{}
    \end{subfigure}
    \caption{Example synthesized images for Jerlov water types 3C-9C \cite{solonenko_inherent_2015}. The first column is the clean source image \cite{islam_fast_2020}. For each figure (a–d): top row shows the reference model; bottom row shows our proposed model. Differences are best observed when zoomed in.}
    \label{fig:res:EUVP}
\end{figure*}

\subsection{Evaluation survey}

In order to evaluate the proposed pipeline on images without references, we collected the mean opinion ranking (MOR) through a survey. Images were randomly selected from the EUVP dataset \cite{islam_fast_2020}, and manually filtered to remove any footage that was either not clear enough or with a horizontal depth too short to showcase the effects of turbidity (e.g., zoomed-in pictures of a fish). A subset of these images was then randomly split into 9 groups of 2, corresponding to the water types IA-9C. Water type I was skipped due to its high similarity with IA and low overall degradation.

Images had their conditions synthetically altered with both the reference model and ours, resulting in 18 pairs. The vertical and horizontal depths were set to 1m and 1-5m, respectively, across all images in the survey. These values are not necessarily accurate for all images, but the process is representative of randomized data generation at scale. For each water type, coefficients were taken from \cite{solonenko_inherent_2015}, and the spectral response of a \textit{Nikon D90} was considered \cite{jiang_what_2013}. All other model parameters are listed in \cref{tab:params}

The 20 survey participants were experienced in the field of computer vision and potentially familiar with synthetic data, though not necessarily with underwater environments. They were shown a set of unedited underwater images of different conditions at the start of the survey as a reference, and were then asked to choose the most realistic image for each pair. The order in which choices appeared was randomized. All survey images can be found in the supplementary, while results are displayed in \cref{fig:res:survey}.

\begin{table}[]
    \centering
    \begin{tabular}{|c|c|c|c|c|}\hline
    $g$ & $\mu$ & $\phi$ & GRF \\\hline
    $0.2$ & $0.3$ & $0.3 \cdot \overline{b}$ & $[0.7 - 1.3]$ \\\hline
    \end{tabular}
    \caption{Model parameters. $\overline{b}$ is the mean across all channels.}
    \label{tab:params}
\end{table}

\begin{figure}[]
\centering
    \begin{subfigure}{\linewidth}
    \centering
        \includegraphics[height=0.73in]{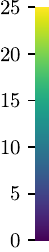}
        \includegraphics[width=0.3\linewidth]
        {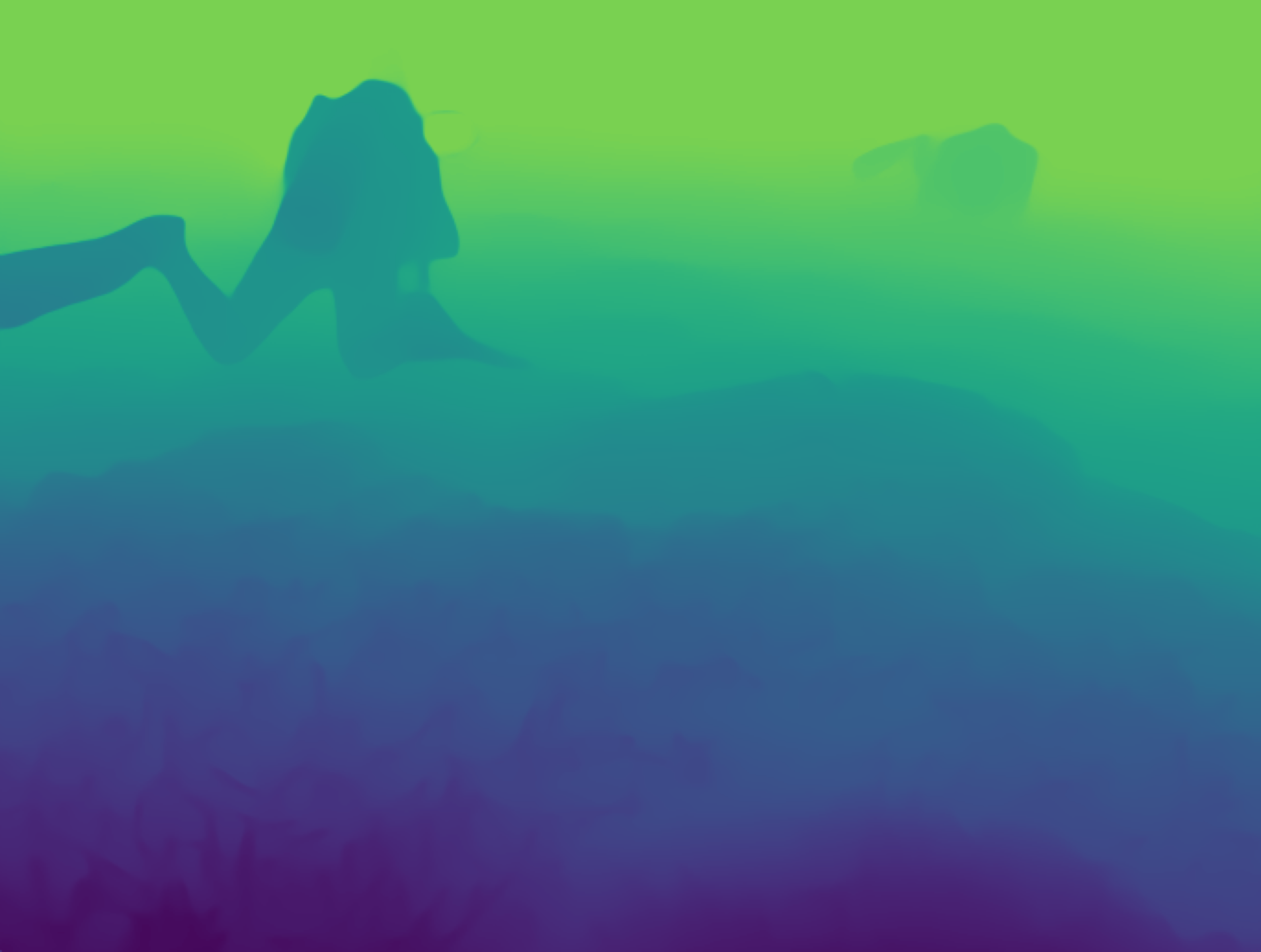}
        \includegraphics[width=0.3\linewidth]{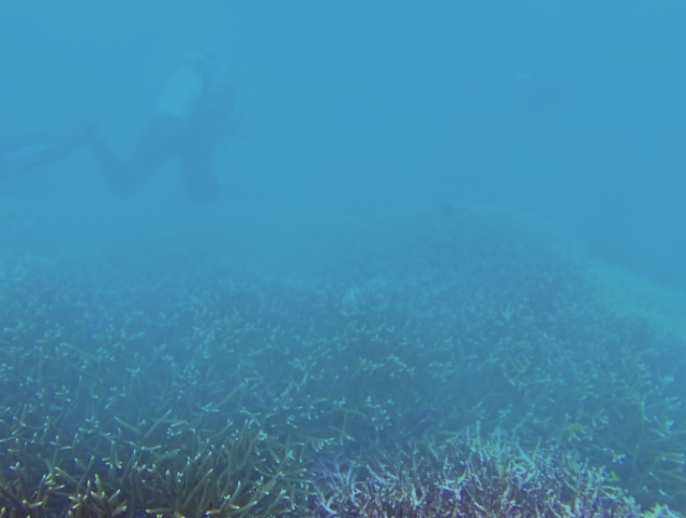}
        \includegraphics[width=0.3\linewidth]{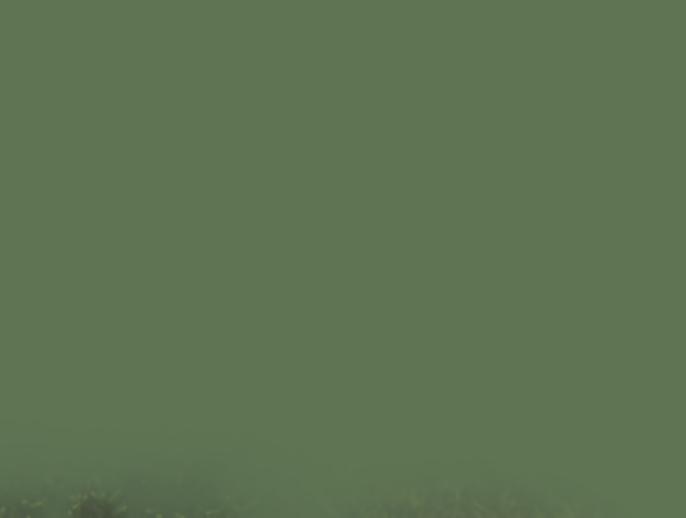}
    \end{subfigure}
        
    \begin{subfigure}{\linewidth}
        \centering
        \includegraphics[height=0.73in]{figures/results/depth/gmn_5594up/colorbar_only.pdf}
        \includegraphics[width=0.3\linewidth]{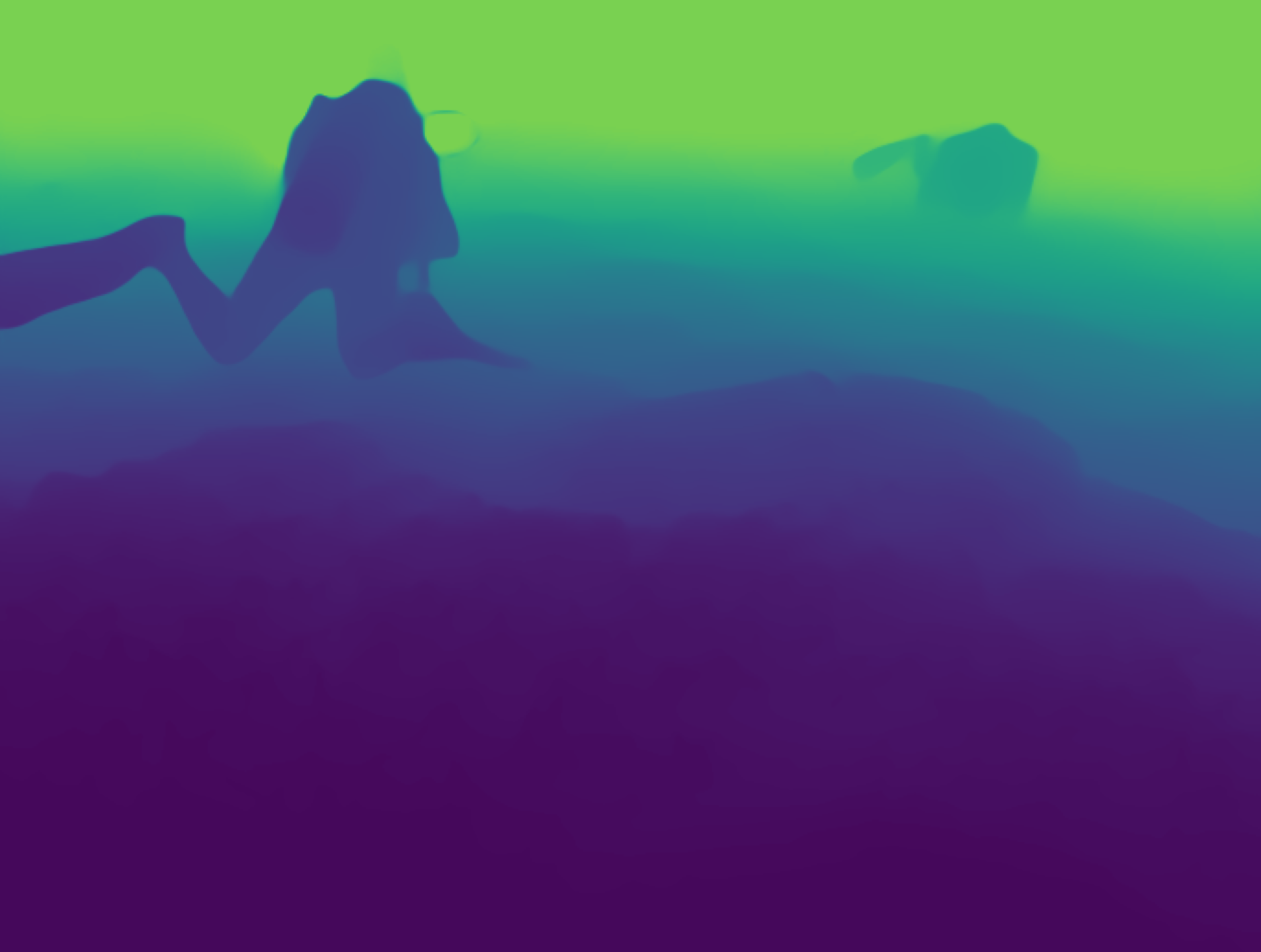}
        \includegraphics[width=0.3\linewidth]{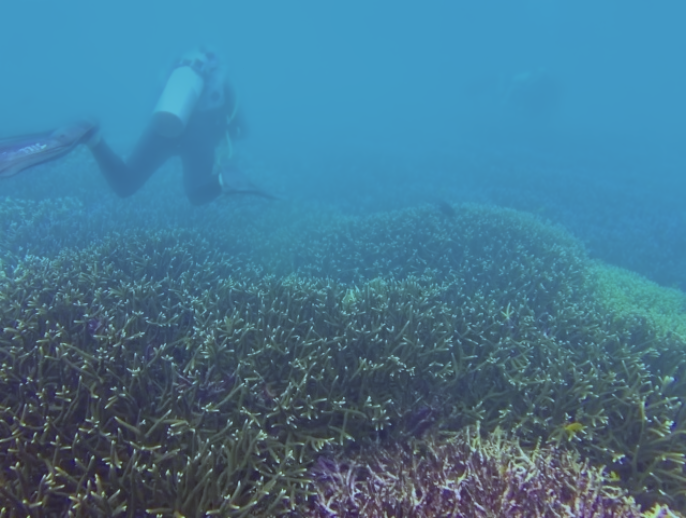}
        \includegraphics[width=0.3\linewidth]{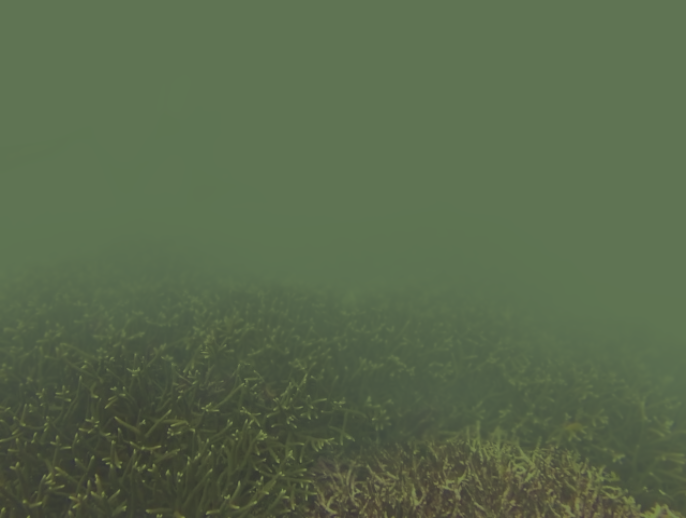}
    \end{subfigure}

    \begin{subfigure}{\linewidth}
        \centering
        \includegraphics[height=0.73in]{figures/results/depth/gmn_5594up/colorbar_only.pdf}
        \includegraphics[width=0.3\linewidth]{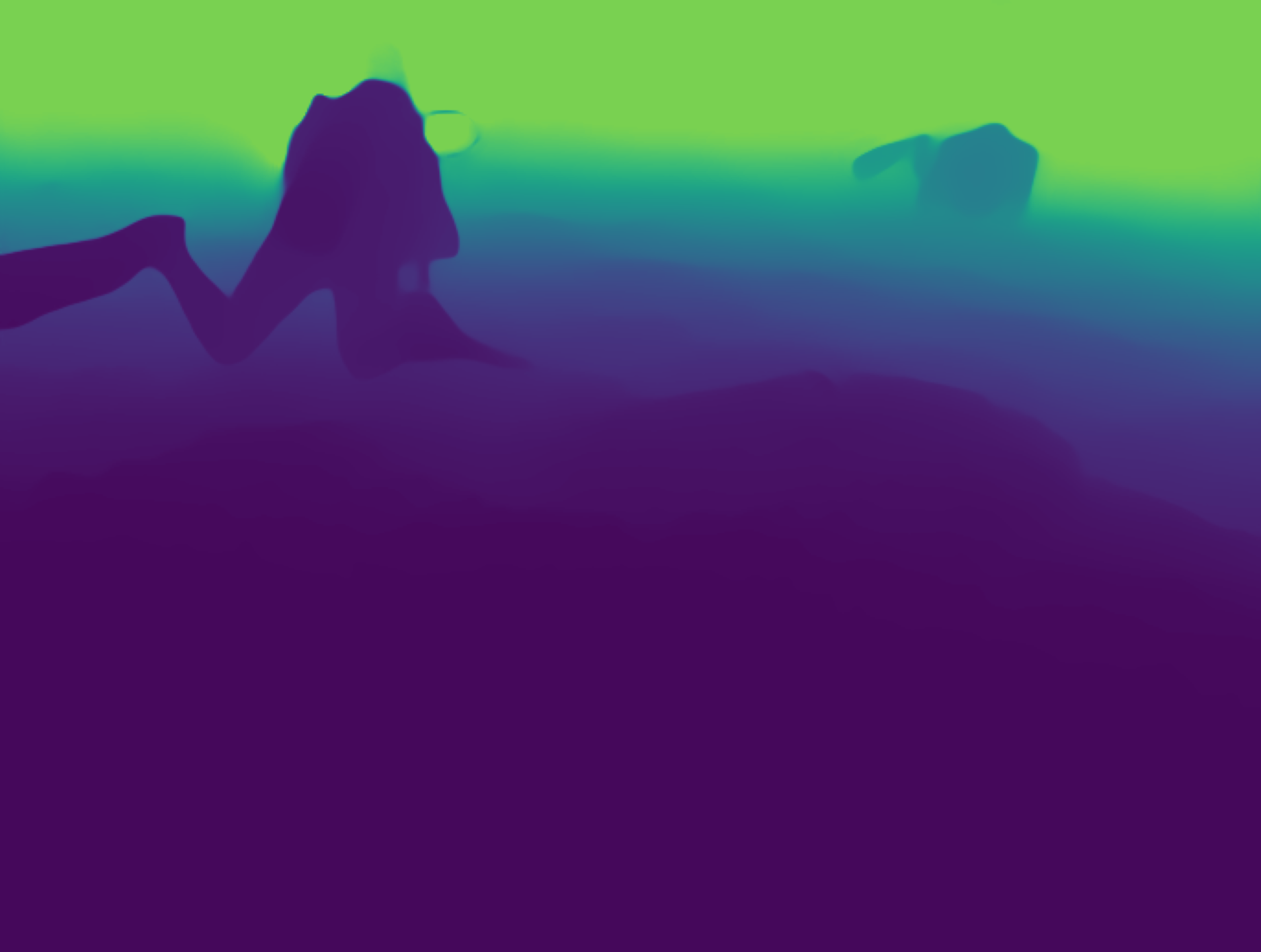}
        \includegraphics[width=0.3\linewidth]{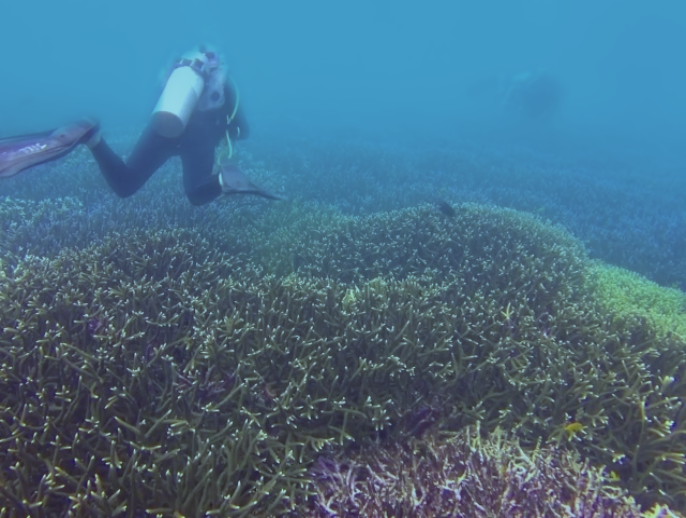}
        \includegraphics[width=0.3\linewidth]{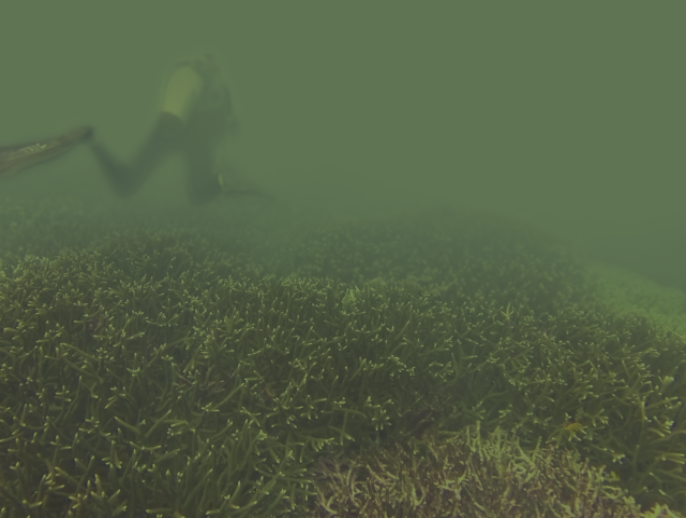}
    \end{subfigure}
    \caption{Effect of depth scaling. Simulation of type II and 7C waters is shown next to the scaled depth map. Top row: 0-20m with linear scaling. Middle row: 0-20m with a gamma value of 3. Bottom row: 0-20m with a gamma value of 6.}
    \label{fig:exp:depth}
\end{figure}

Randomly scaling the horizontal depth, as done here and in prior work \cite{liu_model-based_2022}, can have undesirable outcomes. To study this effect, we alter depth scaling on an image from the EUVP dataset \cite{islam_fast_2020}, while maintaining all other parameters consistent. The results can be seen in \cref{fig:exp:depth}. Finally, a few example images with customized depth and parameters can be seen in \cref{fig:res:EUVP}.

\begin{figure}[]
    \centering
    \includegraphics[width=0.9\linewidth]{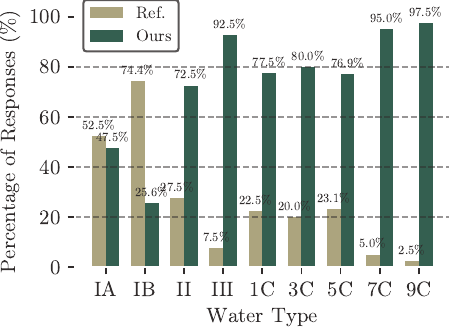}
    \caption{Survey results. Question: \textit{Given some examples of real underwater images, choose the one that looks more realistic.}}
    \label{fig:res:survey}
\end{figure}

\section{Discussion}
\label{sec:discussion}

Contrary to the conclusions of \cite{schechner_clear_2004}, our experiments show that image blur is noticeable, even at short distances in turbid conditions. This can be seen in \cref{fig:res:lab}, and is highlighted in \cref{fig:blurdisplay}, particularly around edges and the baseplate details. While part of this discrepancy can be attributed to the empirical constant $\phi$ in \cref{eq:fscatter}, we believe it is also largely due to an overestimated attenuation. 

As scattering coefficients increase, the reference model assumes a large $\beta$, resulting in a heavily reduced direct signal and overpwoering backscatter. However, when considering that the scattered rays reaching the camera contain information and not only \enquote{blur}, the effective attenuation decreases enough for the blur to be visible.

Within this framework of effective scattering coefficients corresponding to specific scattering angle ranges, the backlight and attenuation intensities can be decoupled to some degree. That is contained within constant $\mu$ (\cref{eq:backlightnew}), which introduces more flexibility and variation to the reference $\frac{b(\lambda)}{\beta(\lambda)}$ term. The practical implications can be seen in \cref{fig:res:lab}, particularly, in the way the last two rows diverge. They both have the same attenuation, but row \textit{d} has a brighter backlight, similar to the ground truth.


Focusing on visual realism, applying GRFs to the depth maps helps the added effects blend more naturally, with transitions that are smoother and less tightly bound to object edges. This results in more gradual and realistic fading. GRFs also help mitigate common artifacts from learning-based depth estimators, such as overly smooth or artificially rounded edges.

Results of the evaluation survey support our claims about realism, with our proposed pipeline being preferred over the reference in 73.9\% of the
responses across all water types and 82.5\% for coastal ones. Notably, the decline in participant choice separation between water types IA and II coincides with a decrease in the forward scattering term, making visual differences increasingly imperceptible.

Despite the improvements, accurate depth remains a weak point of this approach to synthesizing data. \cref{fig:exp:depth} showcases the large divergence between images as the depth scaling is altered. Note how the minimum and maximum values remain identical, and it is only a gamma term that is modified. The potential impact that inaccurate depth can have on downstream tasks remains to be studied.

\section{Conclusion}
\label{sec:conclusion}

In this work, we introduced an improved synthetic underwater data generation process based on the IFM. It incorporates medium inhomogeneity and the forward scattering term, which is commonly overlooked in traditional models. Qualitative evaluations demonstrate that by integrating these components into the pipeline, our approach yields significantly more realistic results, particularly in turbid conditions.

\section{Acknowledgments}
This work was funded by the Pioneer Centre for Artificial Intelligence in Denmark.

\FloatBarrier
{
    \small
    \bibliographystyle{ieeenat_fullname}
    \bibliography{main, references}

\begin{thebibliography}{30}
\providecommand{\natexlab}[1]{#1}
\providecommand{\url}[1]{\texttt{#1}}
\expandafter\ifx\csname urlstyle\endcsname\relax
  \providecommand{\doi}[1]{doi: #1}\else
  \providecommand{\doi}{doi: \begingroup \urlstyle{rm}\Url}\fi

\bibitem[Akkaynak and Treibitz(2018)]{akkaynak_revised_2018}
Derya Akkaynak and Tali Treibitz.
\newblock A {Revised} {Underwater} {Image} {Formation} {Model}.
\newblock In \emph{2018 {IEEE}/{CVF} {Conference} on {Computer} {Vision} and {Pattern} {Recognition}}, pages 6723--6732, 2018.
\newblock ISSN: 2575-7075.

\bibitem[Akkaynak et~al.(2017)Akkaynak, Treibitz, Shlesinger, Loya, Tamir, and Iluz]{akkaynak_what_2017}
Derya Akkaynak, Tali Treibitz, Tom Shlesinger, Yossi Loya, Raz Tamir, and David Iluz.
\newblock What is the {Space} of {Attenuation} {Coefficients} in {Underwater} {Computer} {Vision}?
\newblock In \emph{2017 {IEEE} {Conference} on {Computer} {Vision} and {Pattern} {Recognition} ({CVPR})}, pages 568--577, 2017.
\newblock ISSN: 1063-6919.

\bibitem[Anwar et~al.(2018)Anwar, Li, and Porikli]{anwar_deep_2018}
Saeed Anwar, Chongyi Li, and Fatih Porikli.
\newblock Deep {Underwater} {Image} {Enhancement}, 2018.
\newblock arXiv:1807.03528 [cs].

\bibitem[Desai et~al.(2021)Desai, Tabib, Reddy, Patil, and Mudenagudi]{desai_ruig_2021}
Chaitra Desai, Ramesh~Ashok Tabib, Sai~Sudheer Reddy, Ujwala Patil, and Uma Mudenagudi.
\newblock {RUIG}: {Realistic} {Underwater} {Image} {Generation} {Towards} {Restoration}.
\newblock In \emph{2021 {IEEE}/{CVF} {Conference} on {Computer} {Vision} and {Pattern} {Recognition} {Workshops} ({CVPRW})}, pages 2181--2189, Nashville, TN, USA, 2021. IEEE.

\bibitem[Desai et~al.(2024)Desai, Benur, Patil, and Mudenagudi]{desai_rsuigm_2024}
Chaitra Desai, Sujay Benur, Ujwala Patil, and Uma Mudenagudi.
\newblock {RSUIGM}: {Realistic} {Synthetic} {Underwater} {Image} {Generation} with {Image} {Formation} {Model}.
\newblock \emph{ACM Trans. Multimedia Comput. Commun. Appl.}, 21\penalty0 (1):\penalty0 10:1--10:22, 2024.

\bibitem[Doxaran et~al.(2016)Doxaran, Leymarie, Nechad, Dogliotti, Ruddick, Gernez, and Knaeps]{doxaran_improved_2016}
David Doxaran, Edouard Leymarie, Bouchra Nechad, Ana Dogliotti, Kevin Ruddick, Pierre Gernez, and Els Knaeps.
\newblock Improved correction methods for field measurements of particulate light backscattering in turbid waters.
\newblock \emph{Optics Express}, 24\penalty0 (4):\penalty0 3615--3637, 2016.
\newblock Publisher: Optica Publishing Group.

\bibitem[Galdran et~al.(2015)Galdran, Pardo, Picón, and Alvarez-Gila]{galdran_automatic_2015}
Adrian Galdran, David Pardo, Artzai Picón, and Aitor Alvarez-Gila.
\newblock Automatic {Red}-{Channel} underwater image restoration.
\newblock \emph{Journal of Visual Communication and Image Representation}, 26:\penalty0 132--145, 2015.

\bibitem[Hou et~al.(2020)Hou, Zhao, Pan, Yang, Tan, and Li]{hou_benchmarking_2020}
Guojia Hou, Xin Zhao, Zhenkuan Pan, Huan Yang, Lu Tan, and Jingming Li.
\newblock Benchmarking {Underwater} {Image} {Enhancement} and {Restoration}, and {Beyond}.
\newblock \emph{IEEE Access}, 8:\penalty0 122078--122091, 2020.
\newblock Conference Name: IEEE Access.

\bibitem[Islam et~al.(2020)Islam, Xia, and Sattar]{islam_fast_2020}
Md~Jahidul Islam, Youya Xia, and Junaed Sattar.
\newblock Fast {Underwater} {Image} {Enhancement} for {Improved} {Visual} {Perception}, 2020.
\newblock arXiv:1903.09766 [cs].

\bibitem[Jaffe(1990)]{jaffe_computer_1990}
J.S. Jaffe.
\newblock Computer modeling and the design of optimal underwater imaging systems.
\newblock \emph{IEEE Journal of Oceanic Engineering}, 15\penalty0 (2):\penalty0 101--111, 1990.

\bibitem[Jerlov(1957)]{jerlov_optical_1957}
N.~G. Jerlov.
\newblock \emph{Optical studies of ocean waters}.
\newblock Elanders boktr., Goteborg, 1957.

\bibitem[Jiang et~al.(2013)Jiang, Liu, Gu, and Susstrunk]{jiang_what_2013}
Jun Jiang, Dengyu Liu, Jinwei Gu, and Sabine Susstrunk.
\newblock What is the space of spectral sensitivity functions for digital color cameras?
\newblock In \emph{2013 {IEEE} {Workshop} on {Applications} of {Computer} {Vision} ({WACV})}, pages 168--179, Clearwater Beach, FL, USA, 2013. IEEE.

\bibitem[Kiefer et~al.()Kiefer, Zust, Kristan, Pers, Tersek, Mudenagudi, Desai, Wiliem, Kreis, Akalwadi, Zhong, Zhang, Liu, Chen, Yang, Fabijanic, Ferreira, Lee, Yao, Kumar, Marcus, Novak, Feng, Cheng, Nguyen, Sheikh, Saric, Li, Lu, Lin, Yang, Cheng, Awad, Muhovic, Quan, Lee, Lee, Guan, Huang, Ni, Lin, Lee, Hsu, Michel, Gross, Jiang, Feng, Lucas, Saleem, Lin, and Weinmann]{kiefer_3rd_nodate}
Benjamin Kiefer, Lojze Zust, Matej Kristan, Janez Pers, Matija Tersek, Uma Mudenagudi, Chaitra Desai, Arnold Wiliem, Marten Kreis, Nikhil Akalwadi, Zhiqiang Zhong, Zhe Zhang, Sujie Liu, Xuran Chen, Yang Yang, Matej Fabijanic, Fausto Ferreira, Seongju Lee, Shanliang Yao, Himanshu Kumar, Aurelius Marcus, Gregor Novak, Yuan Feng, Annie Cheng, Thien Nguyen, Jannik Sheikh, Josip Saric, Zhuoxiao Li, Yutang Lu, Yipeng Lin, Xiang Yang, Ching-Heng Cheng, Ali Awad, Jon Muhovic, Yitong Quan, Junseok Lee, Kyoobin Lee, Runwei Guan, Xiaoyu Huang, Yi Ni, Tzu-Yu Lin, Chia-Ming Lee, Chih-Chung Hsu, Andreas Michel, Wolfgang Gross, Nan Jiang, Fei Feng, Evan Lucas, Ashraf Saleem, Yu-Fan Lin, and Martin Weinmann.
\newblock 3rd {Workshop} on {Maritime} {Computer} {Vision} ({MaCVi}) 2025: {Challenge} {Results}.

\bibitem[Li et~al.(2020)Li, Anwar, and Porikli]{li_underwater_2020}
Chongyi Li, Saeed Anwar, and Fatih Porikli.
\newblock Underwater scene prior inspired deep underwater image and video enhancement.
\newblock \emph{Pattern Recognition}, 98:\penalty0 107038, 2020.

\bibitem[Li et~al.(2017)Li, Skinner, Eustice, and Johnson-Roberson]{li_watergan_2017}
Jie Li, Katherine~A. Skinner, Ryan~M. Eustice, and Matthew Johnson-Roberson.
\newblock {WaterGAN}: {Unsupervised} {Generative} {Network} to {Enable} {Real}-time {Color} {Correction} of {Monocular} {Underwater} {Images}.
\newblock \emph{IEEE Robotics and Automation Letters}, pages 1--1, 2017.

\bibitem[Li et~al.(2025)Li, Dong, He, Yang, Zhou, and Hu]{li_multimodal_2025}
Xin Li, Yanni Dong, Junge He, Xin Yang, Wenlong Zhou, and Xiangyun Hu.
\newblock Multimodal {U}-{Net}: {A} {Novel} {Approach} for 2-{D} {Inversion} of {Magnetotelluric} {Data}.
\newblock \emph{IEEE Transactions on Geoscience and Remote Sensing}, 63:\penalty0 1--12, 2025.

\bibitem[Li and Snavely(2018)]{li_megadepth_2018}
Zhengqi Li and Noah Snavely.
\newblock {MegaDepth}: {Learning} {Single}-{View} {Depth} {Prediction} from {Internet} {Photos}.
\newblock In \emph{2018 {IEEE}/{CVF} {Conference} on {Computer} {Vision} and {Pattern} {Recognition}}, pages 2041--2050, Salt Lake City, UT, USA, 2018. IEEE.

\bibitem[Liu et~al.(2022)Liu, Xu, Zhang, Sun, Yang, Li, Li, and Quan]{liu_model-based_2022}
Yidan Liu, Huiping Xu, Bing Zhang, Kelin Sun, Jingchuan Yang, Bo Li, Chen Li, and Xiangqian Quan.
\newblock Model-{Based} {Underwater} {Image} {Simulation} and {Learning}-{Based} {Underwater} {Image} {Enhancement} {Method}.
\newblock \emph{Information}, 13\penalty0 (4):\penalty0 187, 2022.

\bibitem[McGlamery(1980)]{mcglamery_computer_1980}
B.~L. McGlamery.
\newblock A {Computer} {Model} {For} {Underwater} {Camera} {Systems}.
\newblock In \emph{Ocean {Optics} {VI}}, pages 221--231. SPIE, 1980.

\bibitem[Mobley(1994)]{mobley_light_1994}
Curtis Mobley.
\newblock \emph{Light and {Water}: {Radiative} {Transfer} in {Natural} {Waters}}.
\newblock 1994.
\newblock Journal Abbreviation: Academic Press Publication Title: Academic Press.

\bibitem[Schechner and Karpel(2004)]{schechner_clear_2004}
Y.Y. Schechner and N. Karpel.
\newblock Clear underwater vision.
\newblock In \emph{Proceedings of the 2004 {IEEE} {Computer} {Society} {Conference} on {Computer} {Vision} and {Pattern} {Recognition}, 2004. {CVPR} 2004.}, pages 536--543, Washington, DC, USA, 2004. IEEE.

\bibitem[Silberman et~al.(2012)Silberman, Hoiem, Kohli, and Fergus]{silberman_indoor_2012}
Nathan Silberman, Derek Hoiem, Pushmeet Kohli, and Rob Fergus.
\newblock Indoor {Segmentation} and {Support} {Inference} from {RGBD} {Images}.
\newblock In \emph{Computer {Vision} – {ECCV} 2012}, pages 746--760, Berlin, Heidelberg, 2012. Springer.

\bibitem[Solonenko and Mobley(2015)]{solonenko_inherent_2015}
Michael~G. Solonenko and Curtis~D. Mobley.
\newblock Inherent optical properties of {Jerlov} water types.
\newblock \emph{Applied Optics}, 54\penalty0 (17):\penalty0 5392--5401, 2015.
\newblock Publisher: Optica Publishing Group.

\bibitem[Sooknanan et~al.(2012)Sooknanan, Kokaram, Corrigan, Baugh, Wilson, and Harte]{sooknanan_improving_2012}
K. Sooknanan, A. Kokaram, D. Corrigan, G. Baugh, J. Wilson, and N. Harte.
\newblock Improving underwater visibility using vignetting correction.
\newblock page 83050M, Burlingame, California, USA, 2012.

\bibitem[Tuchow et~al.(2016)Tuchow, Broughton, and Kudela]{tuchow_sensitivity_2016}
Noah Tuchow, Jennifer Broughton, and Raphael Kudela.
\newblock Sensitivity analysis of volume scattering phase functions.
\newblock \emph{Optics Express}, 24\penalty0 (16):\penalty0 18559--18570, 2016.
\newblock Publisher: Optica Publishing Group.

\bibitem[Ueda et~al.(2019)Ueda, Yamada, and Tanaka]{ueda_underwater_2019}
Takumi Ueda, Koki Yamada, and Yuichi Tanaka.
\newblock Underwater {Image} {Synthesis} from {RGB}-{D} {Images} and its {Application} to {Deep} {Underwater} {Image} {Restoration}.
\newblock In \emph{2019 {IEEE} {International} {Conference} on {Image} {Processing} ({ICIP})}, pages 2115--2119, 2019.
\newblock ISSN: 2381-8549.

\bibitem[Wandelt(2013)]{wandelt_gaussian_2013}
Benjamin~D. Wandelt.
\newblock Gaussian {Random} {Fields} in {Cosmostatistics}.
\newblock In \emph{Astrostatistical {Challenges} for the {New} {Astronomy}}, pages 87--105. Springer, New York, NY, 2013.

\bibitem[Wang et~al.(2019)Wang, Zhou, Han, Zhu, and Zheng]{wang_uwgan_2019}
Nan Wang, Yabin Zhou, F. Han, H. Zhu, and Yaojing Zheng.
\newblock {UWGAN}: {Underwater} {GAN} for {Real}-world {Underwater} {Color} {Restoration} and {Dehazing}.
\newblock \emph{ArXiv}, 2019.

\bibitem[Wang et~al.(2025)Wang, Ma, Li, Zhang, Mi, and Fu]{wang_underwater_2025}
Yulin Wang, Yueming Ma, Yuanyuan Li, Jiqing Zhang, Zetian Mi, and Xianping Fu.
\newblock Underwater {Vignetting} {Image} {Correction} {Based} on {Binary} {Polynomial} {Regularization} and {Latent} {Low}-{Rank} {Representation}.
\newblock \emph{IEEE Transactions on Circuits and Systems for Video Technology}, 35\penalty0 (4):\penalty0 3410--3425, 2025.

\bibitem[Yang et~al.(2024)Yang, Kang, Huang, Zhao, Xu, Feng, and Zhao]{yang_depth_2024}
Lihe Yang, Bingyi Kang, Zilong Huang, Zhen Zhao, Xiaogang Xu, Jiashi Feng, and Hengshuang Zhao.
\newblock Depth {Anything} {V2}, 2024.
\newblock arXiv:2406.09414 [cs].

\end{thebibliography}
}

\end{document}


\maketitle

\begin{table*}[b]
    \centering
    \begin{tabular}{|c|c|c|c|c|c|c|}\hline
        $a_c$ & $b_c $& $\beta_{eff}/G_c$ & g & mu & phi & GRF  \\\hline
        \multicolumn{7}{|c|}{Reference model (second row of Fig. 4) }\\\hline
        $[0.72, 0.77, 1.00]$ & $[9.5, 10.25, 13.25] $& $[10.22, 11.02, 14.25]$ & 1 & 1 &0 & $[0.3 - 1.7]$\\
        $[1.06, 1.14, 1.47]$ & $[15.58, 16.81, 21.73] $& $[16.64, 17.95, 23.20]$ & 1& 1& 0 & $[0.3 - 1.7]$\\
        $[1.39, 1.50, 1.94]$ & $[21.66, 23.37, 30.21] $& $[23.05, 24.87, 32.15]$ & 1& 1& 0 & $[0.3 - 1.7]$\\
        $[1.56, 1.69, 2.18]$ & $[24.70, 26.65, 34.45] $& $[26.26, 28.34, 36.63]$ & 1& 1& 0 & $[0.3 - 1.7]$\\\hline
        \multicolumn{7}{|c|}{Reference model with adjusted coefficients (third row of Fig. 4)}\\\hline
 
        $[0.72, 0.77, 1.00]$ & $[9.5, 10.25, 13.25] $& $[2.62, 2.82, 3.65]$ & 0.2& 0.2& 0 & $[0.3 - 1.7]$\\
        $[1.06, 1.14, 1.47]$ & $[15.58, 16.81, 21.73] $& $[4.17, 4.50, 5.82]$ & 0.2& 0.2& 0 & $[0.3 - 1.7]$\\
        $[1.39, 1.50, 1.94]$ & $[21.66, 23.37, 30.21] $& $[5.73, 6.18, 7.99]$ & 0.2& 0.2& 0 & $[0.3 - 1.7]$\\
        $[1.56, 1.69, 2.18]$ & $[24.70, 26.65, 34.45] $& $[6.50, 7.02, 9.07]]$ & 0.2& 0.2& 0 & $[0.3 - 1.7]$\\\hline
        \multicolumn{7}{|c|}{Ours (fourth row of Fig. 4) }\\\hline
        $[0.72, 0.77, 1.00]$ & $[9.5, 10.25, 13.25] $& $[2.62, 2.82, 3.65]$ & 0.2& 0.3& $0.3\cdot mean(b)$ & $[0.3 - 1.7]$\\
        $[1.06, 1.14, 1.47]$ & $[15.58, 16.81, 21.73] $& $[4.17, 4.50, 5.82]$ & 0.2& 0.3& $0.3\cdot mean(b)$ & $[0.3 - 1.7]$\\
        $[1.39, 1.50, 1.94]$ & $[21.66, 23.37, 30.21] $& $[5.73, 6.18, 7.99]$ & 0.2& 0.3& $0.3\cdot mean(b)$ & $[0.3 - 1.7]$\\
        $[1.56, 1.69, 2.18]$ & $[24.70, 26.65, 34.45] $& $[6.50, 7.02, 9.07]$ & 0.2& 0.3& $0.3\cdot mean(b)$ & $[0.3 - 1.7]$\\\hline
    \end{tabular}
    \caption{Model parameters for results in Fig. 4 of the paper. Each row of the table corresponds to one of the turbid images. The first column in Fig. 4 shows the clean input image and as such is not included here. All images share: $d=0.5$ and $z(x)=[0.3-0.7]$}
    \label{tab:my_label}
\end{table*}

\begin{table*}[b]
    \centering
    \begin{tabular}{|c|c|c|c|c|c|c|}\hline
        d & z(x) & z gamma & g & mu & phi & GRF \\\hline
        1 & [0.5 - 10] & 5 & 0.3 & 0.3 &  $0.8 \cdot mean(b)$ & $[0.3 - 1.7]$\\
        1 & [0.5 - 7] & 4 & 0.3 & 0.3 &  $0.8 \cdot mean(b)$ & $[0.3 - 1.7]$\\
        1 & [0.5 - 10] & 4 & 0.3 & 0.3 &  $0.8 \cdot mean(b)$ & $[0.3 - 1.7]$\\
        1 & [0.25 - 2] & 2.5 & 0.3 & 0.3 & $0.8 \cdot mean(b)$ & $[0.3 - 1.7]$\\\hline
    \end{tabular}
    \caption{Model parameters for the example images in Fig. 7 of the paper. Each row corresponds to one of the images. The attenuation coefficients were chosen according to Jerlov water types.}
    \label{tab:my_label}
\end{table*}

\begin{figure*}[h]
\centering
    \begin{subfigure}{0.35\textwidth}
       \includegraphics[width=0.49\linewidth]{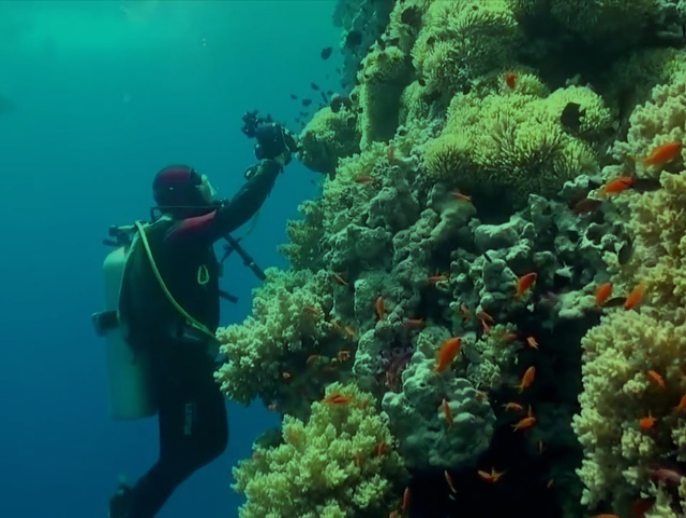}
        \includegraphics[width=0.49\linewidth]{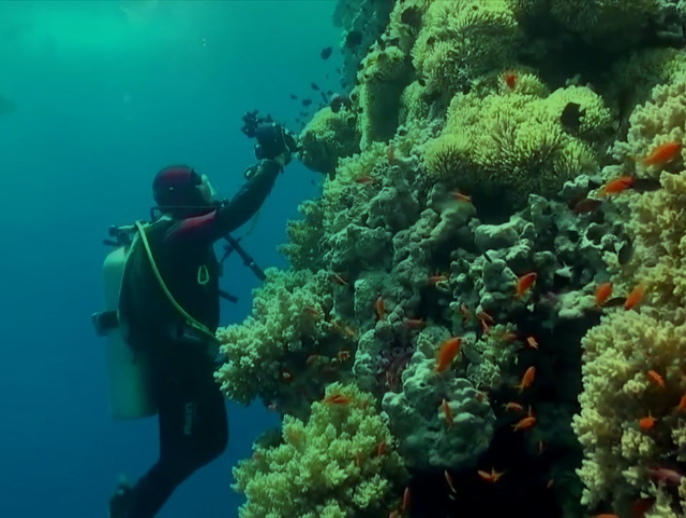}
   \end{subfigure}
    \begin{subfigure}{0.35\textwidth}
       \includegraphics[width=0.49\linewidth]{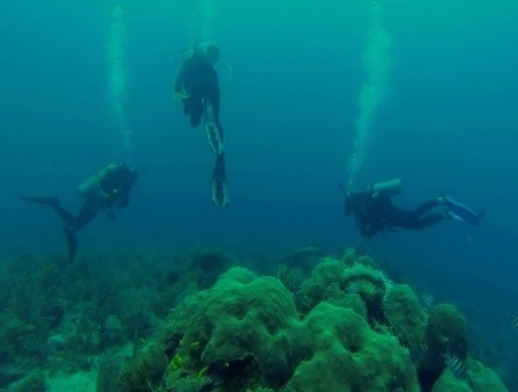}
        \includegraphics[width=0.49\linewidth]{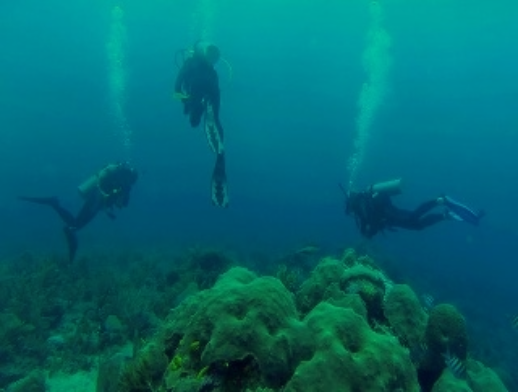}
   \end{subfigure}
   
    \begin{subfigure}{0.35\textwidth}
       \includegraphics[width=0.49\linewidth]{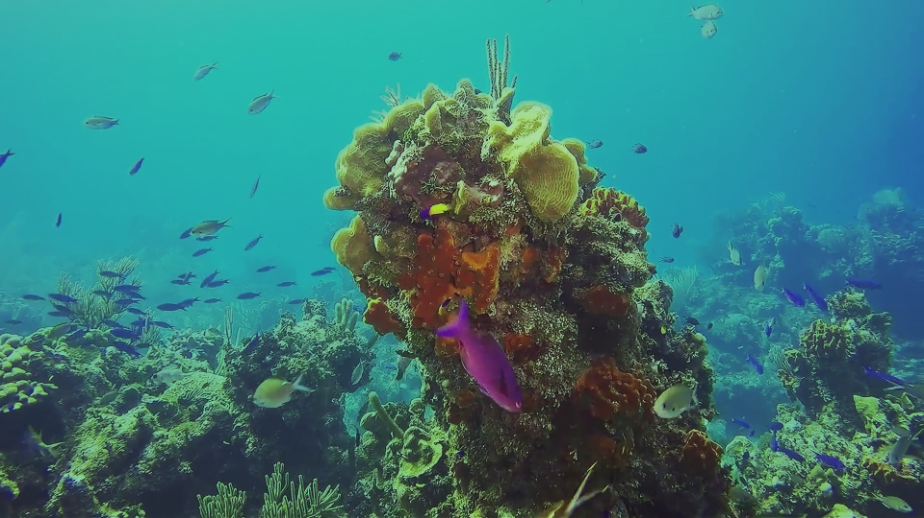}
        \includegraphics[width=0.49\linewidth]{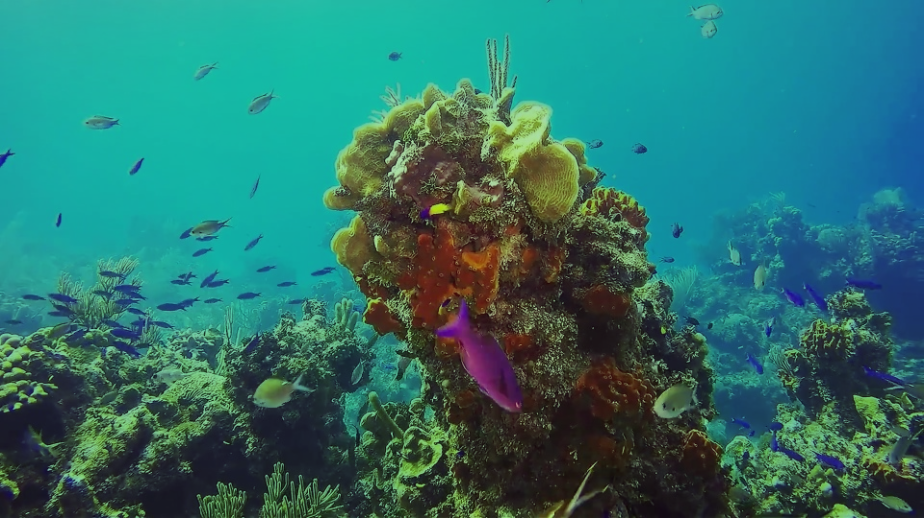}
   \end{subfigure}
    \begin{subfigure}{0.35\textwidth}
       \includegraphics[width=0.49\linewidth]{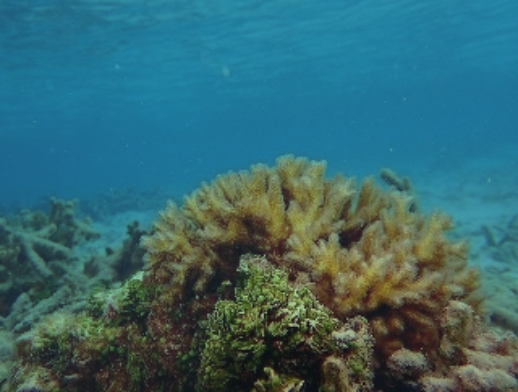}
        \includegraphics[width=0.49\linewidth]{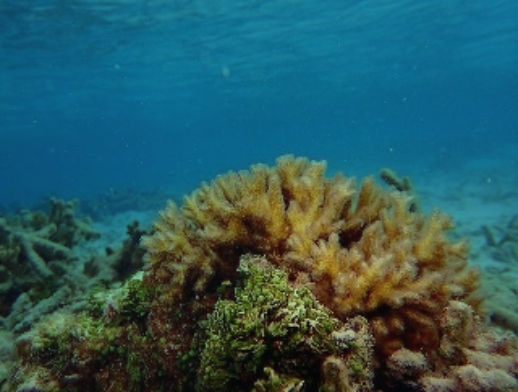}
   \end{subfigure}
   
    \begin{subfigure}{0.35\textwidth}
       \includegraphics[width=0.49\linewidth]{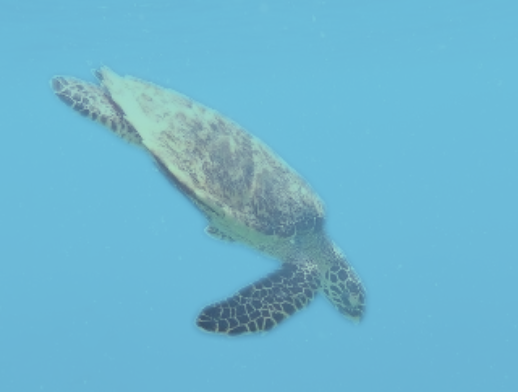}
        \includegraphics[width=0.49\linewidth]{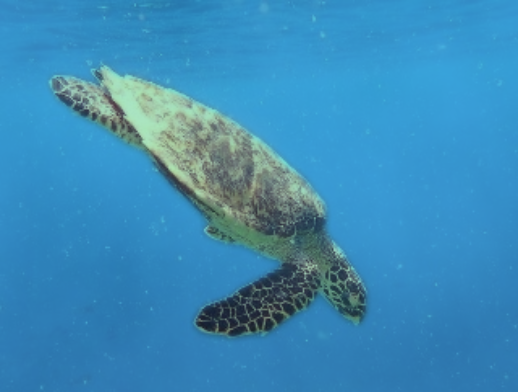}
   \end{subfigure}
    \begin{subfigure}{0.35\textwidth}
       \includegraphics[width=0.49\linewidth]{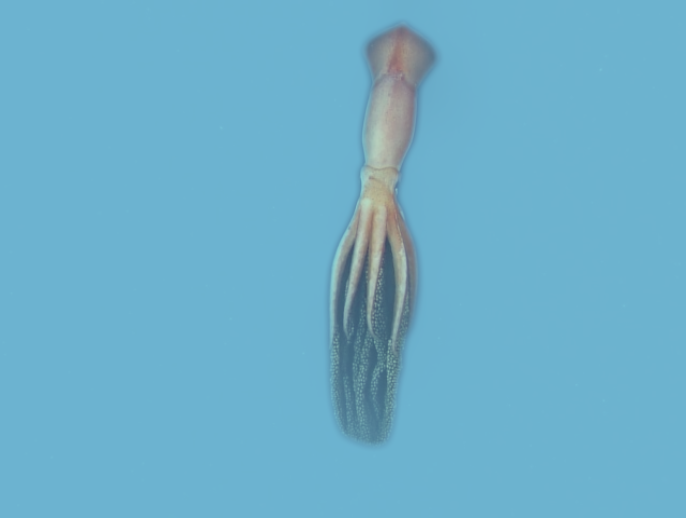}
        \includegraphics[width=0.49\linewidth]{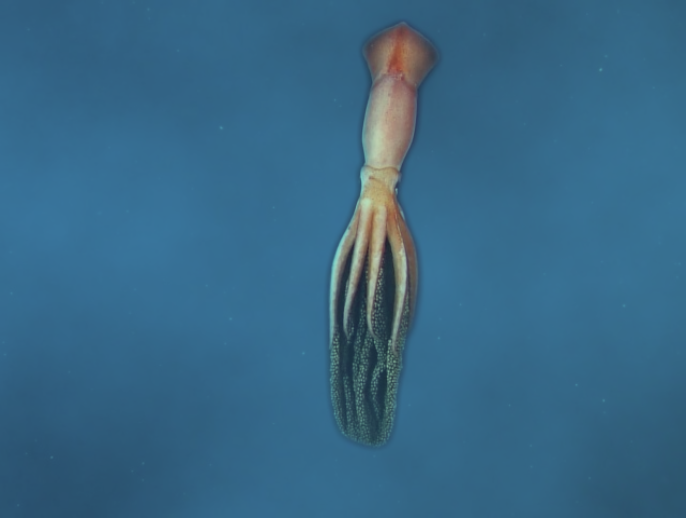}
   \end{subfigure}

    \begin{subfigure}{0.35\textwidth}
       \includegraphics[width=0.49\linewidth]{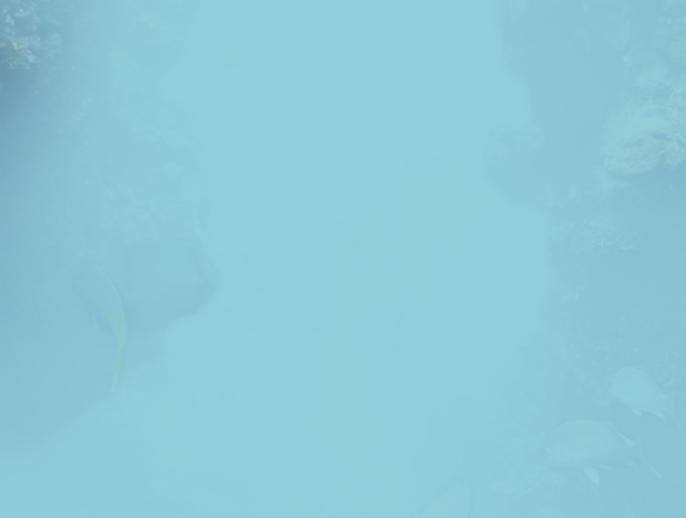}
        \includegraphics[width=0.49\linewidth]{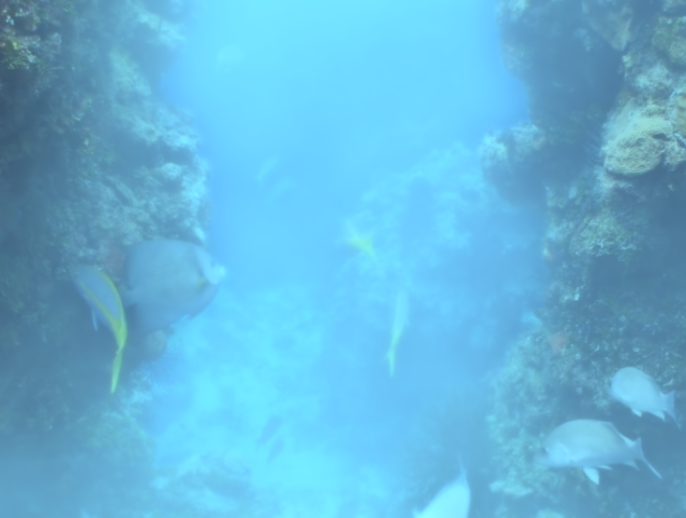}
   \end{subfigure}
    \begin{subfigure}{0.35\textwidth}
       \includegraphics[width=0.49\linewidth]{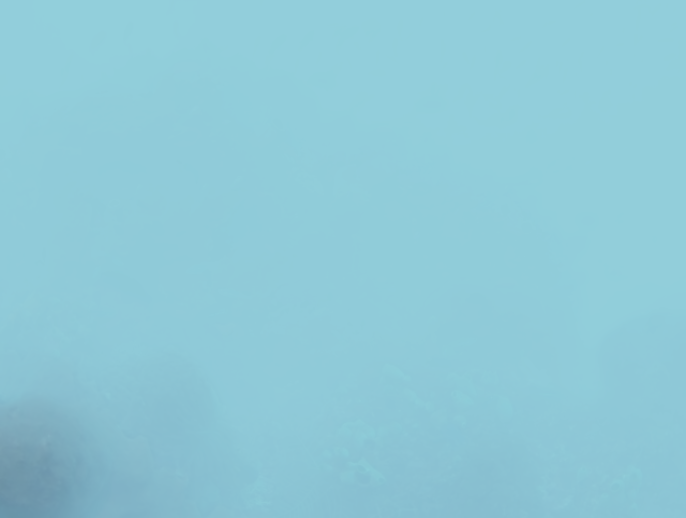}
        \includegraphics[width=0.49\linewidth]{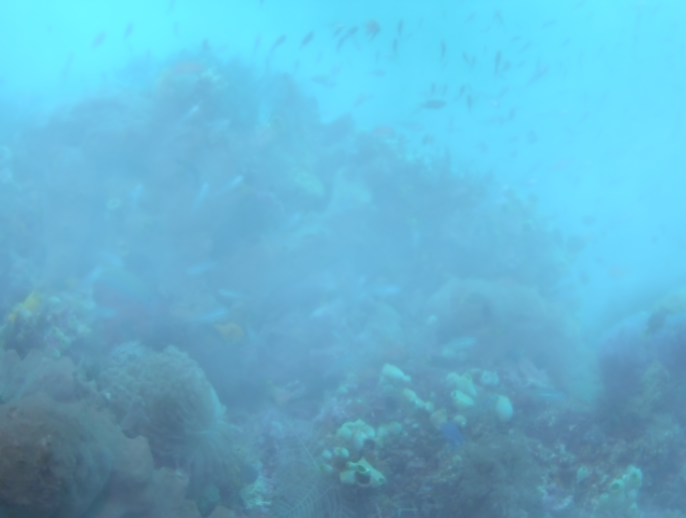}
   \end{subfigure}

    \begin{subfigure}{0.35\textwidth}
       \includegraphics[width=0.49\linewidth]{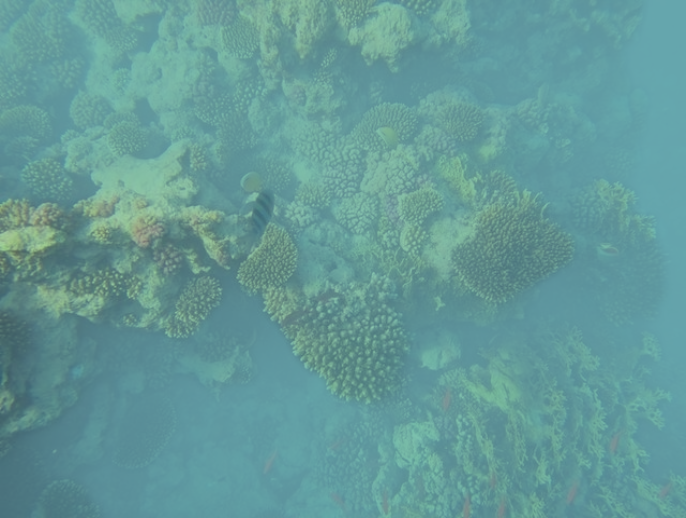}
        \includegraphics[width=0.49\linewidth]{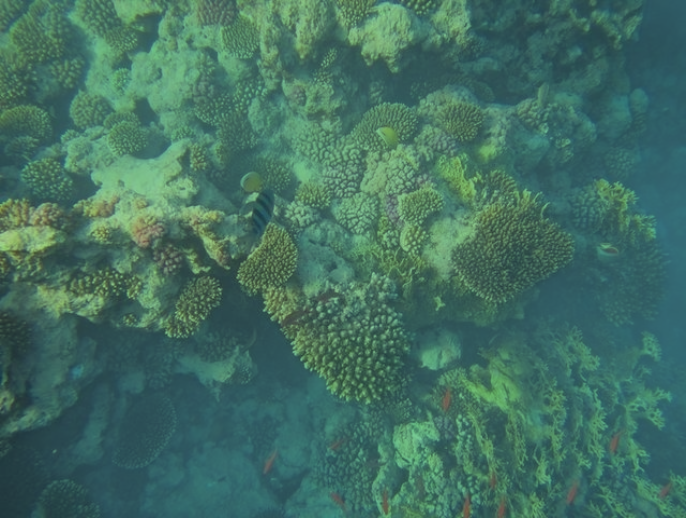}
   \end{subfigure}
    \begin{subfigure}{0.35\textwidth}
       \includegraphics[width=0.49\linewidth]{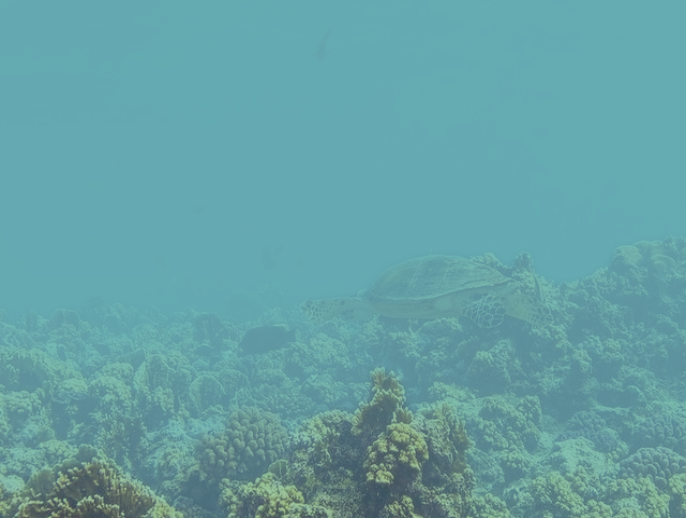}
        \includegraphics[width=0.49\linewidth]{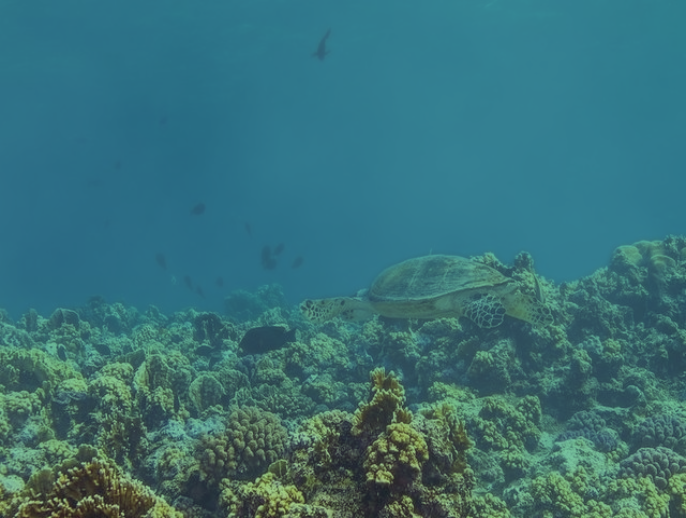}
   \end{subfigure}

    \begin{subfigure}{0.35\textwidth}
       \includegraphics[width=0.49\linewidth]{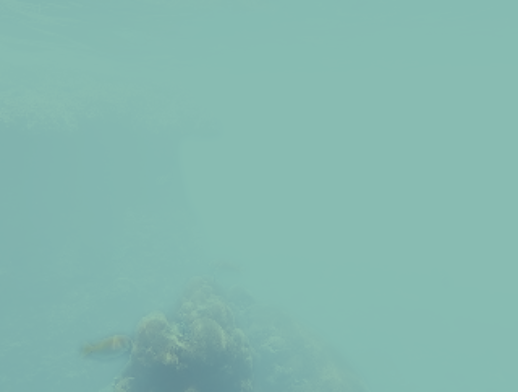}
        \includegraphics[width=0.49\linewidth]{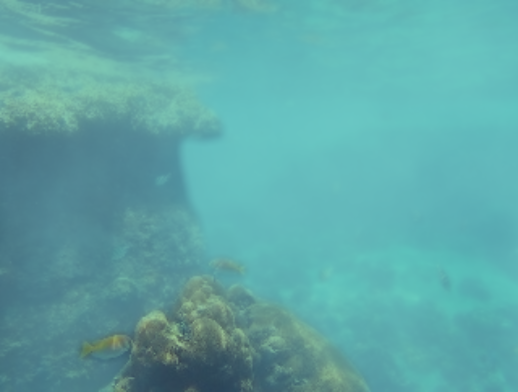}
   \end{subfigure}
    \begin{subfigure}{0.35\textwidth}
       \includegraphics[width=0.49\linewidth]{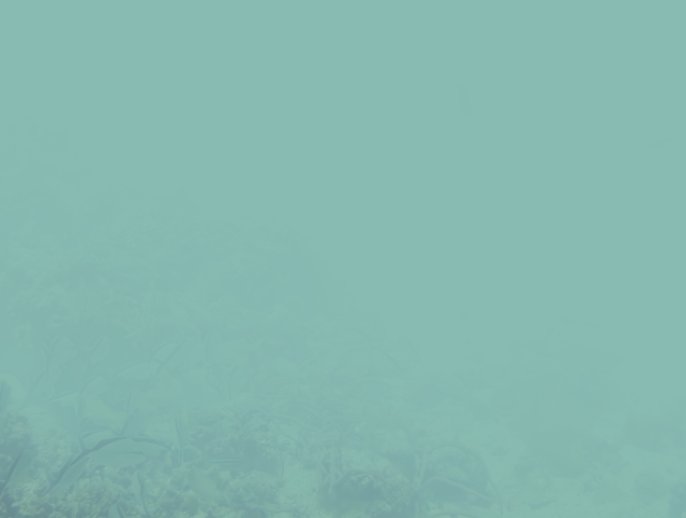}
        \includegraphics[width=0.49\linewidth]{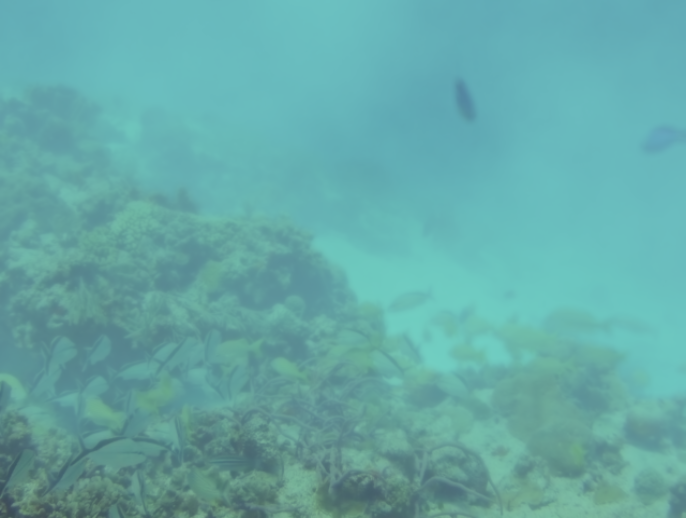}
   \end{subfigure}

    \begin{subfigure}{0.35\textwidth}
       \includegraphics[width=0.49\linewidth]{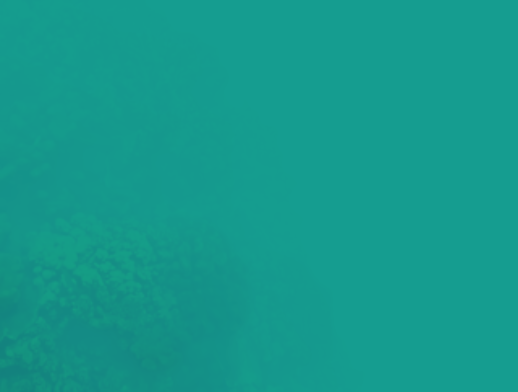}
        \includegraphics[width=0.49\linewidth]{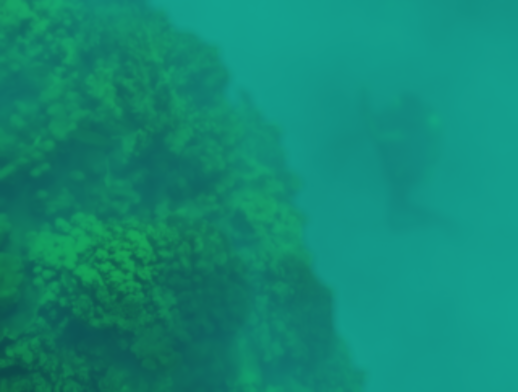}
   \end{subfigure}
    \begin{subfigure}{0.35\textwidth}
       \includegraphics[width=0.49\linewidth]{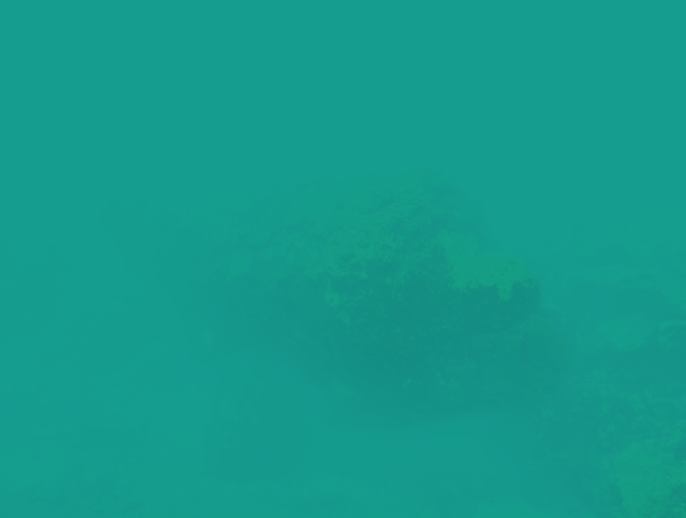}
        \includegraphics[width=0.49\linewidth]{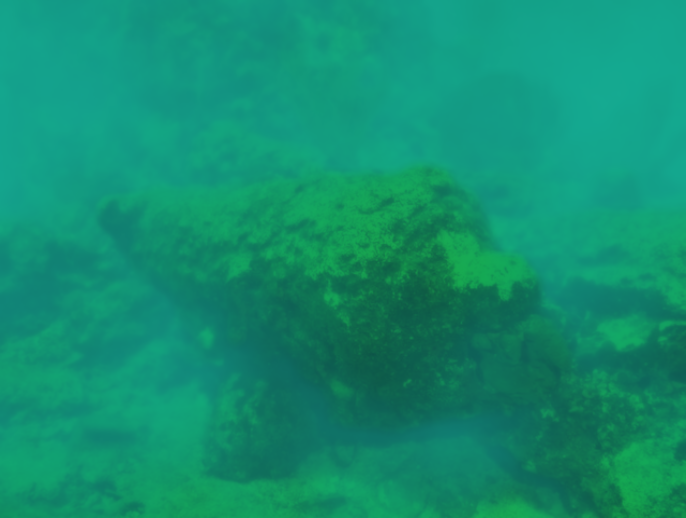}
   \end{subfigure}

    \begin{subfigure}{0.35\textwidth}
       \includegraphics[width=0.49\linewidth]{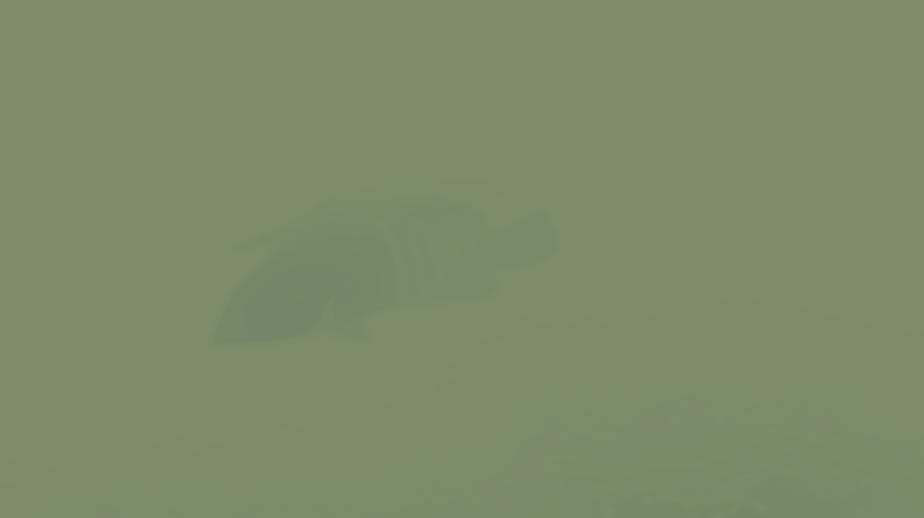}
        \includegraphics[width=0.49\linewidth]{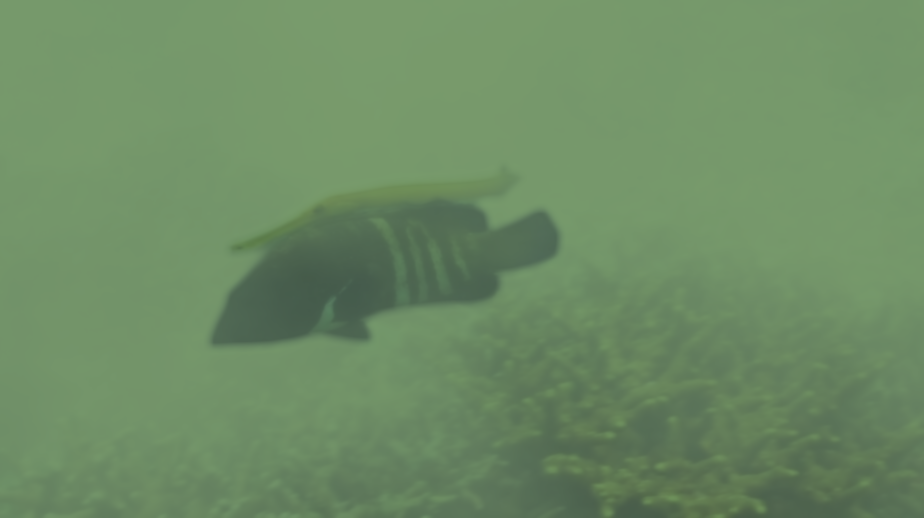}
   \end{subfigure}
    \begin{subfigure}{0.35\textwidth}
       \includegraphics[width=0.49\linewidth]{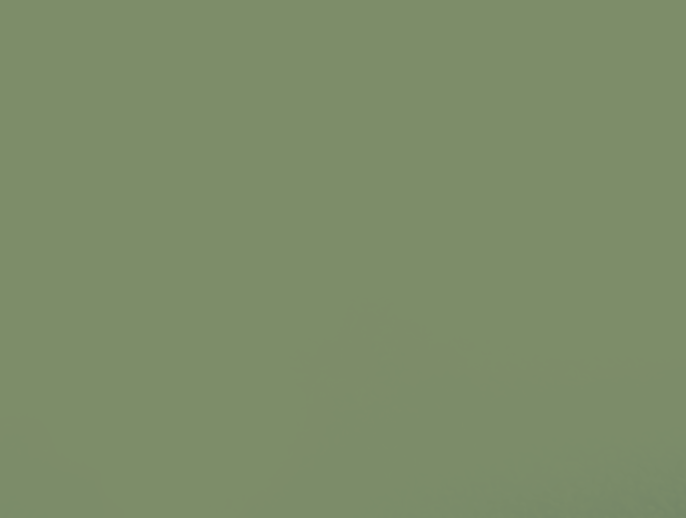}
        \includegraphics[width=0.49\linewidth]{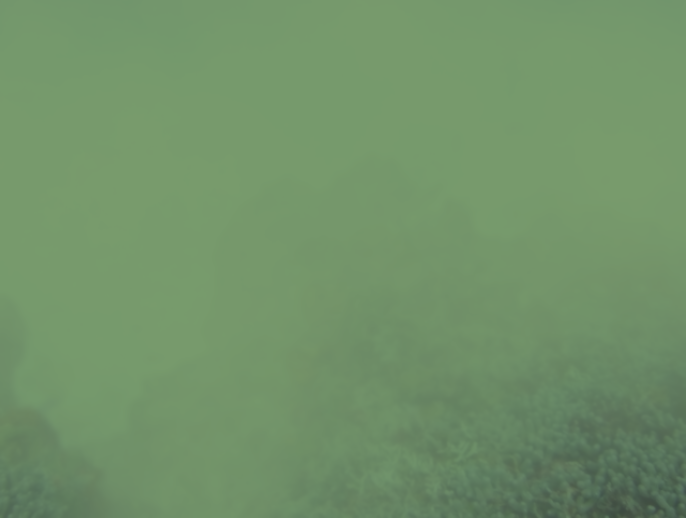}
   \end{subfigure}

    \begin{subfigure}{0.35\textwidth}
       \includegraphics[width=0.49\linewidth]{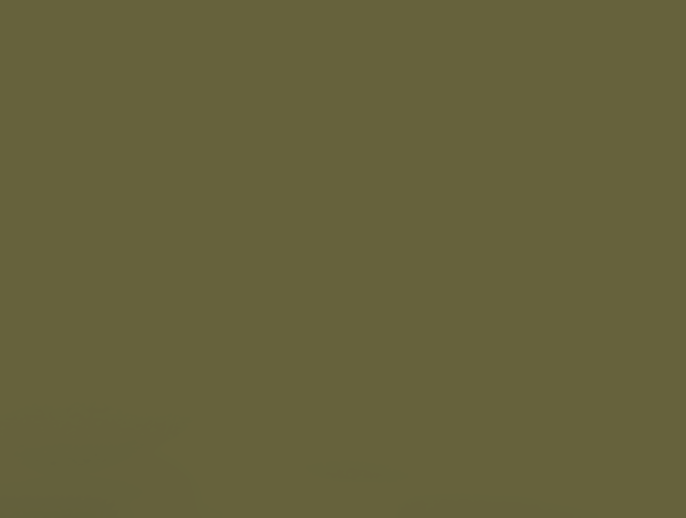}
        \includegraphics[width=0.49\linewidth]{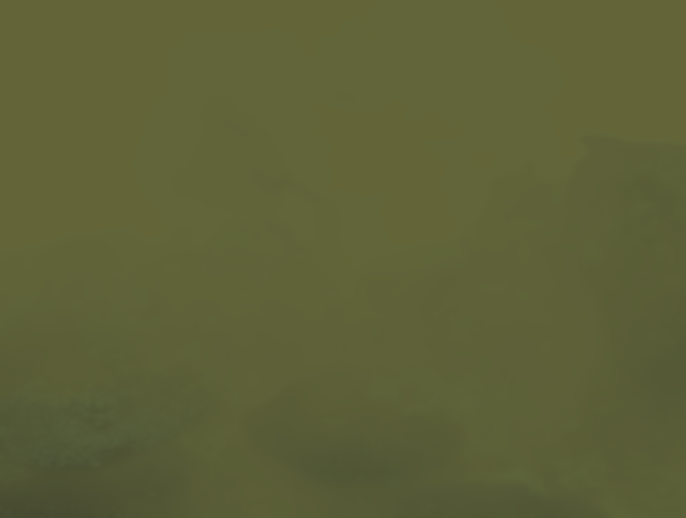}
   \end{subfigure}
    \begin{subfigure}{0.35\textwidth}
       \includegraphics[width=0.49\linewidth]{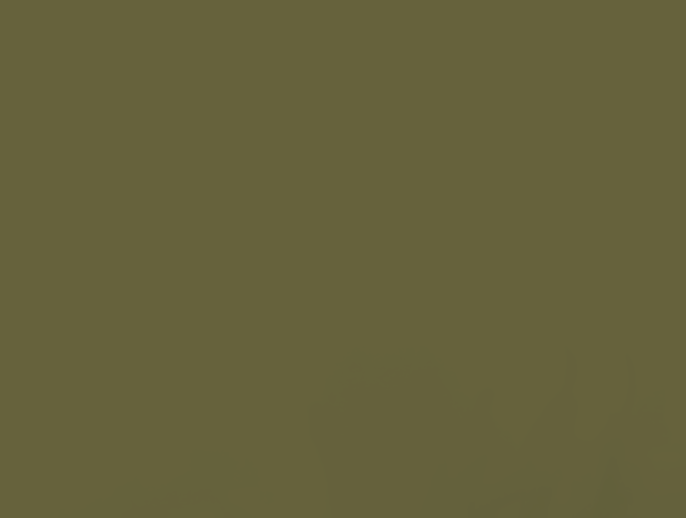}
        \includegraphics[width=0.49\linewidth]{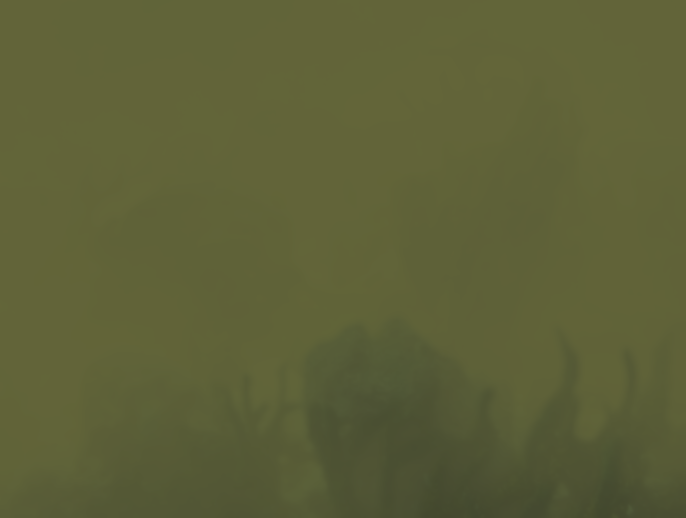}
   \end{subfigure}

   \label{}
   \caption{Survey images. Rows correspond to water types IA - 9C. For all images the following parameters were used: $g=0.2$, $\mu=0.3$, $\phi=0.3\cdot mean(b)$, $d=1m$, $z(x)=[1-5m]$, $GRF=[0.7-1.3]$ }
\end{figure*}